\documentclass[10pt,twocolumn,letterpaper]{article}

\usepackage[pagenumbers]{cvpr} %

\definecolor{cvprblue}{rgb}{0.21,0.49,0.74}
\usepackage[pagebackref,breaklinks,colorlinks,allcolors=cvprblue]{hyperref}

\usepackage[T1]{fontenc}
\usepackage{booktabs}
\usepackage[table,dvipsnames]{xcolor}
\usepackage{makecell}
\usepackage{rotating}
\usepackage{tablefootnote}
\usepackage{enumitem}
\usepackage{subcaption}
\usepackage{multirow}
\usepackage{tcolorbox}
\usepackage{listings}
\usepackage[frozencache,cachedir=minted-cache]{minted}
\usepackage{pifont}
\usepackage{titletoc}
\usepackage{wrapfig}
\usepackage{graphicx}   
\usepackage{adjustbox}
\usepackage{amssymb}
\usepackage{stfloats}
\usepackage{cuted}
\usepackage{capt-of}
\usepackage{fontawesome5}
\usepackage[accsupp]{axessibility}  %

\definecolor{star}{HTML}{FFBE7A}

\definecolor{first}{HTML}{98CC70}
\definecolor{second}{HTML}{DFE674}
\definecolor{third}{HTML}{FFECB2}
\definecolor{below}{HTML}{FF8888}
\definecolor{equal}{HTML}{FFCCCC}

\newcommand{\cmark}{\textcolor{LimeGreen}{\ding{51}}}
\newcommand{\xmark}{\textcolor{red}{\ding{55}}}
\newcommand{\onestar}{\ding{73}}
\newcommand{\twostar}{\ding{73}\ding{73}}
\newcommand{\threestar}{\ding{73}\ding{73}\ding{73}}

\newcommand{\hflogo}{\raisebox{-0.25\height}{\includegraphics[height=1.45em]{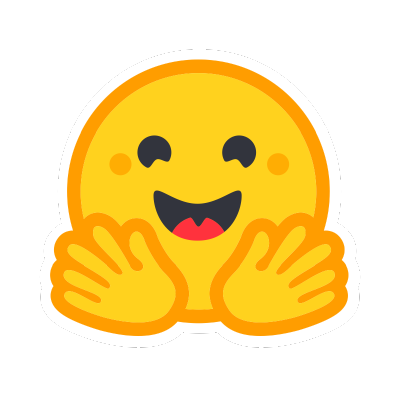}}}
\newcommand{\ghlogo}{\raisebox{-0.2\height}{\includegraphics[height=1.3em]{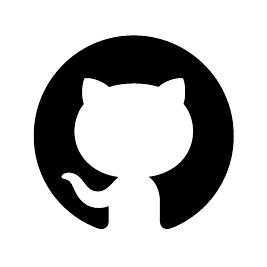}}}
\newcommand\blfootnote[1]{%
  \begingroup
  \renewcommand\thefootnote{}\footnote{#1}%
  \addtocounter{footnote}{-1}%
  \endgroup
}

\def\paperID{8083} %
\def\confName{CVPR}
\def\confYear{2026}

\title{\textsc{VS-Bench}: Evaluating VLMs for Strategic Abilities\\ in Multi-Agent Environments\vspace{-2mm}}

\author{Zelai Xu$^{1*}$, Zhexuan Xu$^{1*}$, Xiangmin Yi$^2$, Huining Yuan$^2$,\\
Mo Guang$^3$, Kaiwen Long$^3$, Xinlei Chen$^2$, Yi Wu$^4$, Chao Yu$^{2\dagger}$, Yu Wang$^{1\dagger}$\\
$^1$EE, Tsinghua University, $^2$SIGS, Tsinghua University, $^3$Li Auto Inc., $^4$IIIS, Tsinghua University\\
$^*$Equal Contribution, $^\dagger$Corresponding Authors\\
{\tt\small zelai.eecs@gmail.com, yuchao@sz.tsinghua.edu.cn, yu-wang@tsinghua.edu.cn}\\
\href{https://vs-bench.github.io/}{{\color{black}\faGlobe}~\,Project Page} \quad
\href{https://github.com/zelaix/VS-Bench}{{\color{black}\ghlogo}~Code} \quad
\href{https://huggingface.co/datasets/zelaix/VS-Bench}{{\color{black}\hflogo}~Dataset}
}

\begin{document}
\maketitle

\begin{strip}
\hfill
\begin{minipage}[t]{0.49\linewidth}
    \centering
    \includegraphics[width=\linewidth,trim={0 10 0 10},clip]{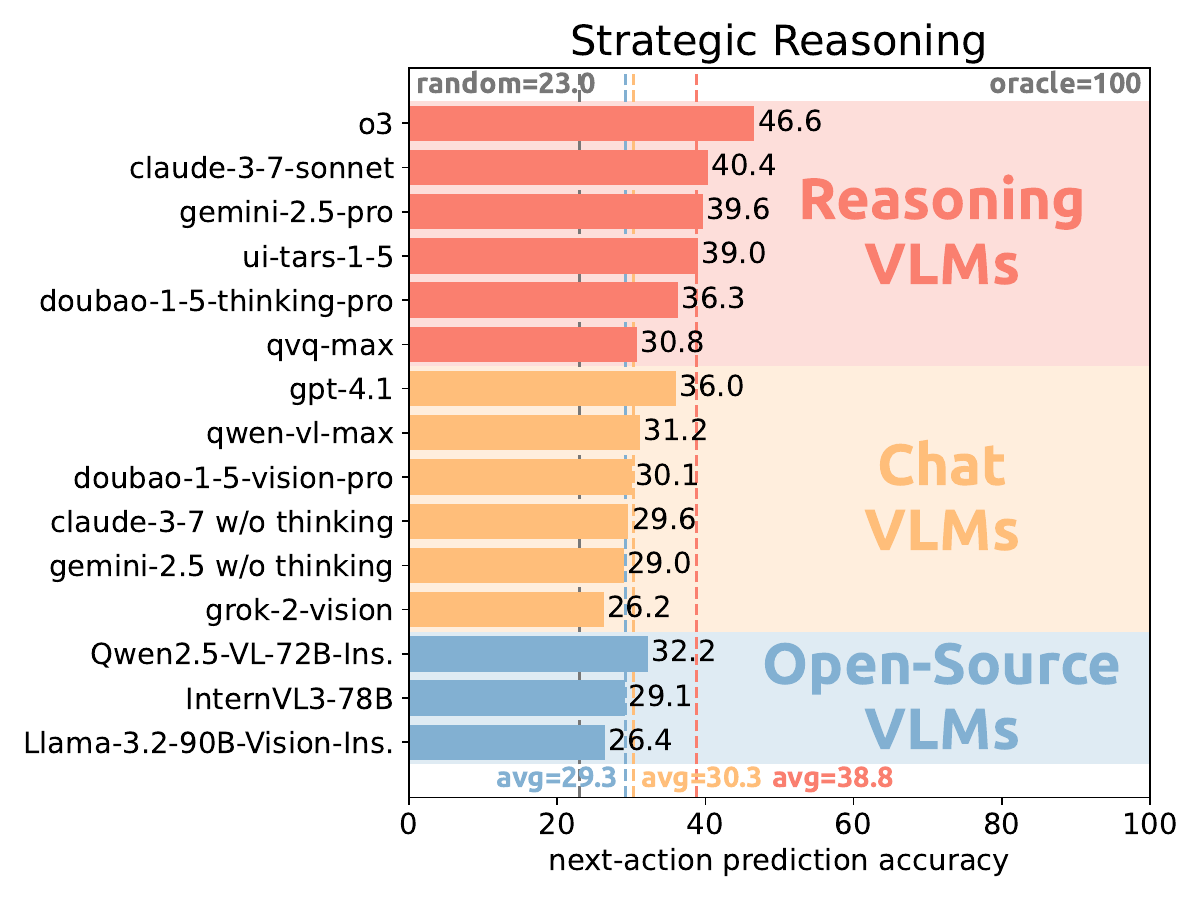}
\end{minipage}
\hfill
\begin{minipage}[t]{0.49\linewidth}
    \centering
    \includegraphics[width=\linewidth,trim={0 10 0 10},clip]{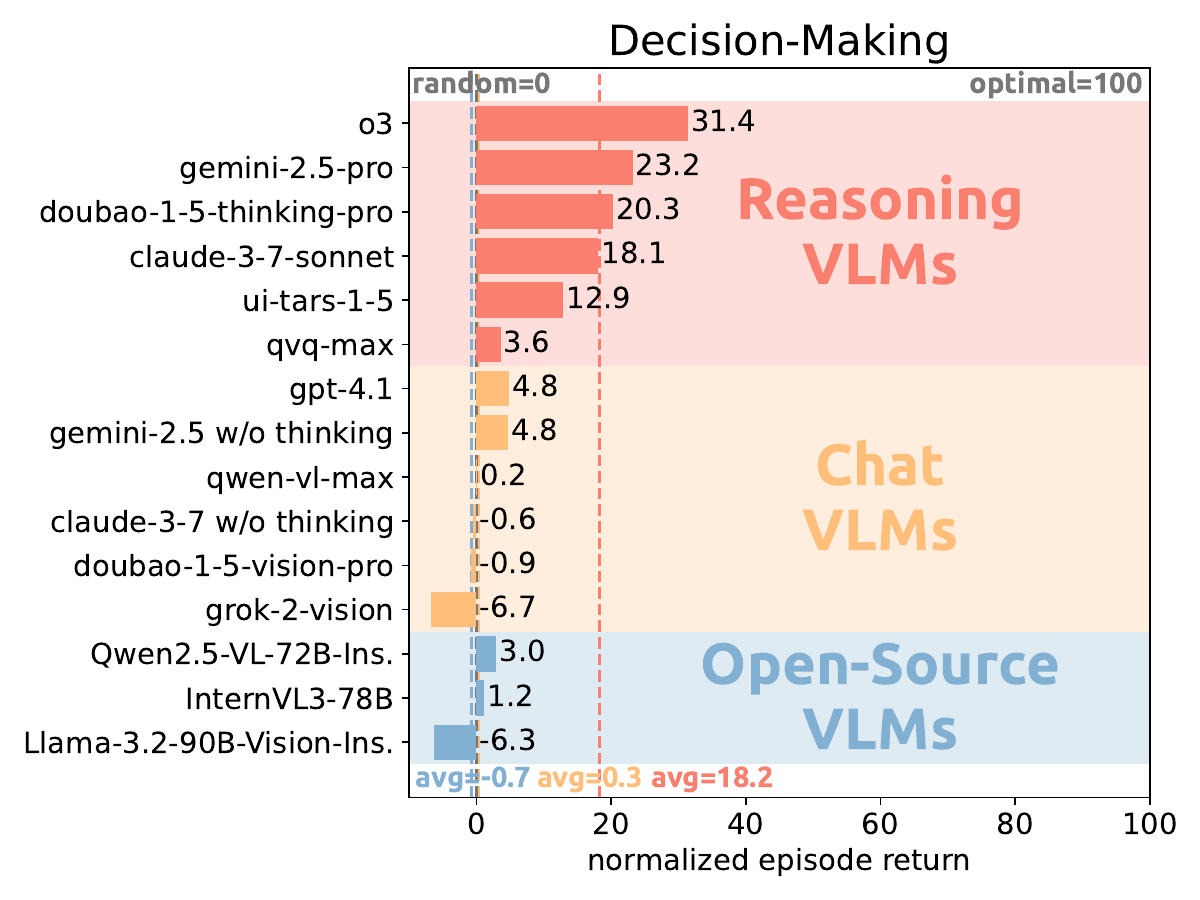}
\end{minipage}
\hfill
\captionof{figure}{Evaluation results of fifteen VLMs on strategic reasoning and decision-making over ten multi-agent environments in \textsc{VS-Bench}.}
\label{fig:teaser}
\end{strip}

\begin{abstract}
Recent advancements in Vision Language Models (VLMs) have expanded their capabilities to interactive agent tasks, yet existing benchmarks remain limited to single-agent or text-only environments.\blfootnote{Published as a conference paper at CVPR 2026 (Oral).} In contrast, real-world scenarios often involve multiple agents interacting within rich visual and textual contexts, posing challenges with both multimodal observations and strategic interactions. To bridge this gap, we introduce \textsc{Visual Strategic Bench} (\textsc{VS-Bench}), a multimodal benchmark that evaluates VLMs for strategic abilities in multi-agent environments. \textsc{VS-Bench} comprises ten vision-grounded environments that cover cooperative, competitive, and mixed-motive interactions. The performance of VLM agents is evaluated across three dimensions: perception measured by element recognition accuracy; strategic reasoning measured by next-action prediction accuracy; and decision-making measured by normalized episode return. Extensive experiments on fifteen leading VLMs show that, although current models exhibit strong perception abilities, there remains a significant gap to optimal performance in reasoning and decision-making, with the best-performing model attaining 46.6\% prediction accuracy and 31.4\% normalized return. We further analyze the key factors influencing performance, conduct human studies, and examine failure modes to provide a deeper understanding of VLMs' strategic abilities. By standardizing the evaluation and highlighting the limitations of existing models, we envision \textsc{VS-Bench} as a foundation for future research on strategic multimodal agents. 

\end{abstract}

\section{Introduction}
\label{sec:intro}
\begin{figure*}[t]
\centering
\includegraphics[width=0.95\linewidth]{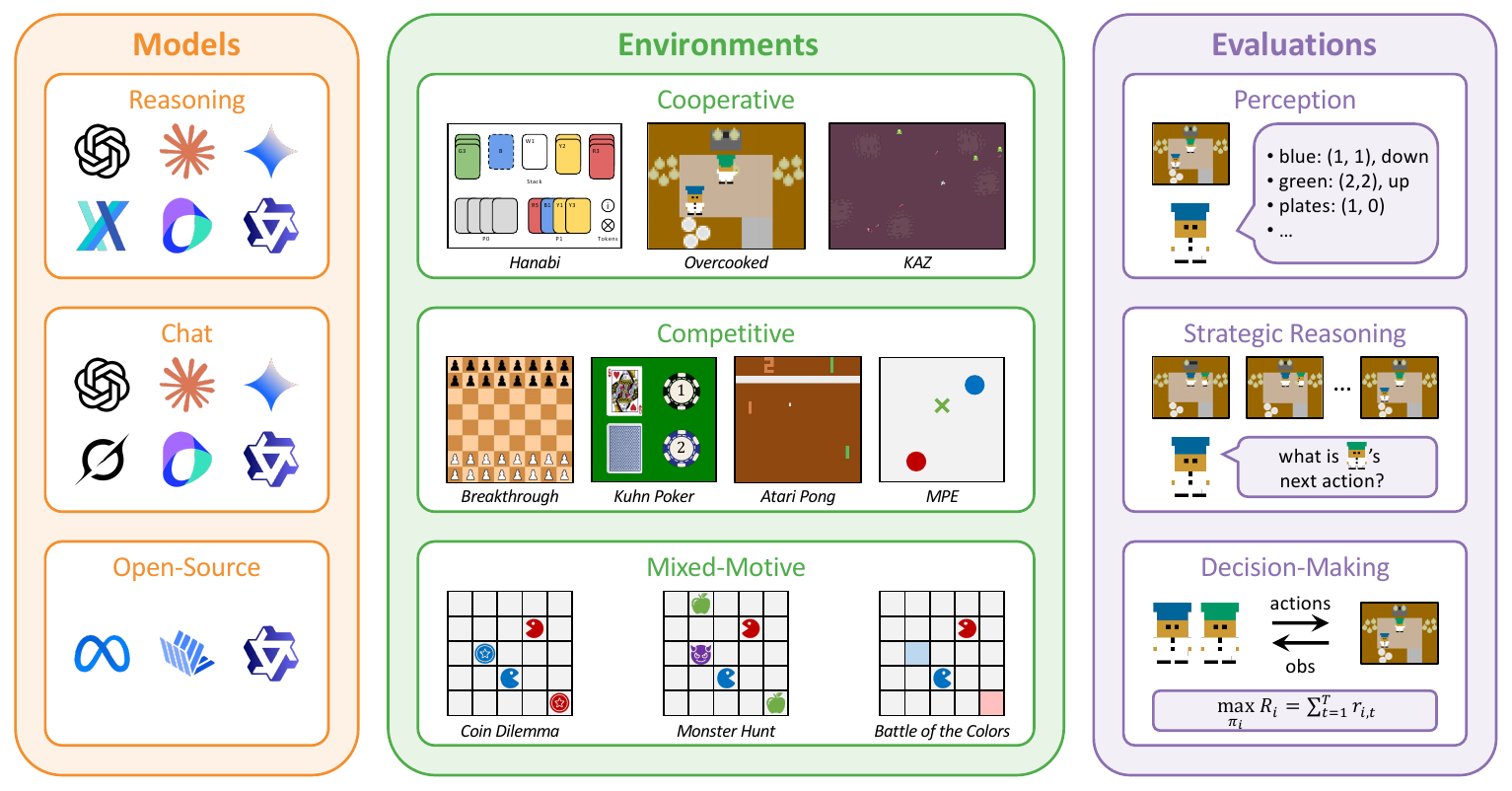}
\caption{Overview of \textsc{VS-Bench}, a multimodal benchmark for evaluating VLMs in multi-agent environments. We evaluate fifteen models in ten vision-grounded environments across three dimensions, including perception measured by element recognition accuracy, strategic reasoning measured by next-action prediction accuracy, and decision-making measured by normalized episode return.}
\label{fig:overview}
\end{figure*}

Vision Language Models (VLMs) have recently unlocked impressive capabilities in open-world perception, multimodal reasoning, and interactive problem-solving~\citep{bai2023qwen,liu2023visual}. 
Driven by these advancements, evaluations of VLMs have progressed beyond static tasks like image captioning~\citep{chen2015microsoft} and visual reasoning~\citep{antol2015vqa,yue2024mmmu} toward agentic benchmarks including software engineering~\citep{chan2024mle,yang2024swe}, computer use~\citep{koh2024visualwebarena,xie2024osworld}, game play~\citep{zhang2023creative,wang2025large}, and embodied control~\citep{guruprasad2024benchmarking,team2025gemini}.

However, existing VLM benchmarks mainly focus on single-agent settings, where one agent reasons and acts in isolation. 
The real world, by contrast, is inherently a multi-agent environment that involves cooperation, competition, and mixed-motive interactions~\citep{ferber1999multi,wooldridge2009introduction}. 
This poses new challenges to the ability of intelligent agents.
First, an agent's outcome depends not only on its own action but also on others' actions, requiring theory-of-mind reasoning~\citep{kosinski2023theory} to infer others' intentions and future moves.
Second, as all agents learn and adapt concurrently, the underlying dynamics become non-stationary, demanding decision-making under uncertainty that optimizes for long-term objectives.
Third, the coexistence of cooperation and competition gives rise to social dilemmas where agents must strategically balance self-interest and collective welfare.
These challenges raise a crucial question that current benchmarks leave underexplored:
\textbf{How capable are VLMs in terms of strategic abilities in multi-agent environments?}

While prior efforts~\citep{agashe2023llm,duan2024gtbench,xie2024can} have explored multi-agent evaluation for Large Language Models (LLMs), these benchmarks remain restricted to text-only environments, limiting their capability to assess agents in multimodal scenarios.
On the one hand, many strategic domains~\citep{bellemare2013arcade,lowe2017multi,carroll2019utility} intrinsically rely on visual observations like game frames that encode rich spatial layouts, motion cues, and temporal dynamics. 
Reducing these visual observations to text-only descriptions inevitably discards critical information that is essential for reasoning and decision-making.
On the other hand, humans naturally receive and convey both visual and textual information when interacting with others.
Purely text-based environments diverge from real-world human-agent interactions and obscure progress toward developing human-compatible intelligent agents.
These limitations underscore the need for a multimodal benchmark that incorporates visual context in multi-agent environments.

To bridge this gap, we introduce \textsc{Visual Strategic Bench} (\textsc{VS-Bench}), a multimodal benchmark that evaluates VLMs' strategic abilities in multi-agent environments.
\textsc{VS-Bench} comprises ten vision-grounded environments that cover three multi-agent dynamics including cooperative, competitive, and mixed-motive interactions.
VLMs are evaluated across three dimensions: perception, strategic reasoning, and decision-making.
Perception requires extracting information from multimodal observations and is evaluated by the recognition accuracy of basic visual elements.
Strategic reasoning is the theory-of-mind ability to infer others' intentions and is evaluated by the prediction accuracy of others' next actions.
Decision-making is the ability to optimize for long-term objectives under non-stationary dynamics and is evaluated by the normalized episode returns.
By jointly analyzing all three perspectives, our benchmark provides a comprehensive evaluation of VLMs' strategic abilities in multi-agent environments.

We evaluate fifteen leading VLMs, including six commercial reasoning models, six commercial chat models, and three open-source models on \textsc{VS-Bench}.
Extensive results show that while current VLMs exhibit strong perception ability, there remains a significant gap to optimal performance in strategic reasoning and decision-making, with the best-performing model attaining $46.6$\% prediction accuracy and $31.4$\% normalized return averaged across all environments.
Notably, although commercial reasoning models generally achieve better results, open-source models can achieve comparable performance with mutual benefit behaviors in some mixed-motive games.
We further perform analyses on multimodal observations, test-time scaling, social behaviors, human study, and failure modes to provide a deeper understanding of VLMs' strategic abilities.

In summary, our contributions are threefold:
\begin{itemize}
    \item We introduce \textsc{VS-Bench}, a multimodal benchmark for evaluating VLMs in ten vision-grounded multi-agent environments across three interaction types.
    \item We consider three evaluation dimensions, including perception, strategic reasoning, and decision-making to provide a comprehensive assessment of VLM agents.
    \item We perform extensive experiments of twelve commercial VLMs and three open-source VLMs and provide in-depth analyses, highlighting significant gaps for future research.
\end{itemize}

\section{\textsc{VS-Bench} environments}
\label{sec:envs}
\begin{table*}[t]
\centering
\caption{Taxonomy and required abilities of the ten multi-agent environments in \textsc{VS-Bench}.}
\label{tab:envs}
\small
\setlength{\tabcolsep}{7.5pt}
\begin{tabular}{cccccccccc}
\toprule
\multirow{2}{*}{Type} & \multirow{2}{*}{Games} & {Full} & {Deter-} & {Simul-} & {Sym-} & {Spatial} & {Theory} & {Long-Term} & {Team} \\
& & Obs. & {ministic} & {taneous} & {metric} & {Perception} & {of Mind} & {Planning} & {Collaboration} \\
\midrule
\multirow{3}{*}{Cooperative} & \textit{Hanabi} & {\small \xmark} & {\small \xmark} & {\small \xmark} & {\small \xmark} & {\small \onestar} &  {\small \threestar} &  {\small \threestar} &  {\small \threestar} \\
& \textit{Overcooked} & {\small \cmark} & {\small \cmark} & {\small \cmark} & {\small \xmark} &  {\small \twostar} &  {\small \twostar} &  {\small \twostar} &  {\small \threestar} \\
& \textit{KAZ} & {\small \cmark} & {\small \xmark} & {\small \cmark} & {\small \xmark} & {\small \threestar} & {\small \onestar} & {\small \twostar} & {\small \twostar} \\
\midrule
\multirow{4}{*}{Competitive} & \textit{Board}
\footnotemark
& {\small \cmark} & {\small \cmark} & {\small \xmark} & {\small \xmark} &  {\small \twostar} &  {\small \threestar} &  {\small \threestar} &  {\small \onestar} \\
& \textit{Poker} & {\small \xmark} & {\small \xmark} & {\small \xmark} & {\small \xmark} &  {\small \onestar} &  {\small \threestar} &  {\small \onestar} &  {\small \onestar} \\
& \textit{Pong} & {\small \cmark} & {\small \xmark} & {\small \cmark} & {\small \cmark} &  {\small \threestar} &  {\small \onestar} &  {\small \onestar} &  {\small \onestar} \\
& \textit{MPE} & {\small \cmark} & {\small \xmark} & {\small \cmark} & {\small \xmark} &  {\small \threestar} &  {\small \twostar} &  {\small \twostar} &  {\small \onestar} \\
\midrule
\multirow{3}{*}{\makecell{Mixed-\\Motive}} & \textit{Dilemma} & {\small \cmark} & {\small \xmark} & {\small \cmark} & {\small \cmark} &  {\small \twostar} &  {\small \twostar} &  {\small \twostar} &  {\small \onestar} \\
& \textit{Hunt} & {\small \cmark} & {\small \xmark} & {\small \cmark} & {\small \cmark} &  {\small \twostar} &  {\small \threestar} &  {\small \twostar} &  {\small \twostar} \\
& \textit{Battle} & {\small \cmark} & {\small \xmark} & {\small \cmark} & {\small \cmark} &  {\small \twostar} &  {\small \twostar} &  {\small \twostar} &  {\small \threestar} \\
\bottomrule
\end{tabular}

\end{table*}
\footnotetext{\textit{Board} corresponds to \textit{Breakthrough}, and the remaining rows below it correspond to \textit{Kuhn \underline{Poker}}, \textit{Atari \underline{Pong}}, \textit{\underline{MPE}}, \textit{Coin \underline{Dilemma}}, \textit{Monster \underline{Hunt}}, and \textit{\underline{Battle} of the Colors}.}

In this section, we first formalize the evaluation of VLMs in multi-agent environments and then introduce ten vision-grounded games comprising \textsc{VS-Bench}. 
These games are carefully curated from game theory and multi-agent reinforcement learning (MARL), each serving as a well-recognized environment in the literature.
We further adapt these games to incorporate image and text observations while preserving their strategic dynamics.
By covering cooperation, competition, and mixed-motive interactions, these games provide a diverse set of multi-agent environments for evaluating VLMs.
An additional set of simpler games and three-player games can be found in Appendix I.

\subsection{Problem formulation}

Multi-agent environments are modeled as Partially Observable Markov Games (POMG)~\citep{shapley1953stochastic}.
A POMG is defined by a tuple $\mathcal{G}=(\mathcal{N}, \mathcal{S}, \{\mathcal{A}_i\}_{i\in\mathcal{N}}, \{\mathcal{O}_i\}_{i\in\mathcal{N}}, \mathcal{P}, \{\mathcal{R}_i\}_{i\in\mathcal{N}}, \gamma)$, where
$\mathcal{N}=\{1,\cdots,n\}$ is the set of agents;
$\mathcal{S}$ is the state space;
$\mathcal{A}_i$ and $\mathcal{O}_i$ are the action space and observation space of agent $i$, respectively;
$\mathcal{P}:\mathcal{S}\times\{\mathcal{A}_i\}_{i\in\mathcal{N}}\rightarrow\Delta(\mathcal{S})$ is the transition function;
$\mathcal{R}_i:\mathcal{S}\times\{\mathcal{A}_i\}_{i\in\mathcal{N}}\rightarrow\mathbb{R}$ is the reward function of agent $i$;
and $\gamma$ is the discount factor.
In step $t$, each agent $i$ receives an observation $o_{i,t}$ and chooses an action $a_{i,t}$ according to its policy $\pi_i$.
Given the current state $s_t$ and the joint action $a_t=(a_{1,t}, \cdots, a_{n,t})$, the environment transitions to the next state $s_{t+1}\sim\mathcal{P}(s_t, a_t)$ and each agent $i$ receive a reward $r_{i,t}=\mathcal{R}_i(s_t, a_t)$. 
The objective of agent $i$ is to maximize its expected accumulated reward $\mathbb{E}_{\pi_1,\cdots,\pi_n}\left[\sum_t\gamma^t r_{i,t}\right]$.
To evaluate VLM agents, we consider a multimodal observation space $\mathcal{O}_i = (\mathcal{I}_i, \mathcal{T}_i)$, where $\mathcal{I}_i$ is the image observation space and $\mathcal{T}_i$ is the text observation space.
We also consider a text action space $\tilde{\mathcal{A}}_i$ and a mapping function $\mathcal{M}:\tilde{\mathcal{A}}_i \rightarrow \mathcal{A}_i$ that converts each text action into the original action space.

\subsection{Multi-agent environments}
\label{sec:envs:envs}

We consider three types of multi-agent environments, including cooperative, competitive, and mixed-motive games.
The taxonomy and required abilities of the games in \textsc{VS-Bench} are listed in Table~\ref{tab:envs}. 
The detailed description of each game and required ability can be found in Appendix J.

\textbf{Cooperative.}
All agents share the same objective in cooperative games. 
Formally, the reward functions in cooperative games are identical for all agents: $\mathcal{R}_1(s, a)=\cdots=\mathcal{R}_n(s, a)$ for all $(s, a)\in\mathcal{S}\times\{\mathcal{A}_i\}_{i\in\mathcal{N}}$.
To achieve strong performance in cooperative games, agents must understand their teammates' intentions under partial observability, divide the tasks to improve efficiency, and coordinate their actions to optimize for the shared objective.
We consider three representative cooperative games in the MARL literature, namely \textit{Hanabi}~\citep{bard2020hanabi}, \textit{Overcooked}~\citep{ghost2016overcooked}, and \textit{Knights Archers Zombies (KAZ)}~\citep{terry2021pettingzoo}.

\textbf{Competitive.}
The objective of each agent strictly contradicts those of the others in competitive games.
Formally, the reward functions in competitive games are zero-sum: $\sum_{i=1}^n\mathcal{R}_i(s, a)=0$ for all $(s, a)\in\mathcal{S}\times\{\mathcal{A}_i\}_{i\in\mathcal{N}}$.
To succeed in competitive games, agents must model their opponents to predict their future moves, stay robust against adversarial exploitation, and adapt to non-stationary dynamics.
We consider four representative competitive games in game theory and MARL literature, namely \textit{Breakthrough}~\citep{troyka2000breakthrough}, \textit{Kuhn Poker}~\citep{kuhn1950simplified}, \textit{Atari Pong}~\citep{atari1972pong}, and \textit{Multi-agent Particle Environment (MPE)}~\citep{lowe2017multi}.

\textbf{Mixed-motive.}
Agents' objectives are partially aligned and partially divergent in mixed-motive games.
Formally, the reward functions are neither identical nor zero-sum, that is, there exists $(s, a)$ such that $\mathcal{R}_i(s, a)\neq \mathcal{R}_j(s, a)$ and $\sum_{i=1}^n\mathcal{R}_i(s, a)\neq0$.
In mixed-motive games, agents must anticipate the hidden intentions of others, balance self-interest and common welfare, and achieve high-payoff equilibria.
We consider three mixed-motive games adapted from classic social dilemmas in game theory, namely \textit{Coin Dilemma}~\citep{lerer2017maintaining}, \textit{Monster Hunt}~\citep{peysakhovich2017prosocial}, \textit{Battle of the Colors}.

\begin{table*}[t]
\centering
\caption{Perception evaluation results. For each environment, the \colorbox{first}{first}, \colorbox{second}{second}, and \colorbox{third}{third} best results are highlighted in green.}
\label{tab:perception}
\scriptsize
\setlength{\tabcolsep}{6.5pt}
\begin{tabular}{cccccccccccc}
\toprule
\multirow{2}{*}{Models\footnotemark}
& \multirow{2}{*}{Overall} & \multicolumn{3}{c}{Cooperative} & \multicolumn{4}{c}{Comptitive} & \multicolumn{3}{c}{Mixed-Motive} \\
& & \textit{Hanabi} & \textit{Overcooked} & \textit{KAZ} & \textit{Board} 
& \textit{Poker}    & \textit{Pong} & \textit{MPE} & \textit{Dilemma} & \textit{Hunt} & \textit{Battle} \\
\midrule
Oracle & $100.0$ & $100.0$ & $100.0$ & $100.0$ & $100.0$ & $100.0$ & $100.0$ & $100.0$ & $100.0$ & $100.0$ & $100.0$ \\
\midrule
\texttt{o3} & \cellcolor{first} $84.9$ & $79.7$ & \cellcolor{second} $69.8$ & \cellcolor{third} $90.1$ & \cellcolor{third} $97.2$ & $99.1$ & $64.6$ & \cellcolor{second} $97.5$ & \cellcolor{second} $85.5$ & \cellcolor{third} $80.2$ & \cellcolor{third} $85.4$ \\
\texttt{gemini-2.5-pro} & \cellcolor{third} $83.4$ & \cellcolor{second} $79.9$ & $54.5$ & $88.6$ & \cellcolor{first} $98.5$ & \cellcolor{first} $100.0$ & \cellcolor{first} $86.5$ & $96.5$ & $76.4$ & $73.0$ & $79.9$ \\
\texttt{claude-3-7-sonnet} & $81.1$ & $73.0$ & $62.8$ & \cellcolor{first} $92.9$ & $75.2$ & $99.5$ & $69.4$ & $89.8$ & \cellcolor{third} $82.6$ & $79.7$ & \cellcolor{second} $85.7$ \\
\texttt{ui-tars-1-5} & $81.5$ & $75.4$ & $59.8$ & $83.0$ & \cellcolor{second} $98.0$ & \cellcolor{first} $100.0$ & $74.4$ & $92.2$ & $76.0$ & $74.9$ & $81.0$ \\
\texttt{doubao-1-5-thinking-pro} & $74.7$ & $42.2$ & $52.4$ & $77.5$ & $91.0$ & $98.6$ & $66.0$ & $87.2$ & $76.7$ & $74.9$ & $80.0$ \\
\texttt{qvq-max} & $74.5$ & $75.1$ & $63.6$ & $74.6$ & $83.3$ & $95.2$ & $50.7$ & $83.6$ & $69.5$ & $72.0$ & $77.7$ \\
\midrule
\texttt{gemini-2.5} w/o thinking & \cellcolor{second} $84.5$ & \cellcolor{second} $79.9$ & $38.8$ & $82.8$ & $88.2$ & $97.2$ & \cellcolor{second} $84.3$ & \cellcolor{third} $97.3$ & \cellcolor{first} $92.5$ & \cellcolor{first} $93.1$ & \cellcolor{first} $90.8$ \\
\texttt{claude-3-7} w/o thinking & $80.5$ & $75.9$ & $59.7$ & $85.6$ & $79.0$ & $99.6$ & $70.1$ & $90.6$ & $81.8$ & \cellcolor{second} $80.4$ & $82.8$ \\
\texttt{gpt-4.1} & $80.3$ & $72.1$ & $62.0$ & \cellcolor{second} $92.6$ & $67.0$ & \cellcolor{first} $100.0$ & $75.4$ & \cellcolor{first} $98.9$ & $76.7$ & $76.8$ & $81.2$ \\
\texttt{qwen-vl-max} & $80.2$ & $76.1$ & \cellcolor{third} $68.2$ & $85.2$ & $81.2$ & $99.2$ & $69.0$ & $88.2$ & $78.4$ & $76.4$ & $80.6$ \\
\texttt{doubao-1-5-vision-pro} & $77.6$ & \cellcolor{first} $80.0$ & $33.1$ & $67.8$ & $89.3$ & \cellcolor{first} $100.0$ & \cellcolor{third} $77.7$ & $90.8$ & $78.0$ & $77.6$ & $81.2$ \\
\texttt{grok-2-vision} & $70.2$ & $75.2$ & $46.8$ & $59.8$ & $80.3$ & $59.5$ & $71.0$ & $78.7$ & $76.4$ & $73.3$ & $81.0$ \\
\midrule
\texttt{Qwen2.5-VL-72B-Ins.} & $80.3$ & $76.0$ & \cellcolor{first} $72.9$ & $85.5$ & $75.1$ & \cellcolor{first} $100.0$ & $65.2$ & $87.7$ & $79.8$ & $79.0$ & $82.4$ \\
\texttt{InternVL3-78B} & $74.1$ & $74.6$ & $43.6$ & $63.8$ & $64.3$ & $99.2$ & $66.7$ & $89.2$ & $81.1$ & $76.7$ & $81.8$ \\
\texttt{Llama-3.2-90B-Vision-Ins.} & $67.8$ & $30.7$ & $58.6$ & $87.1$ & $59.7$ & $81.6$ & $59.5$ & $94.3$ & $68.5$ & $66.0$ & $72.4$ \\
\midrule
Random & $0.0$ & $0.0$ & $0.0$ & $0.0$ & $0.0$ & $0.0$ & $0.0$ & $0.0$ & $0.0$ & $0.0$ & $0.0$ \\
\bottomrule
\end{tabular}

\end{table*}
\footnotetext{Specific model versions can be found in Appendix C.}

\section{Evaluating VLMs in \textsc{VS-Bench}}
\label{sec:results}
To comprehensively evaluate VLMs in multi-agent environments, we consider three dimensions including perception, strategic reasoning, and decision-making.
We also provide several findings that highlight the limitations of existing VLMs and motivate our analysis in the next section.

\textbf{Model setup.}
We select fifteen state-of-the-art VLMs for evaluation.
For commercial VLMs, we select six reasoning models and six chat models from OpenAI GPT~\citep{openai2025introducing,openai2025openai}, Anthropic Claude~\citep{anthropic2025claude}, Google Gemini~\citep{google2025gemini}, xAI Grok~\citep{xai2024grok}, Qwen~\citep{qwen2025qvq}, and Doubao~\citep{bytedance2025doubao}.
For open-source VLMs, we select three leading models from Llama-3.2-Vision~\citep{meta2024llama}, InternVL3~\citep{zhu2025internvl3}, and Qwen2.5-VL~\citep{bai2025qwen2}. 
We set the temperature to $1.0$ and the maximum number of output tokens to $8\mathrm{k}$ for all models.
We also set the maximum number of reasoning tokens to $16\mathrm{k}$ for reasoning models.
When encountering a cutoff for reaching maximum tokens, we dynamically extend the tokens to the model's limit.
Detailed descriptions of model setups can be found in Appendix C.

\subsection{Perception}
\label{sec:results:perception}

Perception is the ability to recognize basic visual information from raw multimodal observations.
This requires agents to accurately recognize entities, spatial relations, and numerical attributes to facilitate downstream reasoning and decision-making.
We evaluate the perception ability of VLMs by measuring their recognition accuracy of basic elements like object position, agent orientation, and game status.
We collect $400$ samples with ground-truth labels from actual game-play in each environment.
Detailed descriptions of perception evaluation can be found in Appendix D.

\textbf{Existing VLMs show competent perception ability.}
The evaluation results in Table~\ref{tab:perception} show that current VLMs achieve high performance in perception with multimodal observations. 
All models achieve at least $67.8$\% overall accuracy and the best model \texttt{o3} attains $84.9$\% overall accuracy.
Reasoning models do not show a significant advantage over chat models or open-source models. 
These results show that all VLMs demonstrate adequate perception ability for subsequent strategic reasoning and decision-making in multi-agent environment.

\begin{table*}[t]
\centering
\caption{Strategic reasoning evaluation results. For each environment, the \colorbox{first}{first}, \colorbox{second}{second}, and \colorbox{third}{third} best results are highlighted in green, while the results \colorbox{below}{below random} are highlighted in red.}
\label{tab:reasoning}
\scriptsize
\setlength{\tabcolsep}{6.5pt}
\begin{tabular}{cccccccccccc}
\toprule
\multirow{2}{*}{Models}
& \multirow{2}{*}{Overall} & \multicolumn{3}{c}{Cooperative} & \multicolumn{4}{c}{Comptitive} & \multicolumn{3}{c}{Mixed-Motive} \\
& & \textit{Hanabi} & \textit{Overcooked} & \textit{KAZ} & \textit{Board} 
& \textit{Poker}    & \textit{Pong} & \textit{MPE} & \textit{Dilemma} & \textit{Hunt} & \textit{Battle} \\
\midrule
Oracle & $100.0$ & $100.0$ & $100.0$ & $100.0$ & $100.0$ & $100.0$ & $100.0$ & $100.0$ & $100.0$ & $100.0$ & $100.0$ \\
\midrule
\texttt{o3} & $46.6$ & \cellcolor{first} $61.2$ & \cellcolor{first} $40.5$ & $30.8$ & \cellcolor{first} $29.0$ & \cellcolor{first} $67.0$ & \cellcolor{below} $25.8$ & $33.8$ & \cellcolor{second} $59.2$ & \cellcolor{first} $57.5$ & \cellcolor{first} $61.5$ \\
\texttt{claude-3-7-sonnet} & $40.4$ & \cellcolor{third} $39.0$ & $26.0$ & $29.0$ & \cellcolor{third} $24.2$ & $65.2$ & \cellcolor{first} $44.8$ & \cellcolor{first} $35.8$ & \cellcolor{third} $53.8$ & \cellcolor{third} $42.5$ & $43.2$ \\
\texttt{gemini-2.5-pro} & $39.6$ & \cellcolor{second} $51.2$ & $22.0$ & $28.8$ & \cellcolor{second} $26.8$ & $59.8$ & \cellcolor{below} $24.5$ & $35.0$ & $52.5$ & $35.8$ & \cellcolor{second} $60.0$ \\
\texttt{ui-tars-1-5} & $39.0$ & $25.5$ & $18.2$ & \cellcolor{third} $38.5$ & $23.2$ & \cellcolor{second} $65.8$ & \cellcolor{below} $30.8$ & $33.5$ & \cellcolor{first} $63.0$ & $41.5$ & \cellcolor{third} $50.0$ \\
\texttt{doubao-1-5-thinking-pro} & $36.3$ & $32.8$ & $26.2$ & $27.5$ & $19.8$ & \cellcolor{second} $65.8$ & \cellcolor{second} $44.2$ & $34.8$ & $26.5$ & \cellcolor{second} $45.2$ & $40.0$ \\
\texttt{qvq-max} & $30.8$ & $32.2$ & $19.0$ & $24.5$ & $21.8$ & $63.0$ & $37.8$ & \cellcolor{third} $35.5$ & \cellcolor{below} $25.2$ & \cellcolor{below} $21.5$ & $27.0$ \\
\midrule
\texttt{gpt-4.1} & $36.0$ & $23.0$ & \cellcolor{second} $27.0$ & $27.8$ & $22.5$ & $58.2$ & $41.5$ & \cellcolor{first} $35.8$ & $49.8$ & $36.8$ & $38.0$ \\
\texttt{qwen-vl-max} & $31.2$ & $26.5$ & $26.0$ & \cellcolor{first} $46.0$ & $19.5$ & $52.2$ & \cellcolor{below} $23.5$ & $31.0$ & $26.2$ & \cellcolor{below} $23.5$ & $37.2$ \\
\texttt{doubao-1-5-vision-pro} & $30.1$ & $15.0$ & $22.2$ & $18.5$ & $15.8$ & $58.2$ & \cellcolor{below} $31.2$ & $31.5$ & $37.2$ & $36.0$ & $34.8$ \\
\texttt{claude-3-7} w/o thinking & $29.6$ & $9.8$ & \cellcolor{below} $16.0$ & $35.8$ & $18.0$ & $57.2$ & \cellcolor{third} $43.2$ & $32.2$ & $31.0$ & \cellcolor{below} $26.0$ & $26.8$ \\
\texttt{gemini-2.5} w/o thinking & $29.0$ & $21.5$ & $19.2$ & $20.8$ & $14.8$ & $56.5$ & $34.0$ & $27.0$ & $31.8$ & $30.5$ & $33.8$ \\
\texttt{grok-2-vision} & $26.2$ & $12.8$ & $17.2$ & $25.5$ & $10.8$ & $59.5$ & \cellcolor{below} $20.8$ & $25.2$ & $30.2$ & $31.5$ & $29.0$ \\
\midrule
\texttt{Qwen2.5-VL-72B-Ins.} & $32.2$ & $26.8$ & \cellcolor{third} $26.5$ & \cellcolor{second} $39.5$ & $23.8$ & $50.8$ & \cellcolor{below} $27.0$ & $34.2$ & $30.0$ & \cellcolor{below} $27.2$ & $36.8$ \\
\texttt{InternVL3-78B} & $29.1$ & $25.2$ & $20.5$ & $24.0$ & $14.0$ & \cellcolor{below} $45.2$ & $34.8$ & $30.2$ & $37.0$ & $30.0$ & $30.2$ \\
\texttt{Llama-3.2-90B-Vision-Ins.} & $26.4$ & $20.0$ & \cellcolor{below} $16.5$ & $20.8$ & $11.8$ & $52.8$ & $36.2$ & $32.8$ & \cellcolor{below} $22.2$ & \cellcolor{below} $26.2$ & \cellcolor{below} $25.0$ \\
\midrule
Random & $23.0$ & $8.8$ & $16.7$ & $16.2$ & $4.2$ & $50.0$ & $33.3$ & $19.5$ & $25.4$ & $29.3$ & $26.5$ \\
\bottomrule
\end{tabular}

\end{table*}

\subsection{Strategic reasoning}
\label{sec:results:reasoning}

Strategic reasoning is the theory-of-mind ability to infer the hidden beliefs, desires, and intentions of other agents~\citep{rao1995bdi,kosinski2023theory}.
This requires agents to think from others' perspectives to answer: \textit{What are the next actions of other agents?}
Strategic reasoning is crucial in multi-agent environments because an agent's reward function depends not only on its own action, but also on other participants' actions.
We evaluate the strategic reasoning ability of VLMs by their prediction accuracy of other agents' next actions on a dataset for each environment.
To ensure a rigorous and thorough evaluation, we selectively collect $400$ samples and construct a predictable, diverse, and balanced dataset for each environments.
Detailed descriptions of dataset construction and strategic reasoning evaluation can be found in Appendix E.

\textbf{Existing VLMs exhibit preliminary strategic reasoning ability but are still far from oracle performance.}
The evaluation results in Fig.~\ref{fig:teaser} and Table~\ref{tab:reasoning} show that current VLMs show basic strategic reasoning ability by surpassing random agents in overall prediction accuracy, yet they still lag behind the oracle results by a noticeable margin.
Most models perform better than random guesses in at least eight of the ten games, demonstrating non-trivial theory-of-mind capability in multi-agent environments.
In general, reasoning models achieve better results than chat models and open-source models, with the best-performing model \texttt{o3} attaining an overall accuracy of $46.6$\% and ranking first in six environments.
Notably, the three leading open-source models achieve an average overall accuracy of $29.2$\%, which is comparable to commercial chat models with a $30.3$\% average overall accuracy.
However, even these most capable models attain less than $50$\% overall accuracy, and some even fail to outperform random.
This deficit is especially pronounced in \textit{Overcooked}, \textit{Atari Pong}, and \textit{Monster Hunt}, which are all adapted from video games.

\begin{table*}[t]
\centering
\caption{Decision-making evaluation results. For each environment, the \colorbox{first}{first}, \colorbox{second}{second}, and \colorbox{third}{third} best results are highlighted in green, while the results \colorbox{below}{below} or \colorbox{equal}{equal to random} are in red.}
\label{tab:decision-making}
\scriptsize
\makebox[\textwidth][c]{\setlength{\tabcolsep}{6.5pt}
\begin{tabular}{cccccccccccc}
\toprule
\multirow{2}{*}{Models}
& \multirow{2}{*}{Overall} & \multicolumn{3}{c}{Cooperative} & \multicolumn{4}{c}{Comptitive} & \multicolumn{3}{c}{Mixed-Motive} \\
& & \textit{Hanabi} & \textit{Overcooked} & \textit{KAZ} & \textit{Board} 
& \textit{Poker}    & \textit{Pong} & \textit{MPE} & \textit{Dilemma} & \textit{Hunt} & \textit{Battle} \\
\midrule
Oracle & $100.0$ & $100.0$ & $100.0$ & $100.0$ & $100.0$ & $100.0$ & $100.0$ & $100.0$ & $100.0$ & $100.0$ & $100.0$ \\
\midrule
\texttt{o3} & \cellcolor{first} $31.4$ & \cellcolor{second} $55.8$ & \cellcolor{second} $15.6$ & \cellcolor{first} $29.5$ & \cellcolor{first} $65.0$ & \cellcolor{second} $61.8$ & \cellcolor{first} $8.6$ & $37.2$ & \cellcolor{below} $-0.4$ & \cellcolor{second} $24.0$ & \cellcolor{second} $16.7$ \\
\texttt{gemini-2.5-pro} & \cellcolor{second} $23.2$ & \cellcolor{third} $32.9$ & \cellcolor{first} $17.1$ & \cellcolor{third} $8.4$ & \cellcolor{second} $55.0$ & $27.3$ & \cellcolor{second} $6.5$ & \cellcolor{second} $38.9$ & \cellcolor{below} $-9.6$ & \cellcolor{third} $21.5$ & \cellcolor{first} $33.8$ \\
\texttt{doubao-1-5-thinking-pro} & \cellcolor{third} $20.3$ & \cellcolor{first} $56.7$ & $10.1$ & \cellcolor{second} $14.1$ & $10.0$ & \cellcolor{third} $48.5$ & \cellcolor{third} $2.9$ & $38.5$ & $0.7$ & $17.2$ & $4.0$ \\
\texttt{claude-3-7-sonnet} & $18.1$ & $6.7$ & \cellcolor{third} $10.1$ & $5.3$ & \cellcolor{third} $20.0$ & \cellcolor{first} $72.9$ & \cellcolor{below} $-0.5$ & \cellcolor{first} $39.4$ & $4.6$ & $19.9$ & $2.5$ \\
\texttt{ui-tars-1-5} & $12.9$ & \cellcolor{equal} $0.0$ & $2.0$ & \cellcolor{below} $-4.7$ & $5.0$ & $36.4$ & \cellcolor{below} $-0.9$ & $19.4$ & \cellcolor{first} $33.1$ & \cellcolor{first} $24.9$ & \cellcolor{third} $13.6$ \\
\texttt{qvq-max} & $3.6$ & \cellcolor{equal} $0.0$ & $2.0$ & \cellcolor{below} $-1.7$ & $5.0$ & \cellcolor{below} $-8.7$ & $0.4$ & \cellcolor{third} $38.6$ & \cellcolor{equal} $0.0$ & $0.7$ & \cellcolor{below} $-0.5$ \\
\midrule
\texttt{gpt-4.1} & $4.8$ & \cellcolor{equal} $0.0$ & \cellcolor{below} $-0.5$ & \cellcolor{below} $-5.3$ & \cellcolor{equal} $0.0$ & \cellcolor{below} $-7.1$ & $0.2$ & $31.5$ & \cellcolor{second} $17.8$ & $11.2$ & $0.5$ \\
\texttt{gemini-2.5} w/o thinking & $4.8$ & \cellcolor{equal} $0.0$ & $2.0$ & \cellcolor{equal} $0.0$ & \cellcolor{equal} $0.0$ & $18.2$ & $1.0$ & $23.8$ & \cellcolor{below} $-0.7$ & $0.7$ & $2.5$ \\
\texttt{qwen-vl-max} & $0.2$ & $1.2$ & \cellcolor{below} $-0.5$ & \cellcolor{below} $-5.3$ & \cellcolor{equal} $0.0$ & \cellcolor{below} $-20.5$ & \cellcolor{below} $-0.3$ & $14.4$ & \cellcolor{below} $-0.4$ & $13.2$ & \cellcolor{below} $-0.5$ \\
\texttt{claude-3-7} w/o thinking & \cellcolor{below} $-0.6$ & \cellcolor{equal} $0.0$ & $2.0$ & $3.5$ & $5.0$ & \cellcolor{below} $-23.6$ & \cellcolor{below} $-0.9$ & $5.6$ & $1.4$ & $0.2$ & $1.0$ \\
\texttt{doubao-1-5-vision-pro} & \cellcolor{below} $-0.9$ & \cellcolor{equal} $0.0$ & \cellcolor{below} $-0.5$ & \cellcolor{below} $-5.3$ & \cellcolor{equal} $0.0$ & \cellcolor{below} $-40.2$ & \cellcolor{below} $-0.9$ & $32.9$ & \cellcolor{below} $-2.1$ & $7.8$ & \cellcolor{below} $-0.5$ \\
\texttt{grok-2-vision} & \cellcolor{below} $-6.7$ & \cellcolor{equal} $0.0$ & $1.5$ & \cellcolor{equal} $0.0$ & \cellcolor{equal} $0.0$ & \cellcolor{below} $-11.8$ & \cellcolor{below} $-0.1$ & \cellcolor{below} $-58.2$ & $1.1$ & \cellcolor{below} $-0.4$ & $0.5$ \\
\midrule
\texttt{Qwen2.5-VL-72B-Ins.} & $3.0$ & $0.8$ & \cellcolor{below} $-0.5$ & \cellcolor{below} $-5.3$ & \cellcolor{equal} $0.0$ & \cellcolor{below} $-3.1$ & \cellcolor{below} $-0.8$ & $19.6$ & \cellcolor{equal} $0.0$ & $19.6$ & \cellcolor{below} $-0.5$ \\
\texttt{InternVL3-78B} & $1.2$ & \cellcolor{equal} $0.0$ & \cellcolor{equal} $0.0$ & \cellcolor{below} $-3.5$ & \cellcolor{equal} $0.0$ & $4.0$ & \cellcolor{below} $-0.9$ & $5.9$ & \cellcolor{third} $5.0$ & \cellcolor{below} $-0.2$ & $1.5$ \\
\texttt{Llama-3.2-90B-Vision-Ins.} & \cellcolor{below} $-6.3$ & \cellcolor{equal} $0.0$ & $1.5$ & \cellcolor{equal} $0.0$ & \cellcolor{equal} $0.0$ & \cellcolor{below} $-39.4$ & \cellcolor{below} $-0.9$ & \cellcolor{below} $-29.6$ & $0.4$ & $3.6$ & $1.0$ \\
\midrule
Random & $0.0$ & $0.0$ & $0.0$ & $0.0$ & $0.0$ & $0.0$ & $0.0$ & $0.0$ & $0.0$ & $0.0$ & $0.0$ \\
\bottomrule
\end{tabular}

}
\end{table*}

\subsection{Decision-making}
\label{sec:results:decision}

Decision-making is the ability to optimize for one's long-term objective under uncertainty~\citep{edwards1954theory}.
This requires agents to prioritize future accumulated returns over immediate gains, adapt to non-stationary dynamics with evolving agents, and balance cooperation and competition for favorable equilibria. 
We evaluate the decision-making ability of VLMs by their normalized episode returns through self-play or interactions with conventional agents in each environment.
We also evaluate random agents and oracle agents to normalize the results so that the normalized return for random agents is $0$ and the normalized return for oracle agents is $100$.
Detailed descriptions of decision-making evaluation can be found in Appendix F.

\textbf{Existing VLMs struggle with decision-making in multi-agent environments.}
The evaluation results in Fig.~\ref{fig:teaser} and Table~\ref{tab:decision-making} show that current VLMs have limited decision-making ability in multi-agent environments, highlighting a significant gap between existing models and oracle performance.
As illustrated by the large swaths of red cells, four out of fifteen models' overall performance is even worse than random agents, indicating their incompetence to optimize long-term return under non-stationary, interdependent multi-agent dynamics. 
Although reasoning models achieve relatively better results than chat models and open-source models, even the best-performing model \texttt{o3} only attains an overall normalized return of $31.4$\%, which is far behind the oracle agent.
Surprisingly, we observe that open-source models can achieve comparable results to reasoning models in some mixed-motive games like \texttt{InternVL3-78B} in \textit{Coin Dilemma} and \texttt{Qwen2.5-VL-72B-Ins.} in \textit{Monster Hunt}.
We also observe that the failures to outperform random agents are concentrated on video games like \textit{Overcooked}, \textit{KAZ}, \textit{Atari Pong}, and \textit{Coin Dilemma}, which underscores the coupled difficulty of multimodal perception and strategic decision-making in \textsc{VS-Bench}.

\section{Analysis}
\label{sec:analysis}
\begin{figure*}[t]
\centering
\includegraphics[width=\linewidth,trim={0 10 0 10},clip]{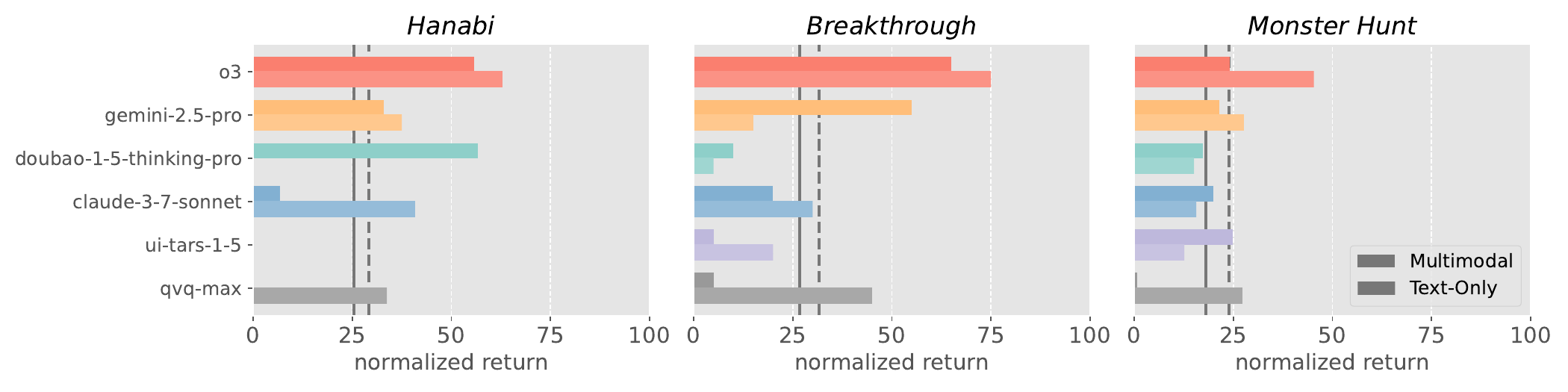}
\caption{Comparison of reasoning VLMs on decision-making with multimodal and text-only observations. The solid and dashed vertical lines represent the average results of two settings.}
\label{fig:multimodal}
\end{figure*}

To investigate the key factors that influence the performance of VLMs in multi-agent environments,
we perform in-depth analyses on multimodal observations, test-time scaling, and social behaviors. 
We also compare VLMs with human performance and provide failure cases for future development.
More experiment results can be found in Appendix G.

\subsection{Multimodal observations}
\label{sec:analysis:multimodal}

Although VLMs achieve good performance in multimodal perception, the evaluations on reasoning and decision-making show that environments with rich visual observations are especially challenging for VLM agents.
To investigate the influence of multimodal observations, we further evaluate VLMs with text-only observations that replace images with corresponding text descriptions. 
We select a board game, a card game, and a video game.
The decision-making results of reasoning models in Fig.~\ref{fig:multimodal} show that VLMs achieve slightly better results with text-only observations, but are still far from oracle results. 
This indicates that the weak decision-making performance arises from the coupled challenges of multimodal observation and strategic interactions in multi-agent environments.

\subsection{Test-time scaling}
\label{sec:analysis:scaling}

\begin{figure*}[t]
\centering
\includegraphics[width=\linewidth,trim={0 10 0 10},clip]{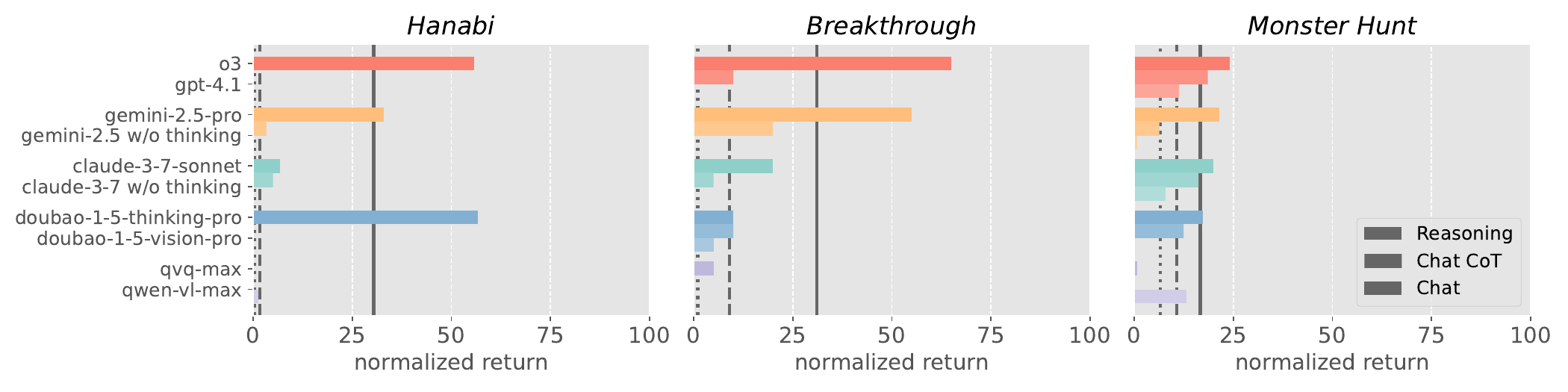}
\caption{Comparison of reasoning VLMs and chat VLMs on decision-making with IO and CoT prompting. The solid, dashed, and dotted vertical lines represent the average results of three settings.}
\label{fig:scaling}
\end{figure*}

We observe in the evaluation results that reasoning models generally achieve better performance than chat models. 
We further investigate the test-time scaling of VLMs in multi-agent environments by using Chain-of-Thought (CoT)~\citep{wei2022chain} prompting for chat models and comparing their performance with reasoning models and chat models with simple IO prompting.
The evaluation results in Fig.~\ref{fig:scaling} show that CoT prompting significantly improves chat models' performance in all three environments, while reasoning models still achieve the best results.
This demonstrates that test-time scaling like reasoning and CoT prompting can substantially improve VLMs' performance.

\begin{figure*}[t]
\centering
\includegraphics[width=0.9\linewidth,trim=0 40 0 0,clip]{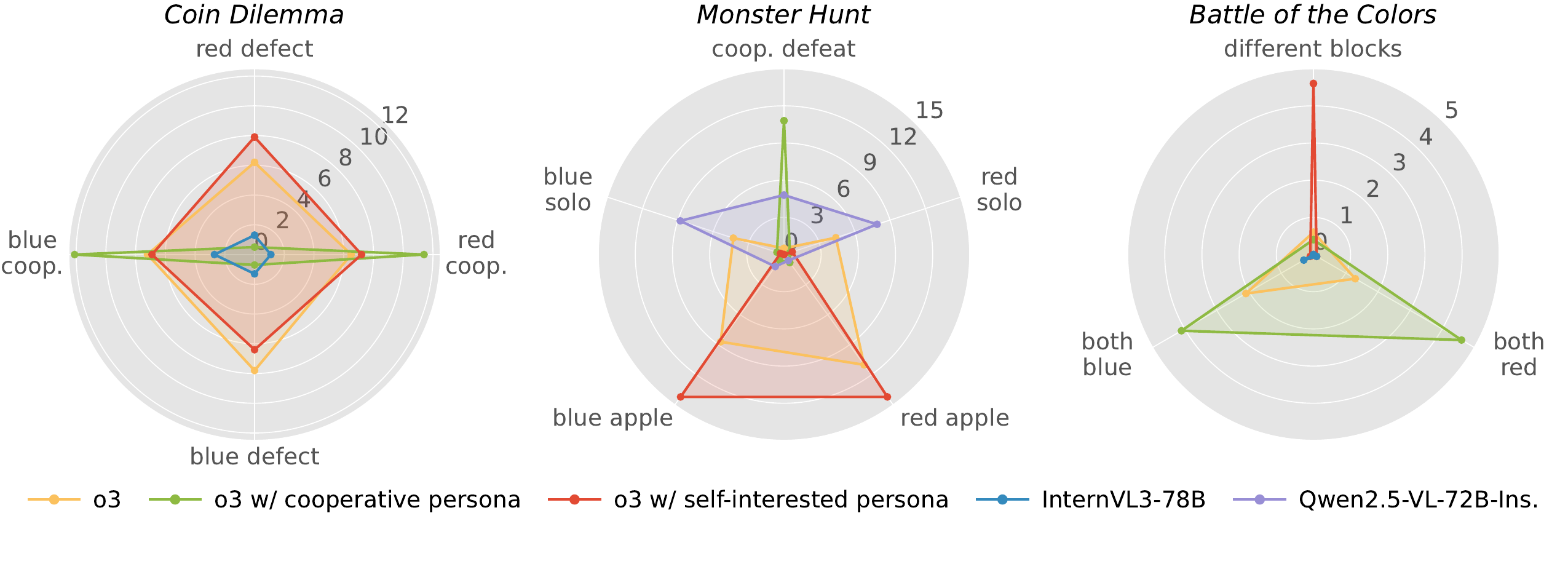}
\caption{Social behaviors of \texttt{o3} with different personas and the best-performing open-source model in each social dilemma game. Dimensions are agents' behaviors described in Appendix J.}
\label{fig:social}
\end{figure*}

\subsection{Social behaviors and personas}
\label{sec:analysis:behaviors}

Another interesting observation in the evaluation results is that open-source models can achieve comparable results to reasoning models in some mixed-motive games. 
We investigate this by visualizing the behaviors of \texttt{o3} and the best-performing open-source models in each social dilemma game.
As shown in Fig.~\ref{fig:social}, in \textit{Coin Dilemma}, \texttt{o3} are better at collecting coins but are also more self-interested, while \texttt{InterVL3-78B} is more inclined to cooperation that leads to a win-win situation.
Similar behaviors can be found in \textit{Monster Hunt}, where the two reasoning models tend to safely eat apples alone, while \texttt{Qwen2.5-VL-72B-Ins.} prefers taking the risk to cooperate with the other agent and defeat the monster together.
We also explicitly prompt \texttt{o3} with self-interested and cooperative personas, and find that different personas significantly change the behaviors and performance of VLMs in social dilemmas.

\subsection{Human studies}
\label{sec:analysis:human}

\begin{figure}
    \centering
    \includegraphics[width=0.4\textwidth,trim=0 10 0 10,clip]{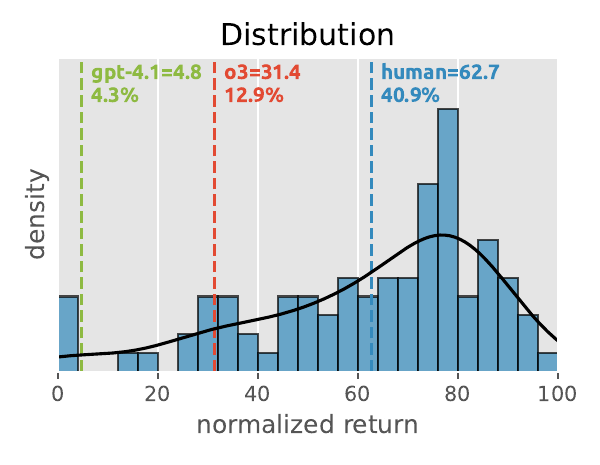}
    \caption{Decision-making results of human participants.}
    \label{fig:human}
\end{figure}

We further conduct experiments with humans in all ten games to better understand the current capability of VLMs in multi-agent environments.
To ensure a fair comparison, we let human participants receive the same observations and play games for the same number of times as the VLMs. 
We collect the normalized episode returns of $26$ human participants and visualize their distribution.
As shown in Fig.~\ref{fig:human}, human participants achieve an average overall normalized return of $62.7$.
For comparison, \texttt{o3} attains an overall normalized return of $31.4$, which surpasses $12.9$\% of human results; \texttt{gpt-4.1} attains an overall normalized return of $4.4$, which surpasses only $4.3$\% of human results.

\subsection{Failure cases}
\label{sec:analysis:failure}

To investigate why VLMs underperform in multi-agent environments, we conduct a qualitative analysis of their failure cases.
In strategic reasoning, two common failure cases are ignoring history and private information. 
For example, in \textit{Hanabi}, players' cards are observable to other agents but not to themselves. 
VLMs often overlook this information asymmetry and incorrectly use their private information to predict the next actions of others.
In decision-making, a common failure case is focusing excessively on one's own actions while ignoring those of others.
For example, in \textit{Breakthrough}, VLMs tend to persistently advance their own pieces and fail to identify vulnerabilities that result in losing the match.
More failure cases can be found in Appendix H.

\section{Related work}
\label{sec:related}
\subsection{Multi-agent environments and benchmarks}

Early work on multi-agent reasoning and decision-making is grounded in classical game theory~\citep{fudenberg1991game,von2007theory}, which models rational interactions among self-interested players and introduces canonical environments like board games~\citep{shannon1950xxii,tesauro1994td}, card games~\citep{kuhn1950simplified,southey2012bayes}, and social dilemmas~\citep{luce2012games,rapoport1965prisoner,rousseau1985discourse}.
Building on these foundations, breakthroughs in multi-agent reinforcement learning (MARL)~\citep{silver2016mastering,brown2019superhuman} have expanded the field toward complex, high-dimensional environments like video games.
These environments cover a diverse range of multi-agent dynamics like cooperation~\citep{carroll2019utility,samvelyan2019starcraft,bard2020hanabi}, competition~\citep{moravvcik2017deepstack,vinyals2019grandmaster}, and mixed-motive interactions~\citep{lowe2017multi,berner2019dota}.
Despite their impressive achievements, agents in these environments are typically specialized for a single task and lack general-purpose abilities for strategic reasoning and decision-making across different domains. 

Recent advancements in Large Language Models (LLMs)~\citep{ouyang2022training,touvron2023llama,guo2025deepseek} have catalyzed a paradigm shift toward generalist agents that can perceive and act in various environments without task-specific training.
A growing body of text-based benchmarks has been proposed to evaluate different facets of LLM agents in multi-agent environments.
LLM-Coordination~\citep{agashe2023llm} analyzes LLM agents' ability in pure cooperative tasks.
GT-Bench~\citep{duan2024gtbench} and GAMA-Bench~\citep{huang2024far} consider non-cooperative games and evaluate LLMs through the lens of game theory. 
Trust Games~\citep{xie2024can} studies the trust behaviors of LLM agents in their interactions.
MAgIC~\citep{xu2023magic} and LLMArena~\citep{chen2024llmarena} include both cooperative and competitive games for a comprehensive evaluation.
However, these works~\cite{zha2026aircopbench} mainly focus on text-only environments, which are different from real-world decision-making that integrates both visual 
and textual observations.
Our work fills this gap by introducing ten vision-grounded games to evaluate VLMs in multi-agent environments.

\subsection{VLM agent benchmarks}

The rapid evolution of Vision Language Models (VLMs)~\citep{bai2023qwen,liu2023visual} has driven evaluation beyond static tasks like image captioning~\citep{chen2015microsoft} and visual reasoning~\citep{antol2015vqa,yue2024mmmu} toward interactive agent environments.
Existing benchmarks can be broadly categorized into four domains: coding, GUI interaction, game environments, and embodied control.
Coding benchmarks~\citep{chan2024mle,li2024mmcode,yang2024swe} consider software engineering and machine learning engineering with both image and text input.
GUI benchmarks evaluate VLMs on graphic interface operations like web browsing~\citep{he2024webvoyager,koh2024visualwebarena,zheng2024gpt}, computer use~\citep{xie2024osworld,bonatti2024windows}, and phone use~\citep{li2024effects,rawles2024androidworld}.
Game benchmarks~\citep{zhang2023creative,liu2024visualagentbench,wang2025large} offer dynamic virtual environments with structured rewards to assess VLMs. 
Embodied benchmarks~\citep{guruprasad2024benchmarking,team2025gemini,yang2025embodiedbench,zha2025enable} evaluate VLMs in vision-driven robotics and physical world interactions.
Nevertheless, these benchmarks predominantly concentrate on single-agent tasks, which overlook the distinctive challenges of multi-agent environments including non-stationary dynamics, interdependent decision-making, and equilibrium selection.
Our work bridges this gap by evaluating VLMs' strategic ability in multi-agent games.

\section{Conclusion}
\label{sec:conclusion}
In this work, we present \textsc{VS-Bench}, a multimodal benchmark for evaluating VLMs' strategic abilities in multi-agent environments.
By introducing ten vision-grounded environments and three evaluation dimensions, including perception, strategic reasoning, and decision-making, we establish a comprehensive multi-agent evaluation of VLMs.
Extensive experiments on fifteen leading VLMs reveal a significant gap between existing models and optimal performance on reasoning and decision-making, highlighting their limitations for future development.
We further provide detailed analyses on multimodal observations, test-time scaling, social behaviors, human studies, and failure cases of VLM agents.
By releasing \textsc{VS-Bench} as an open platform, we seek to spur research on strategic multimodal agents that excel in vision-grounded multi-agent environments.

\section*{Acknowledgement}
This work is supported by the National Natural Science Foundation of China
(No.62406159, 62325405), Ant Group, Beijing National Research Center for Information Science, Technology (BNRist), and Beijing Innovation Center for Future Chips.

{
    \small
    \bibliographystyle{ieeenat_fullname}
    \bibliography{main}
}

\newpage
\appendix
\onecolumn

\section{Ethics statement}
\textbf{Positive research and societal value.}
VS-Bench targets a core capability that future AI systems will increasingly need: making strategic, multi-step decisions while perceiving the world through vision and language. By standardising how this ability is measured, the benchmark can accelerate reproducible research on safer, more reliable multimodal agents. Concretely, it enables (1) principled comparisons across models, (2) diagnostic analyses that pinpoint specific failure modes such as myopic play or poor opponent modelling, and (3) a shared testbed for developing methods that foster cooperation, fairness, or robustness in complex interactive settings. Beyond academic progress, stronger decision-making agents could benefit applications like assistive household robotics, disaster-response swarms, automated traffic control, and large-scale scientific simulations where coordination and strategic planning are essential.

\textbf{Risk of misuse and dual-use considerations.}
At the same time, more capable agents that reason strategically can be repurposed for adversarial or deceptive objectives — for example, collusive price-setting, automated disinformation campaigns, or the coordination of autonomous weapons systems. VS-Bench lowers the barrier to evaluating such capabilities, potentially making it easier to select or fine-tune models for harmful ends. To mitigate this, we will (1) release only simulated environments that do not directly embody real-world attack surfaces, (2) distribute the benchmark and evaluation code under licenses that forbid the use of our assets in weaponised or surveillance applications, and (3) encourage follow-up work on safety safeguards (e.g., opponent-aware alignment checks) by providing explicit hooks for auditing model rationales and behaviors. During review, we intentionally omit any URLs or repository links to preserve anonymity.

\textbf{Human subjects protection.}
To minimize potential risks to human participants, we conduct the human studies with departmental approval and enroll individuals only after obtaining informed consent. The consent form explicitly stated that the games involve cooperative, competitive, and mixed-motive interactions between participants. Participants are paid for taking part in this study and can choose to withdraw from the study at any time without consequences.

\textbf{Privacy and data ethics.}
All VS-Bench environments are synthetic with no personally identifiable information or copyrighted third-party imagery that is not permissively licensed. Replays, logs, and intermediate states are derived entirely from simulation and will be released under an open license to avoid common privacy pitfalls in dataset creation and facilitate unrestricted academic use. During review, we intentionally omit any URLs or repository links to preserve anonymity.

\section{Limitations}
\label{app:limitations}

\textbf{Evaluation metric.}
For decision-making evaluation, we measure the episode return in self-play and interaction with conventional agents. 
Incorporating results against a population of diverse opponents provides a more comprehensive assessment of the generalization and adaptability of VLM agents. 

\textbf{Number of agents.}
Many real-world scenarios involve more than two participants. 
Although our current environments mainly focus on two-player games, some of them naturally support more than two agents. 
At present, we have conducted preliminary decision-making experiments in three-player settings for both \textit{Hanabi} and the \textit{Coin Dilemma} in Appendix~\ref{app:results:multi}. 
We believe that extending our benchmark to include games with even more agents is a promising direction for future work.

\section{Models configuration details}
\label{app:models}
\begin{table}[ht]
\centering
\scriptsize
\caption{Model configurations used in the evaluation.}
\makebox[\textwidth][c]{\setlength{\tabcolsep}{8pt}
\begin{tabular}{cccccc}
\toprule
\textbf{Models} & \textbf{Version} & \textbf{Evaluated} & \textbf{Reasoning} & \textbf{Multimodal} & \textbf{Open-Source} \\
\midrule
\texttt{o3}                          & \texttt{o3-2025-04-16}               & \cmark & \cmark & \cmark & \xmark \\
\texttt{gemini-2.5-pro}             & \texttt{gemini-2.5-pro-preview-03-25}   & \cmark & \cmark & \cmark & \xmark \\
\texttt{claude-3-7-sonnet}            & \texttt{claude-3-7-sonnet-20250219}       & \cmark & \cmark & \cmark & \xmark \\
\texttt{doubao-1-5-thinking-pro}      & \texttt{doubao-1-5-thinking-pro-m-250428} & \cmark & \cmark & \cmark & \xmark \\
\texttt{ui-tars-1-5}                   & \texttt{doubao-1-5-ui-tars-250428} & \cmark & \cmark & \cmark & \xmark \\
\texttt{qvq-max}                      & \texttt{qvq-max-2025-03-25}               & \cmark & \cmark & \cmark & \xmark \\
\midrule
\texttt{gpt-4.1}                      & \texttt{gpt-4.1-2025-04-14}               & \cmark & \xmark & \cmark & \xmark \\
\texttt{gemini-2.5} w/o thinking      & \texttt{gemini-2.5-flash-preview-04-17}   & \cmark & \xmark & \cmark & \xmark \\
\texttt{claude-3-7} w/o thinking      & \texttt{claude-3-7-sonnet-20250219}       & \cmark & \xmark & \cmark & \xmark \\
\texttt{grok-2-vision}                & \texttt{grok-2-vision-1212}               & \cmark & \xmark & \cmark & \xmark \\
\texttt{doubao-1-5-vision-pro}        & \texttt{doubao-1-5-vision-pro-250328}        & \cmark & \xmark & \cmark & \xmark \\
\texttt{qwen-vl-max}                  & \texttt{qwen-vl-max-2025-04-08}           & \cmark & \xmark & \cmark & \xmark \\
\midrule

\texttt{Llama-3.2-90B--Ins.} & \texttt{Llama-3.2-90B-Vision-Instruct}
& 
\cmark & \xmark & \cmark & \cmark \\

\texttt{InternVL3-78B} & 
\texttt{InternVL3-78B} & 
\cmark & \xmark & \cmark & \cmark \\

\texttt{Qwen2.5-VL-72B-Ins.} & 
\texttt{Qwen2.5-VL-72B-Instruct} & 
\cmark & \xmark & \cmark & \cmark \\

\midrule
\texttt{claude-4-opus}               & N/A & \xmark & \cmark & \cmark & \xmark \\
\texttt{grok-4}                       & N/A & \xmark & \cmark & \xmark & \xmark \\
\texttt{grok-4-fast}                       & N/A & \xmark & \cmark & \cmark & \xmark \\
\texttt{deepseek-r1}                  & N/A & \xmark & \cmark & \xmark & \cmark \\
\texttt{deepseek-v3}                  & N/A & \xmark & \xmark & \xmark & \cmark \\
\bottomrule
\end{tabular}
\setlength{\tabcolsep}{6pt}
}
\label{tab:model-config}
\end{table}

In this work, we evaluated the state-of-the-art VLMs released before September~1,~2025.
All models in our experiments are listed in Table~\ref{tab:model-config}.
For each model, the table specifies the exact version, whether it is a reasoning model, supports multimodal inputs, and is open‑source. 
Because our environments provide both visual and linguistic observations, we do not evaluate models lacking multimodal input support, such as \texttt{grok-4}, \texttt{deepseek-r1}, \texttt{deepseek-v3}.
At present, \texttt{claude-4-opus} is not evaluated due to its high cost (exceeding~\$1000), and \texttt{grok-4-fast} is excluded because it was released on September~20,~2025, which is after our specified model release deadline.

\section{Perception evaluation details}
\label{app:perception}
In this section, we describe the construction of the perception datasets and the evaluation protocols used to assess the perception abilities of VLMs across different environments. For each game, we construct a dataset of 400 samples.

\subsection{Hanabi}
\label{app:perception:hanabi}

The samples are uniformly drawn from the trajectories of \texttt{o3} and \texttt{gemini-2.5-pro}, the most capable reasoning models that explore a more diverse range of game states.

\begin{figure}[ht]
  \centering
  \begin{minipage}[b]{0.44\textwidth}
    \centering
    \vspace{0pt}
    \includegraphics[width=\linewidth]{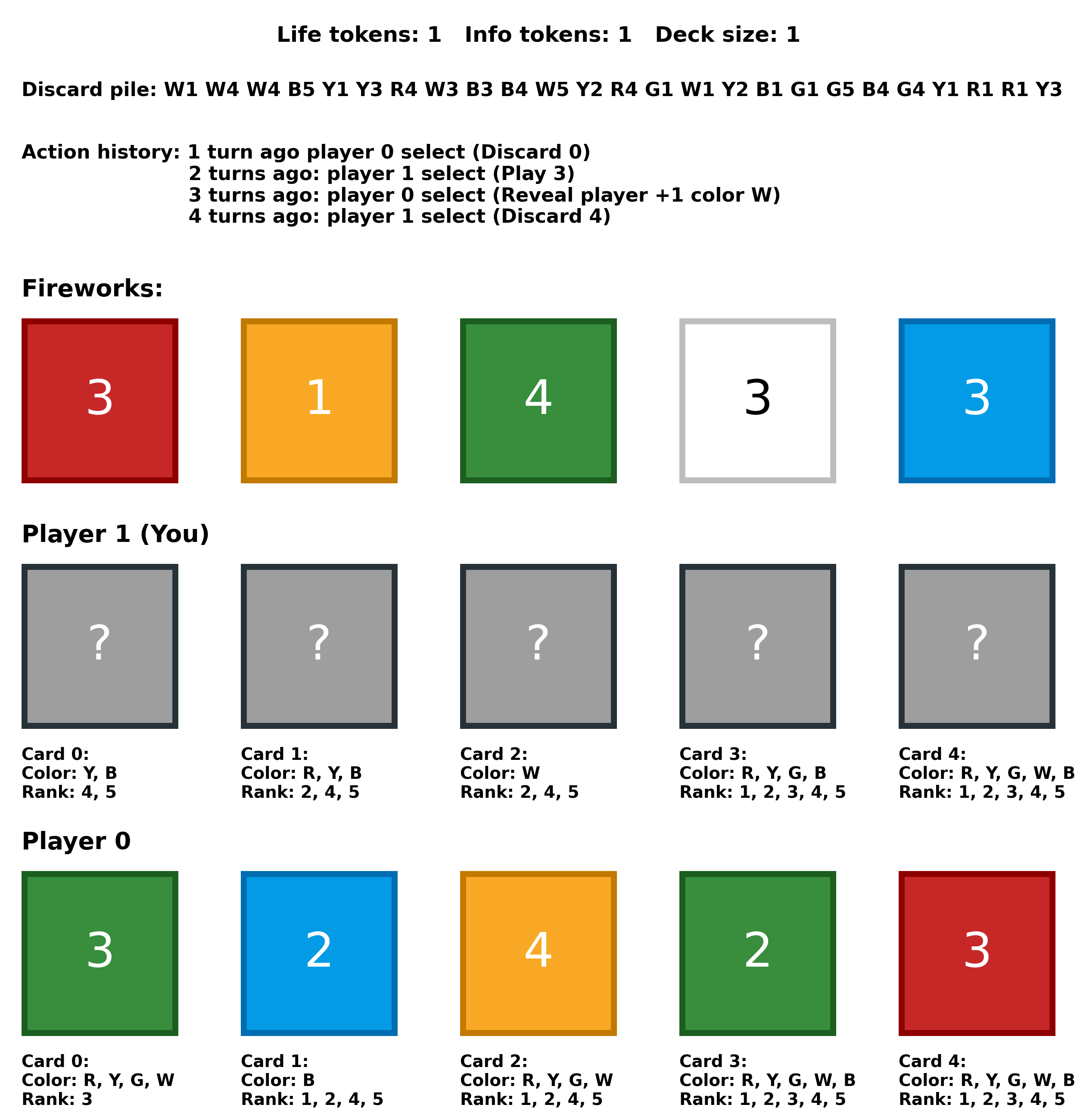}
    \caption{\textit{Hanabi}.}
    \label{fig:app:hanabi}
  \end{minipage}\hfill
  \begin{minipage}[b]{0.5\textwidth}
    \centering
    \raisebox{10mm}{\includegraphics[width=\linewidth]{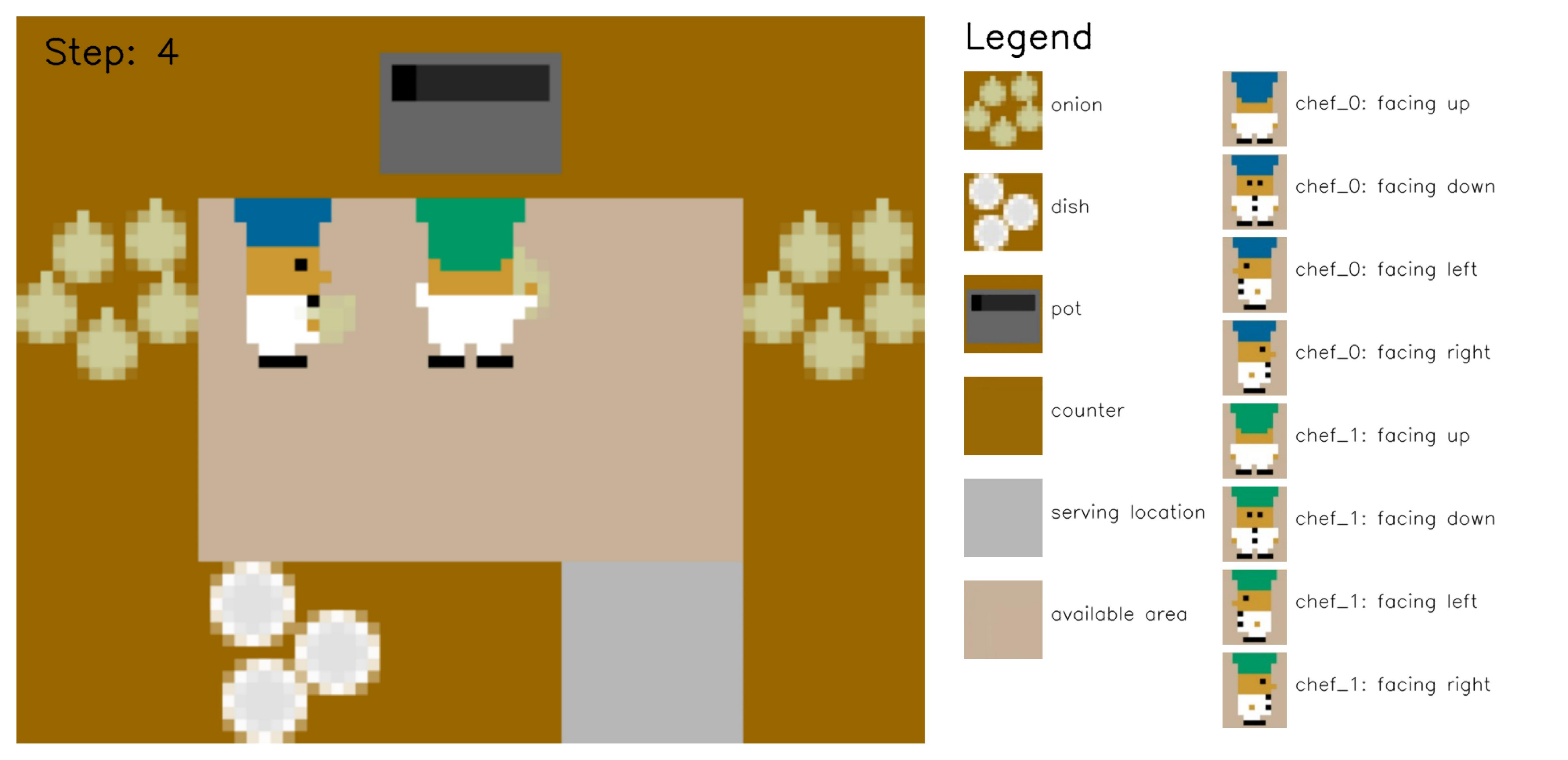}}
    \caption{\textit{Overcooked}.}
    \label{fig:app:overcooked}
  \end{minipage}
\end{figure}

Given the rich information present in the \textit{Hanabi} environment, we select four key elements as evaluation targets: \textit{life tokens}, \textit{info tokens}, \textit{fireworks}, and \textit{cardinfo}. For each sample, a screenshot of the game state (e.g., Fig.~\ref{fig:app:hanabi}) is provided, and the model is prompted to infer the values of the specified elements.
Specifically, \textit{fireworks} consists of five color-based stacks, while \textit{cardinfo} represents ten cards (five per player), each requiring the model to output 11 units of information: card value (known for the opponent's hand, but "?" for the player's own hand), and the color/number knowledge that each player currently has about the cards. In total, each sample requires predicting 117 units of information: 1 for \textit{life tokens}, 1 for \textit{info tokens}, 5 for \textit{fireworks}, and 110 (11 × 10) for \textit{cardinfo}. Each unit is equally weighted, and the final accuracy is normalized to the range \([0, 1]\).

\subsection{Overcooked}
\label{app:perception:overcooked}

This dataset is derived from the human experiment data provided by Overcooked-AI~\cite{carroll2019utility}, which contains gameplay trajectories recorded from multiple human participants.

For the \textit{Overcooked} environment, each state is represented as a $4 \times 5$ grid. For each sample, a screenshot corresponds to this $4$-row by $5$-column layout (e.g., Fig.~\ref{fig:app:overcooked}), and the model is prompted to infer the content of each grid cell, producing a $4 \times 5$ matrix as output.

Each grid cell may contain objects such as tables, pots, onions, dishes, service desks, or empty areas. To simplify model predictions, we map each possible object to a symbolic representation (e.g., \texttt{P} stands for pot). The model’s score is computed as the number of correctly predicted grid cells out of 20, normalized to the range \([0, 1]\).

\begin{figure}[t]
  \centering
\begin{minipage}[b]{0.45\textwidth}
    \centering
    \includegraphics[width=\textwidth]{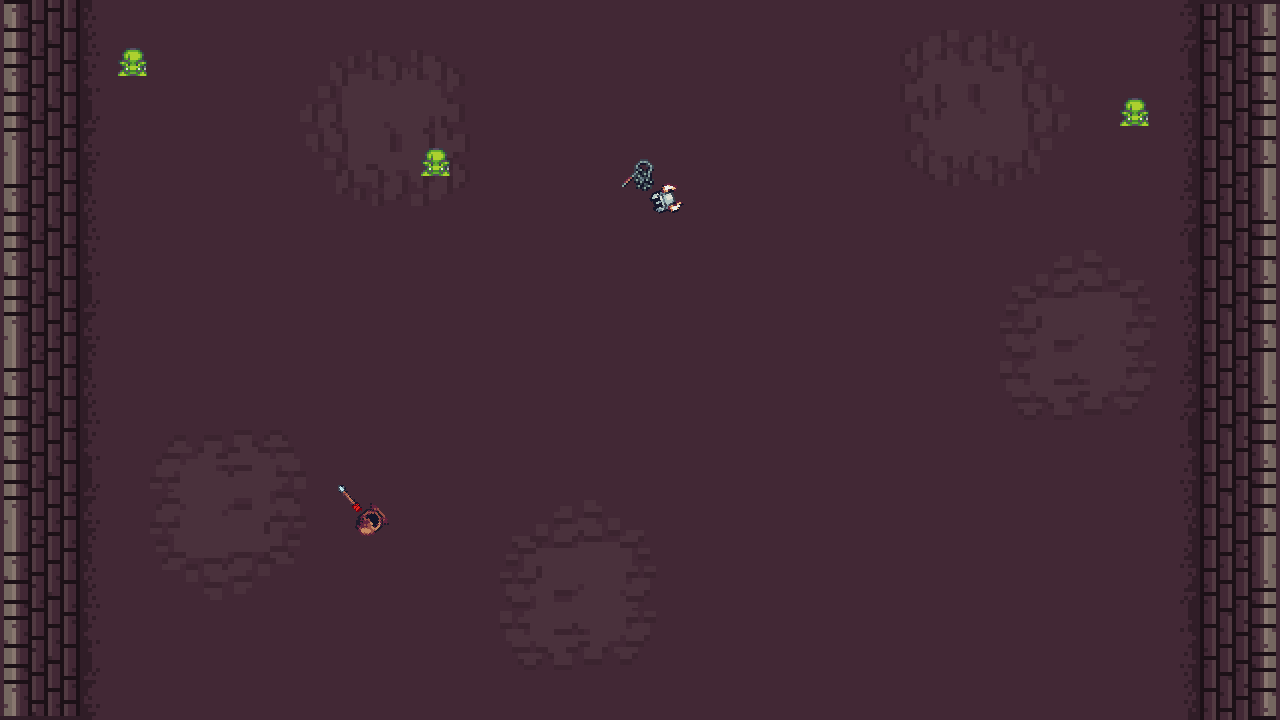}
    \caption{Origin \textit{KAZ}.}
    \label{fig:app:kaz-origin}
\end{minipage}\hfill
\begin{minipage}[b]{0.45\textwidth}
\centering
    \includegraphics[width=\textwidth]{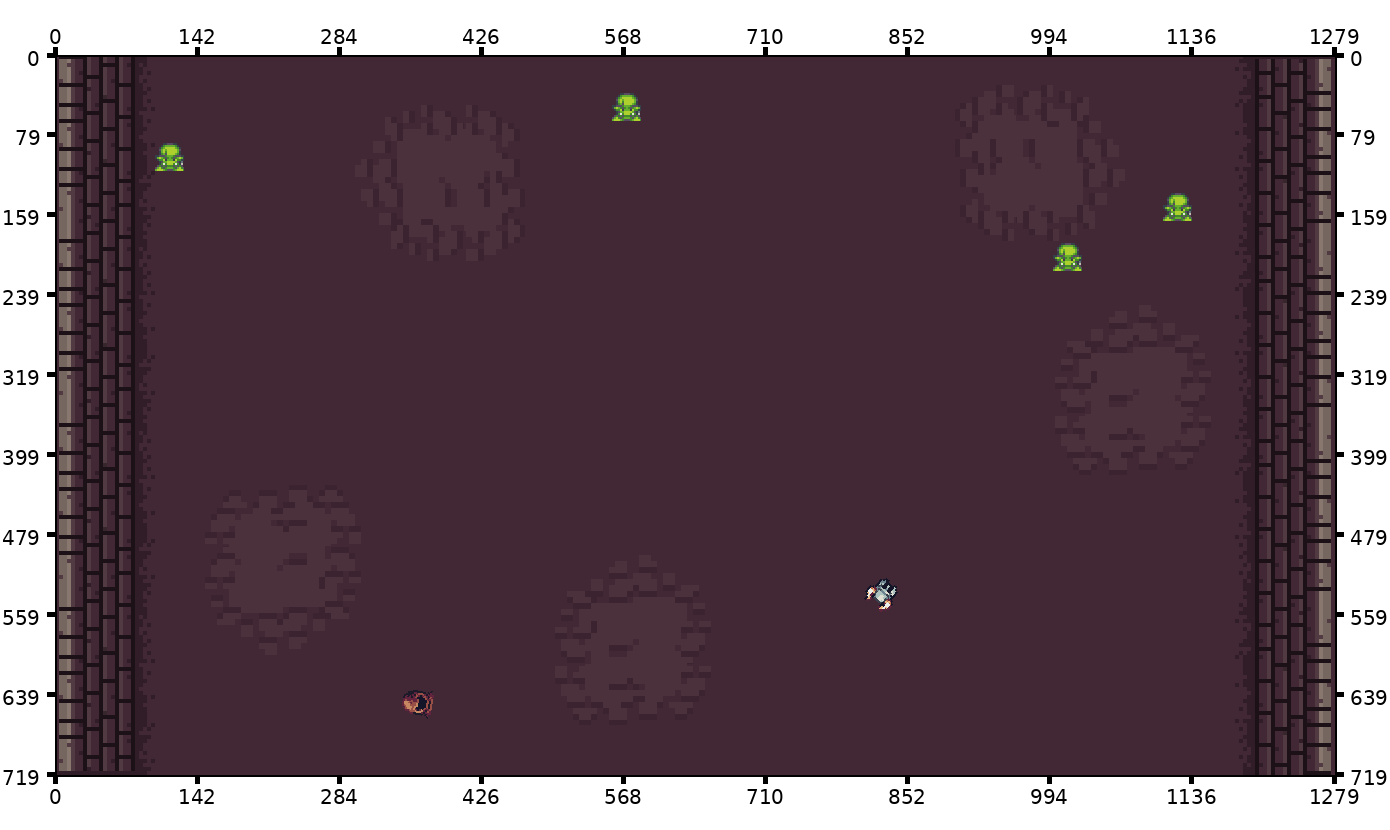}
    \caption{\textit{KAZ} with coordinate axes.}
    \label{fig:app:kaz-axes}
\end{minipage}
\end{figure}

\subsection{Knights Archers Zombies}
\label{app:perception:kaz}

The samples are uniformly drawn from trajectories generated by a \texttt{random} agent. This choice is motivated by the fact that the game may involve a large number of zombies. To simplify perception for VLMs, we limit the maximum number of zombies to five. Under this constraint, we observe that the \texttt{random} agent explores a broader distribution of game states compared to other VLMs, thereby providing a more comprehensive benchmark for evaluating perception ability.

The original \textit{KAZ} environment is continuous (e.g., Fig.~\ref{fig:app:kaz-origin}). To facilitate visual recognition by VLMs, we implement an additional wrapper that overlays coordinate axes and scales on the environment. For each sample, a screenshot of the game state with coordinate axes is provided (e.g., Fig.~\ref{fig:app:kaz-axes}), and the model is prompted to infer the following elements: the coordinates of the archer and the knight, the number of zombies present (up to five), and the positions of all detected zombies. The four elements are weighted as follows: $0.2$, $0.2$, $0.2$, and $0.4$, respectively.

For zombie count, exact matching is required to receive full credit. For coordinate predictions, full credit is awarded if the predicted position lies within a predefined radius centered on the ground-truth location. Beyond this radius, the score decays proportionally to the distance from the correct position. In our setting, the radius is fixed at $100$.

\subsection{Breakthrough}
\label{app:perception:breakthrough}

The samples are uniformly drawn from trajectories generated by a \texttt{random} agent. This choice is motivated by the observation that it can explore board states that other VLMs are unlikely to reach, thereby providing a more comprehensive benchmark for evaluating perception ability.

\begin{figure}[h]
  \centering
\begin{minipage}[b]{0.45\textwidth}
    \centering
    \includegraphics[width=\textwidth]{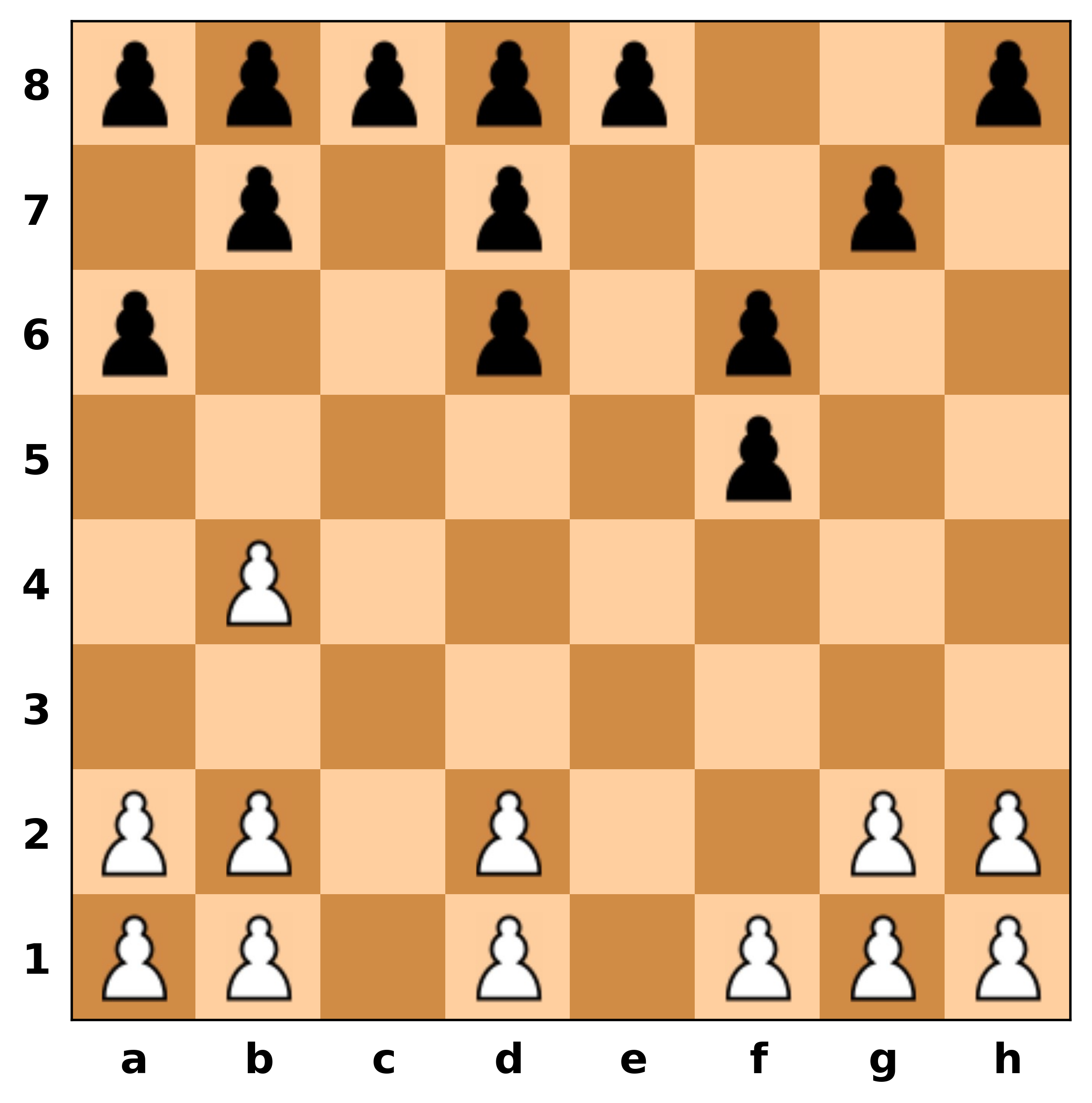}
    \caption{\textit{Breakthrough}.}
    \label{fig:app:breakthrough}
\end{minipage}\hfill
\begin{minipage}[b]{0.45\textwidth}
    \centering
    \includegraphics[width=\textwidth]{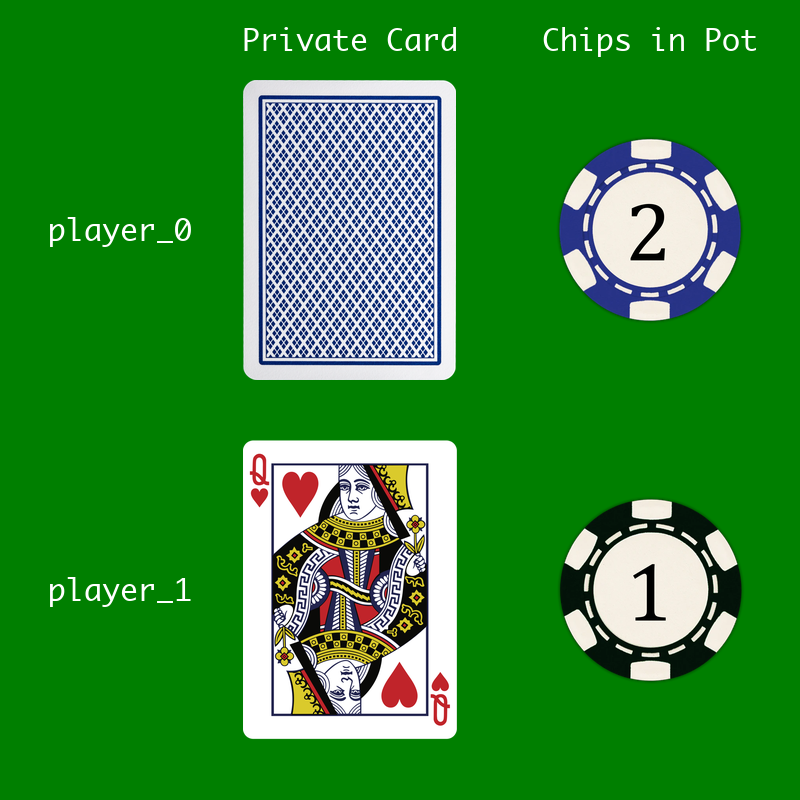}
    \caption{\textit{Kuhn Poker}.}
    \label{fig:app:kuhn_cards}
\end{minipage}

\end{figure}

In the \textit{Breakthrough} environment, each state is represented as an $8 \times 8$ grid. Accordingly, a screenshot of the game state (e.g., Fig.~\ref{fig:app:breakthrough}) is provided for each sample, and the model is prompted to infer the content of each grid cell, ultimately producing an $8 \times 8$ matrix as output.

Each grid cell may contain one of three possible objects: \texttt{b} for a black piece, \texttt{w} for a white piece, and \texttt{.} for an empty cell. The model’s score is computed as the number of correctly predicted grid cells out of 64, normalized to the range \([0, 1]\).

\subsection{Kuhn poker}
\label{app:perception:poker}

The samples are uniformly drawn from trajectories generated by \texttt{random} agent, since the overall state space is relatively small and a random agent is able to traverse it more effectively.

In \textit{Kuhn poker}, the environment contains only four variables: the card values of both players (with the opponent’s card being hidden and thus marked as \texttt{unknown}), and the chips in pot of both players (e.g., Fig.~\ref{fig:app:kuhn_cards}). Accordingly, the model is prompted to infer these four elements, each assigned an equal weight, and the final score is normalized to the range \([0, 1]\).

\begin{figure}[h]
  \centering
\begin{minipage}[b]{0.45\textwidth}
    \centering
    \includegraphics[height=15em]{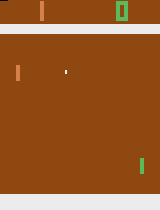}
    \caption{Origin \textit{Atari Pong}.}
    \label{fig:app:pong-origin}
\end{minipage}\hfill
\begin{minipage}[b]{0.45\textwidth}
\centering
    \includegraphics[height=15em]{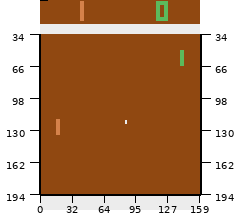}
    \caption{\textit{Atari Pong} with coordinate axes.}
    \label{fig:app:pong-axes}
\end{minipage}
\end{figure}

\subsection{Atari Pong}
\label{app:perception:pong}

The samples are uniformly drawn from trajectories generated by \texttt{o4-mini} and \texttt{doubao-1-5-thinking-pro}, both of which exhibit strong performance in the game.

The original \textit{Atari Pong} environment is continuous (e.g., Fig.~\ref{fig:app:pong-origin}). To facilitate visual recognition by VLMs, we implement an additional wrapper that overlays coordinate axes and scales on the environment. For each sample, a screenshot of the game state with coordinate axes is provided (e.g., Fig.~\ref{fig:app:pong-axes}), and the model is prompted to infer the following elements: the positions of the two paddles, the position of the ball, and the current score, with weights of $4:4:2$, respectively.

Since the paddles move only vertically, the model is required to predict their $y$-coordinates only. For the ball, the model is required to predict both $x$- and $y$-coordinates. A prediction receives full credit if it lies within a predefined radius of the ground truth; beyond this radius, the score decays with distance. The radius is fixed at $10$ in our setting.

\subsection{MPE}
\label{app:perception:mpe}

The samples are uniformly drawn from trajectories generated by \texttt{random} agents, which provide a broader distribution of game states compared to other VLMs.

\begin{figure}[h]
\begin{minipage}[b]{0.3\textwidth}
    \centering
    \includegraphics[width=\textwidth]{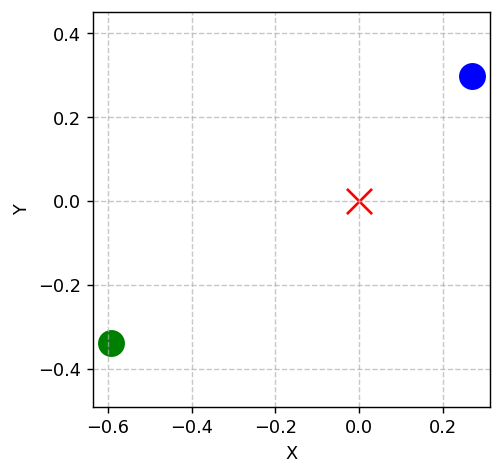}
    \caption{\textit{MPE}.}
    \label{fig:app:mpe}
\end{minipage}
\begin{minipage}[b]{0.6\textwidth}
    \centering
    \includegraphics[trim=20 0 20 0,clip,width=\textwidth]{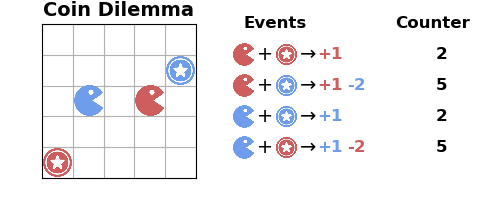}
    \caption{\textit{Coin Dilemma}.}
    \label{fig:app:dilemma}
\end{minipage}
\end{figure}

In the \textit{MPE} environment, there are only two variables representing the coordinates of the two players and a target fixed at the origin (e.g., Fig.~\ref{fig:app:mpe}). Accordingly, the model is prompted to infer the $x$- and $y$-coordinates of both players. For coordinate predictions, full credit is awarded if the predicted position lies within a predefined radius centered on the ground-truth location. Beyond this radius, the score decays proportionally to the distance from the correct position. In our setting, the radius is set to half of the unit spacing of the coordinate axis.

\subsection{Coin Dilemma}
\label{app:perception:dilemma}

The sampling strategy and dataset construction follow the same procedure as in the strategic reasoning setting (Appendix~\ref{app:reasoning:dilemma}).

In the \textit{Coin Dilemma} environment, each state is represented as a $5 \times 5$ grid. Accordingly, a screenshot of the game state (e.g., Fig.~\ref{fig:app:dilemma}) is provided for each sample, and the model is prompted to infer the content of each grid cell, ultimately producing a $5 \times 5$ matrix as output.

Each grid cell may contain the following objects: an empty cell, a red coin, a blue coin, a red player, or a blue player. The model’s score is computed as the number of correctly predicted grid cells out of 25, normalized to the range \([0, 1]\).

\subsection{Monster Hunt}
\label{app:perception:hunt}

The sampling strategy and dataset construction follow the same procedure as in the strategic reasoning setting (Appendix~\ref{app:reasoning:hunt}).

\begin{figure}[h]
\begin{minipage}[b]{0.49\textwidth}
\centering
    \includegraphics[width=\textwidth]{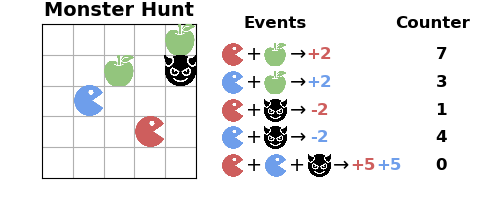}
    \caption{\textit{Monster Hunt}.}
    \label{fig:app:hunt}
\end{minipage}
\begin{minipage}[b]{0.49\textwidth}
\centering
    \includegraphics[width=\textwidth]{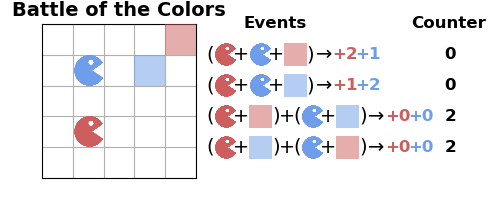}
    \caption{\textit{Battle of the Colors}.}
    \label{fig:app:battle}
\end{minipage}
\end{figure}

In the \textit{Monster Hunt} environment, each state is represented as a $5 \times 5$ grid. Accordingly, a screenshot of the game state (e.g., Fig.~\ref{fig:app:hunt}) is provided for each sample, and the model is prompted to infer the content of each grid cell, ultimately producing a $5 \times 5$ matrix as output.

Each grid cell may contain the following objects: an empty cell, an apple, a red player, a blue player, or a monster. The model’s score is computed as the number of correctly predicted grid cells out of 25, normalized to the range \([0, 1]\).

\subsection{Battle of the Colors}
\label{app:perception:battle}

The sampling strategy and dataset construction follow the same procedure as in the strategic reasoning setting (Appendix~\ref{app:reasoning:battle}).

In the \textit{Battle of the Colors} environment, each state is represented as a $5 \times 5$ grid. Accordingly, a screenshot of the game state (e.g., Fig.~\ref{fig:app:battle}) is provided for each sample, and the model is prompted to infer the content of each grid cell, ultimately producing a $5 \times 5$ matrix as output.

Each grid cell may contain the following objects: an empty cell, a red block, a blue block, a red player or a blue player. The model’s score is computed as the number of correctly predicted grid cells out of 25, normalized to the range \([0, 1]\).

\section{Strategic reasoning evaluation details}
\label{app:reasoning}
In this section, we describe the dataset generation procedure for each environment to evaluate the strategic reasoning of VLMs. Each game dataset contains 400 samples.

\subsection{Hanabi}
\label{app:reasoning:hanabi}

The dataset is generated in three steps.
First, to include data with different gameplay levels, 90\% of the samples are obtained through mutual prediction between two reasoning models, while the remaining 10\% are generated by a chat model predicting a reasoning model’s move. 
We chose \texttt{doubao-1-5-thinking-pro} and \texttt{doubao-1-5-vision-pro}, both of which demonstrated top‑tier decision‑making performance, to represent the reasoning and chat models, respectively.
Second, to balance the ratio of different actions to approximate human gameplay, we analyze the gameplay data of leading VLM agents and set the ratio of <Play>:<Discard>:<Reveal> to $2:3:4$ in our dataset.
Third, we balance the agent order and step index in the dataset so that the first and second agents each account for 50\% samples and the step index is uniformly distributed across the entire game sequence.

\subsection{Overcooked}
\label{app:reasoning:overcooked}

This dataset is derived from the human experiment data provided by Overcooked-AI~\cite{carroll2019utility}, which comprises game trajectories recorded from multiple human participants. 
We focus on the trial-train subset of these data, with instances of invalid actions filtered out from the trajectories.
We then use random sampling on these filtered trajectories to ensure comprehensive coverage of possible game states.
Each sample comprises a sequence of four consecutive game frames.
Additionally, we apply constraints to simulate realistic game scenarios and control the distribution of target actions. Specifically, the proportion of <STAY> action among the possible actions is limited to 10\%. 
The dataset is balanced for two chefs, each accounting for 50\% samples of the dataset.

\subsection{Knights Archers Zombies}
\label{app:reasoning:kaz}

The dataset is uniformly sampled from the logged trajectories of the two best-performing models in the decision-making process: \texttt{o4-mini} and \texttt{doubao-1-5-thinking-pro}. The next actions taken by these VLM agents are used as ground truth labels. Since the environment involves two types of agents, the knight and the archer, we ensure an equal split of samples, with $50\%$ corresponding to each agent type.

\subsection{Breakthrough}
\label{app:reasoning:breakthrough}

All samples are generated by a minimax algorithm with alpha–beta pruning, a widely adopted baseline in \textit{Breakthrough} research~\cite{lorentz2013programming,saffidine2012solving}. 
Since minimax search does not always reach terminal positions to determine win–loss outcomes, we implement a state evaluation function: upon reaching a fixed search depth, we compute the difference between the maximum effective forward advancement of our deepest piece and that of the opponent’s deepest piece, then normalize this difference to obtain a reward for the state.
We configure minimax agents with maximum search depths for the first and second players as $(3,4)$, $(3,5)$, $(4,5)$, $(4,6)$, $(4,4)$, and $(5,5)$, respectively, and sample step indices uniformly to ensure coverage of diverse game states.

\subsection{Kuhn poker}
\label{app:reasoning:poker}

Since \textit{Kuhn Poker} admits a continuum of mixed Nash equilibria parameterized by a single probability $\alpha \in [0, 1/3]$, representing the likelihood of betting when holding a Jack~\cite{kuhn1950simplified}, we evaluate a discrete subset of strategies. 
Specifically, we consider all pairwise matchups among three representative values of $\alpha = 0, 1/6, 1/3$, yielding nine distinct strategy combinations. 
For each combination, we simulate $600$ head-to-head games and uniformly sample $400$ game states in total to construct the final dataset.

\subsection{Atari Pong}
\label{app:reasoning:pong}

We uniformly sample $400$ state transitions from the logged trajectories of two strong reasoning models in the decision-making process, namely \texttt{o4-mini} and \texttt{doubao-1-5-thinking-pro}. 
The subsequent actions taken by these VLM agents are treated as ground-truth labels. 
To construct the evaluation setting, we modify the prompts such that the VLMs are required to control the left paddle (originally the built-in bot’s paddle) and predict the corresponding actions.

\subsection{MPE}
\label{app:reasoning:mpe}

The MPE environment consists of two agents navigating a two-dimensional plane. We uniformly sample $400$ state transitions from the logged trajectories of three strong reasoning models in the decision-making process, namely \texttt{o4-mini}, \texttt{claude-3.7-sonnet}, and \texttt{doubao-1-5-thinking-pro}. The subsequent actions taken by these VLM agents are treated as ground-truth labels. It is important to emphasize that we only log trajectories in which the VLMs assume the blocker role; consequently, the prediction task is restricted to the blocker’s actions.

\subsection{Coin Dilemma}
\label{app:reasoning:dilemma}

We consider two types of heuristic strategies for playing \textit{Coin Dilemma} and generate the dataset by simulating game play with these strategies:
\begin{enumerate}[leftmargin=2em]
  \item \textbf{Common welfare}: the agent only collects the coin of its own color.
  \item \textbf{Self interest}: the agent always collects the closest coin, regardless of the color. 
\end{enumerate}

Concretely, we collect samples from six settings, resulting in a dataset of $400$ samples:
\begin{enumerate}[leftmargin=2em]
  \item \textbf{Common welfare + common welfare}: collect 100 samples.
  \item \textbf{Self interest + self interest}: collect 100 samples.
  \item \textbf{Common welfare + self interest}: collect 50 samples.
  \item \textbf{Self interest + common welfare}: collect 50 samples.
  \item \textbf{Random + self interest}: collect 50 samples.
  \item \textbf{Self interest + random}: collect 50 samples.
\end{enumerate}

For \textit{Coin Dilemma}, different actions can lead to the same transition in the environment. Therefore, the actions with the same outcome as the ground truth action are all considered correct.
For example, if a player is at the top-left corner of the grid map, then actions <UP> and <LEFT> are both considered correct with ground truth <STAY> as they all result in no movement of the player.

\subsection{Monster Hunt}
\label{app:reasoning:hunt}

We consider four types of heuristic strategies for playing \textit{Monster Hunt} and generate the dataset by simulating game play with these strategies:
\begin{enumerate}[leftmargin=2em]
  \item \textbf{Common welfare 1}: the agent always moves towards the monster.
  \item \textbf{Common welfare 2}: the agent first moves to the middle block of the grid map and stays there to wait for the other agent and the monster.
  \item \textbf{Common welfare 3}: the agent first moves to a corner of the grid map and stays there to wait for the other agent and the monster.
  \item \textbf{Self interest}: the agent always moves towards the closest apple.
\end{enumerate}

Concretely, we collect samples from six settings, resulting in a dataset of $400$ samples:
\begin{enumerate}[leftmargin=2em]
  \item \textbf{Common welfare 1 + common welfare 1}: collect 80 samples.
  \item \textbf{Common welfare 2 + common welfare 2}: collect 80 samples.
  \item \textbf{Common welfare 3 + common welfare 3}: collect 80 samples.
  \item \textbf{Self interest + self interest}: collect 80 samples.
  \item \textbf{Random + self interest}: collect 40 samples.
  \item \textbf{Self interest + random}: collect 40 samples.
\end{enumerate}

For \textit{Monster Hunt}, different actions with the same outcome as the ground truth action are all considered correct, as in \textit{Coin Dilemma}.

\subsection{Battle of the Colors}
\label{app:reasoning:battle}

We consider four types of heuristic strategies for playing \textit{Battle of the Colors} and generate the dataset by simulating game play with these strategies:
\begin{enumerate}[leftmargin=2em]
  \item \textbf{Common welfare}: the agent always moves to the closest color block (to both players) and stays there to wait for the other player.
  \item \textbf{Self interest}: the agent always moves to the block of its own color.
  \item \textbf{Biased red}: the agent always moves to the red block.
  \item \textbf{Biased blue}: the agent always moves to the blue block. 
\end{enumerate}

Concretely, we collect samples from six settings, resulting in a dataset of $400$ samples:
\begin{enumerate}[leftmargin=2em]
  \item \textbf{Common welfare + common welfare}: collect 100 samples.
  \item \textbf{Self interest + self interest}: collect 100 samples.
  \item \textbf{Common welfare + self interest}: collect 50 samples.
  \item \textbf{Self interest + common welfare}: collect 50 samples.
  \item \textbf{Biased red + biased red}: collect 50 samples.
  \item \textbf{Biased blue + biased blue}: collect 50 samples.
\end{enumerate}

For \textit{Battle of the Colors}, different actions with the same outcome as the ground truth action are all considered correct, as in \textit{Coin Dilemma}.

\section{Decision-making evaluation details}
\label{app:decision}
In this section, we describe the methodology for evaluating the decision-making abilities of VLMs by assigning appropriate rewards after each step of interaction.

\subsection{Hanabi}
\label{app:decision:hanabi}

In \textit{Hanabi}, we consider two kinds of return.

\textbf{Standard return.} 
This is the standard return in existing \textit{Hanabi} environments~\cite{bard2020hanabi}, and we use it in the main text results. If all life tokens are consumed before the fireworks are completed, the agents get a reward of $0$. If all fireworks stacks are built successfully, the agents get a return of $25$. If the above two early terminal conditions are not reached and the deck is exhausted, the game continues for one additional round, and the return is the sum of the ranks of the fireworks.

\textbf{Firework return.} 
Since many VLM agents often consume all the life tokens and result in a zero return, we also consider another return that relaxes the ``zero‐out'' penalty upon losing all life tokens. More specifically, if all life tokens are consumed before the fireworks are completed, the return is the sum of the ranks of the firework piles. This return measures the progress of the fireworks and is reported in the raw decision-making results without normalization in Table~\ref{tab:decision-making-raw}.

For each model, we perform $10$ self‐play games and report the average standard return and the firework return. 
These results are then normalized and compared with a random baseline and an optimal policy as reported in Multi-Agent PPO (MAPPO)~\cite{yu2022surprising}.

\subsection{Overcooked}
\label{app:decision:overcooked}

In \textit{Overcooked}, each episode is limited to $50$ steps. Within these steps, two chefs cooperatively cook soup and deliver the cooked soup to the service desk. The two agents share a common return, which includes three process rewards and a delivery reward.

\textbf{Process rewards.} 
To measure the progress of making a soup, the agents get $2$ reward for each of the three events: (1) An agent successfully adds an onion to a cooking pot; (2) An agent picks up a dish when a pot contains onions or cooking is in progress; (3) An agent successfully plates a finished soup using a dish. Completing a three-onion soup leads to a total of $10$ process rewards.

\textbf{Delivery reward.} 
To align with the game's goal of successfully delivering specified dishes, the agents get $10$ reward upon successful delivery to the service desk. 

For the 3-onion soup recipe, successfully completing and delivering one soup yields a total return of $20$, consisting of a process reward of $10$ and a final reward of $10$. We evaluate each VLM over $10$ episodes of self-play, in which both agents employ the same model. The accumulated reward across an episode is reported as the episode return, and scores are further normalized with respect to the returns achieved by random and optimal agents. Notably, the optimal agents are able to complete two deliveries within a single episode.

\subsection{Knights Archers Zombies}
\label{app:decision:kaz}

In \textit{Knights Archers Zombies} (KAZ), the return is defined as the total number of zombies eliminated by the end of the game. 
The game terminates under two possible conditions: (i) reaching the predefined maximum horizon of $200$ steps, or (ii) when either of the two players is contacted by a zombie and dies. 
Although the environment inherently supports multi-agent settings, we adopt a two-agent configuration in the decision-making context considered in this work, with one archer and one knight. 
Moreover, a new zombie is spawned every $10$ steps, implying that optimal agents can achieve up to $20$ points in a single episode.

\subsection{Breakthrough}
\label{app:decision:breakthrough}

In \textit{Breakthrough}, the return is determined by the final outcome: the winning agent gets a reward of $+1$, the losing agent gets a reward of $-1$, and draws are not possible.
We selected a moderately strong MCTS agent as our baseline, configured with an exploration constant $c=2.0$, a maximum of $100$ simulations per move, and a rollout count of $10$. 
Each model is evaluated for $20$ games against this MCTS agent, with $10$ games as the first agent and $10$ games as the second agent.
For the optimal policy, we use a minimax agent with alpha–beta pruning and a maximum search depth of $5$, using a state evaluation function as described in Appendix~\ref{app:reasoning:breakthrough}. 
Although minimax is not guaranteed to be optimal for \textit{Breakthrough}, it achieved a perfect win rate against the MCTS agent in our experiment, making it a reasonable choice as the optimal policy in this study.

\subsection{Kuhn poker}
\label{app:decision:poker}

In \textit{Kuhn Poker}, the return of each agent is the net chips won or lost at the end of the game.
We consider a Nash equilibrium (NE) strategy as described in Appendix~\ref{app:reasoning:poker} with $\alpha=0$.
Each model is evaluated for ten runs of $120$ games against this NE agent, with $60$ games as the first agent and $60$ games as the second agent.
We then calculate the mean and standard deviation of the ten runs and normalize the result.
The optimal agent in \textit{Kuhn poker} is the agent with the NE strategy.

\subsection{MPE}
\label{app:decision:mpe}

In the \textit{MPE} environment, there are two types of players: a \textit{charger} and a \textit{blocker}. Since it is a competitive game, we designate the VLMs as the blocker, competing against a built-in agent that plays as the charger. The built-in agent is rule-based, with a strategy that always moves directly toward the target. 

The reward function of the blocker consists of two components:

\textbf{Push Charger}: defined as the distance between the charger and the target.

\textbf{Near Target}: defined as the negative distance between the blocker and the target.

In summary, the blocker is incentivized to stay as close to the target as possible while simultaneously keeping the charger as far away from the target as possible. In our setting, we adopt a heuristic value of $0$ as the optimal baseline.

\subsection{Atari Pong}
\label{app:decision:pong}

In \textit{Atari Pong}, we adopt frame stacking of $4$ frames to pass dynamic information to the VLM agent. We also employ a sticky action probability of $0.25$ and perform a random number (between $1$ and $30$) of <STAY> steps at the beginning of an episode to achieve randomness. These settings have been common practice in related works, such as DQN~\cite{mnih2013playing}. We also consider two kinds of reward.

\textbf{Score reward.}
The agent scores one point when the ball passes the opponent's paddle and gets a score reward of $+1$.
The game ends when an agent scores three points.

\textbf{Step reward.}
As many VLMs fail to score even one point against the built-in bot, the score reward alone becomes too sparse to distinguish the performance of different models.
Therefore, we design a continuous reward proportional to the number of steps that VLMs survived against the built-in bot.

We evaluate each VLM for $10$ episodes against the built-in bot, and the overall return is the weighted sum of the normalized score reward with a $0.9$ weight and the normalized step reward with a $0.1$ weight.
The optimal agent is an RL agent that scores three points against the built-in bot.

\subsection{Coin Dilemma}
\label{app:decision:dilemma}

In \textit{Coin Dilemma}, the agents receive rewards on different game events:
\begin{enumerate}[leftmargin=2em]
  \item \textbf{Red agent collects a red coin}: red agent gets $+1$ reward.
  \item \textbf{Red agent collects a blue coin}: red agent gets $+1$ reward, blue agent get $-2$ reward.
  \item \textbf{Blue agent collects a blue coin}: blue agent gets $+1$ reward.
  \item \textbf{Blue agent collects a red coin}: blue agent gets $+1$ reward, red agent get $-2$ reward.
\end{enumerate}

We evaluate the return of each VLM for $10$ episodes of self-play, where the red and blue agents use the same model. We further normalize these returns with respect to the returns of the random agents and the optimal agents that always move towards the coin of their own color.

\subsection{Monster Hunt}
\label{app:decision:hunt}

In \textit{Monster Hunt}, the agents receive rewards on different game events:
\begin{enumerate}[leftmargin=2em]
  \item \textbf{Red agent eats an apple}: red agent gets $+2$ reward.
  \item \textbf{Blue agent eats an apple}: blue agent gets $+2$ reward.
  \item \textbf{Red agent encounters the monster alone}: red agent gets $-2$ reward.
  \item \textbf{Blue agent encounters the monster alone}: blue agent gets $-2$ reward.
  \item \textbf{Both agents defeat the monster together}: both agents get $+5$ reward.
\end{enumerate}

We evaluate the return of each VLM for $10$ episodes of self-play, where the red and blue agents use the same model. We further normalize these returns with respect to the returns of the random agents and the optimal agents that always move to the middle block and stay there to wait for the monster.

\subsection{Battle of the Colors}
\label{app:decision:battle}

In \textit{Battle of the Colors}, the agents receive rewards on different game events:
\begin{enumerate}[leftmargin=2em]
  \item \textbf{Both players on the red block}: red agent gets $+2$ reward, blue agent gets $+1$ reward.
  \item \textbf{Both players on the blue block}: blue agent gets $+2$ reward, red agent gets $+1$ reward.
  \item \textbf{Two players on different blocks}: both agents get $0$ reward.
\end{enumerate}

We evaluate the return of each VLM for $10$ episodes of self-play, where the red and blue agents use the same model. We further normalize these returns with respect to the returns of the random agents and the optimal agents that always move to the closest block to the two players.

\subsection{Raw results without normalization}

The decision-making results in all environments without normalization are listed in Table~\ref{tab:decision-making-raw}. We also report the firework return for \textit{Hanabi} and the score reward and step reward for \textit{Atari Pong}.

\begin{table}[H]
\centering
\scriptsize
\caption{Raw decision-making results without normalization. 
The \textit{Hanabi} standard\textsuperscript{1} and firework\textsuperscript{2} returns are described in Appendix~\ref{app:decision:hanabi}, 
while the \textit{Pong} score\textsuperscript{3} and step\textsuperscript{4} rewards are described in Appendix~\ref{app:decision:pong}.}
\begin{adjustbox}{angle=90, max width=\textheight} 
  \setlength{\tabcolsep}{4pt}
\begin{tabular}{ccccccccccccc}
\toprule
\multirow{3}{*}{Models} & \multicolumn{4}{c}{Cooperative} & \multicolumn{5}{c}{Competitive} & \multicolumn{3}{c}{Mixed-Motive} \\
 & \multicolumn{2}{c}{\textit{Hanabi}} & \multirow{2}{*}{\textit{Overcooked}} & \multirow{2}{*}{\textit{KAZ}} & \multirow{2}{*}{\textit{Board}} & \multirow{2}{*}{\textit{Poker}} & \multicolumn{2}{c}{\textit{Pong}} & \multirow{2}{*}{\textit{MPE}} & \multirow{2}{*}{\textit{Dilemma}} & \multirow{2}{*}{\textit{Hunt}} & \multirow{2}{*}{\textit{Battle}} \\
 & standard\textsuperscript{1} & firework\textsuperscript{2} &  &  &  &  & score\textsuperscript{3} & step\textsuperscript{4} &  &  &  &  \\

\midrule

Optimal & $24.0$ & $24.0$ & $40.0$ & $20.0$ & $1.0$ & $0.0$ & $3.0$ & $398.0$ & $0.0$ & $27.9$ & $176.2$ & $59.7$ \\

\midrule

\texttt{o3} & $13.4_{\pm5.0}$ & $13.6_{\pm4.5}$ & $6.4_{\pm1.2}$ & $6.6_{\pm3.1}$ & $0.6_{\pm0.8}$ & $-0.0_{\pm0.0}$ & $0.2_{\pm0.4}$ & $219.4_{\pm113.4}$ & $-6.5_{\pm9.3}$ & $-0.3_{\pm4.9}$ & $27.0_{\pm6.3}$ & $10.2_{\pm3.1}$ \\

\texttt{o4-mini} & $10.3_{\pm7.3}$ & $13.3_{\pm2.9}$ & $7.0_{\pm2.7}$ & $3.7_{\pm2.5}$ & $-0.4_{\pm0.9}$ & $-0.0_{\pm0.0}$ & $0.3_{\pm0.5}$ & $205.2_{\pm91.0}$ & $-6.6_{\pm8.3}$ & $-1.5_{\pm6.0}$ & $28.6_{\pm16.1}$ & $2.4_{\pm3.2}$ \\

\texttt{gemini-2.5-pro} & $7.9_{\pm8.5}$ & $11.6_{\pm5.2}$ & $7.0_{\pm2.2}$ & $2.6_{\pm1.6}$ & $0.1_{\pm1.0}$ & $-0.1_{\pm0.1}$ & $0.1_{\pm0.3}$ & $246.7_{\pm65.2}$ & $-6.1_{\pm8.0}$ & $-2.9_{\pm5.5}$ & $22.0_{\pm11.7}$ & $20.4_{\pm3.8}$ \\

\texttt{gemini-2.5-flash} & $6.5_{\pm8.6}$ & $10.7_{\pm5.3}$ & $3.6_{\pm2.1}$ & $3.7_{\pm1.7}$ & $-0.6_{\pm0.5}$ & $-0.1_{\pm0.1}$ & $0.0_{\pm0.0}$ & $194.4_{\pm53.2}$ & $-7.7_{\pm8.9}$ & $2.6_{\pm7.2}$ & $31.2_{\pm11.4}$ & $19.8_{\pm5.1}$ \\

\texttt{claude-3.7-sonnet} & $1.6_{\pm5.1}$ & $9.7_{\pm3.9}$ & $4.2_{\pm1.4}$ & $2.0_{\pm1.4}$ & $-0.6_{\pm0.8}$ & $-0.0_{\pm0.0}$ & $0.0_{\pm0.0}$ & $133.7_{\pm27.8}$ & $-5.9_{\pm8.5}$ & $1.1_{\pm4.3}$ & $18.8_{\pm6.9}$ & $1.8_{\pm2.8}$ \\

\texttt{ui-tars-1-5} & $0.0_{\pm0.0}$ & $4.9_{\pm1.7}$ & $1.0_{\pm1.3}$ & $0.1_{\pm0.3}$ & $-0.9_{\pm0.3}$ & $-0.1_{\pm0.0}$ & $0.0_{\pm0.0}$ & $121.4_{\pm8.4}$ & $-11.2_{\pm12.9}$ & $9.1_{\pm4.0}$ & $28.6_{\pm6.5}$ & $8.4_{\pm1.7}$ \\

\texttt{doubao-1.5-thinking-pro} & $13.6_{\pm5.5}$ & $14.1_{\pm4.0}$ & $4.2_{\pm1.9}$ & $3.7_{\pm1.7}$ & $-0.8_{\pm0.4}$ & $-0.0_{\pm0.0}$ & $0.0_{\pm0.0}$ & $230.5_{\pm72.6}$ & $-6.2_{\pm10.3}$ & $0.0_{\pm0.9}$ & $13.6_{\pm22.3}$ & $2.7_{\pm2.8}$ \\

\texttt{qvq-max} & $0.0_{\pm0.0}$ & $4.9_{\pm2.9}$ & $1.0_{\pm1.3}$ & $0.7_{\pm0.9}$ & $-0.9_{\pm0.3}$ & $-0.1_{\pm0.0}$ & $0.0_{\pm0.0}$ & $158.2_{\pm46.9}$ & $-6.1_{\pm8.3}$ & $-0.2_{\pm0.6}$ & $-18.8_{\pm8.8}$ & $0.0_{\pm0.0}$ \\

\midrule

\texttt{gemini-2.5} w/o thinking & $0.0_{\pm0.0}$ & $3.8_{\pm1.6}$ & $1.0_{\pm1.6}$ & $1.0_{\pm0.8}$ & $-1.0_{\pm0.0}$ & $-0.1_{\pm0.0}$ & $0.0_{\pm0.0}$ & $175.9_{\pm41.5}$ & $-10.0_{\pm7.7}$ & $-0.4_{\pm1.2}$ & $-18.8_{\pm17.4}$ & $1.8_{\pm2.0}$ \\

\texttt{claude-3.7} w/o thinking & $0.0_{\pm0.0}$ & $2.9_{\pm0.9}$ & $1.0_{\pm1.6}$ & $1.7_{\pm1.7}$ & $-0.9_{\pm0.3}$ & $-0.1_{\pm0.1}$ & $0.0_{\pm0.0}$ & $121.4_{\pm8.4}$ & $-14.8_{\pm14.0}$ & $0.2_{\pm2.6}$ & $-19.8_{\pm16.0}$ & $0.9_{\pm1.4}$ \\

\texttt{gpt-4.1} & $0.0_{\pm0.0}$ & $3.6_{\pm1.4}$ & $0.0_{\pm0.0}$ & $0.0_{\pm0.0}$ & $-1.0_{\pm0.0}$ & $-0.1_{\pm0.0}$ & $0.0_{\pm0.0}$ & $151.8_{\pm41.5}$ & $-8.0_{\pm6.6}$ & $4.8_{\pm1.9}$ & $1.8_{\pm10.9}$ & $0.6_{\pm1.2}$ \\

\texttt{qwen-vl-max} & $0.3_{\pm0.5}$ & $0.3_{\pm0.5}$ & $0.0_{\pm0.0}$ & $0.0_{\pm0.0}$ & $-1.0_{\pm0.0}$ & $-0.1_{\pm0.1}$ & $0.0_{\pm0.0}$ & $139.7_{\pm29.3}$ & $-12.5_{\pm12.0}$ & $-0.3_{\pm0.8}$ & $5.8_{\pm39.6}$ & $0.0_{\pm0.0}$ \\

\texttt{doubao-1.5-vision-pro} & $0.0_{\pm0.0}$ & $4.6_{\pm1.1}$ & $0.0_{\pm0.0}$ & $0.0_{\pm0.0}$ & $-1.0_{\pm0.0}$ & $-0.1_{\pm0.1}$ & $0.0_{\pm0.0}$ & $121.4_{\pm8.4}$ & $-7.7_{\pm9.3}$ & $-0.8_{\pm1.5}$ & $-4.8_{\pm16.1}$ & $0.0_{\pm0.0}$ \\

\texttt{grok-2-vision} & $0.0_{\pm0.0}$ & $1.6_{\pm1.0}$ & $0.8_{\pm1.3}$ & $1.0_{\pm1.4}$ & $-1.0_{\pm0.0}$ & $-0.1_{\pm0.1}$ & $0.2_{\pm0.4}$ & $152.6_{\pm45.9}$ & $-31.6_{\pm24.7}$ & $0.1_{\pm2.0}$ & $-21.0_{\pm11.4}$ & $0.6_{\pm1.2}$ \\

\midrule

\texttt{Qwen2.5-VL-72B-Ins.} & $0.2_{\pm0.4}$ & $0.2_{\pm0.4}$ & $0.0_{\pm0.0}$ & $0.0_{\pm0.0}$ & $-1.0_{\pm0.0}$ & $-0.1_{\pm0.1}$ & $0.0_{\pm0.0}$ & $123.8_{\pm4.6}$ & $-11.1_{\pm10.5}$ & $-0.2_{\pm0.8}$ & $18.2_{\pm50.4}$ & $0.0_{\pm0.0}$ \\

\texttt{InternVL3-78B} & $0.0_{\pm0.0}$ & $2.4_{\pm1.0}$ & $0.2_{\pm0.6}$ & $0.3_{\pm0.5}$ & $-1.0_{\pm0.0}$ & $-0.1_{\pm0.1}$ & $0.0_{\pm0.0}$ & $121.4_{\pm8.4}$ & $-14.8_{\pm14.8}$ & $1.2_{\pm2.0}$ & $-20.6_{\pm10.8}$ & $1.2_{\pm1.5}$ \\

\texttt{Llama-3.2-90B-Vision-Ins.} & $0.0_{\pm0.0}$ & $1.2_{\pm1.6}$ & $0.8_{\pm1.3}$ & $1.0_{\pm0.8}$ & $-1.0_{\pm0.0}$ & $-0.1_{\pm0.1}$ & $0.0_{\pm0.0}$ & $121.4_{\pm8.4}$ & $-24.1_{\pm17.6}$ & $-0.1_{\pm0.9}$ & $-13.2_{\pm9.6}$ & $0.9_{\pm1.4}$ \\

\midrule

Random & $0.0$ & $1.2$ & $0.2$ & $1.0$ & $-1.0$ & $-0.1$ & $0.0$ & $147.2$ & $-16.3$ & $-0.2$ & $-20.2$ & $0.3$ \\

\bottomrule
\end{tabular}
\setlength{\tabcolsep}{6pt}

\end{adjustbox}
\label{tab:decision-making-raw}
\end{table}

\section{Additional experiment results}
\label{app:results}
\begin{table}[h]
\centering
\scriptsize
\caption{Strategic reasoning results on multimodal input and CoT prompting. }
\makebox[\textwidth][c]{\setlength{\tabcolsep}{7.5pt}
\centering
\begin{tabular}{cccccccccc}
\toprule
\multirow{2}{*}{Model} & \multicolumn{3}{c}{\textit{Hanabi}} & \multicolumn{3}{c}{\textit{Breakthrough}} & \multicolumn{3}{c}{\textit{Monster Hunt}} \\
& Text-Only & Multimodal & CoT & Text-Only & Multimodal & CoT& Text-Only & Multimodal & CoT \\
\midrule
Optimal & $100.0$ & $100.0$ & $100.0$ & $100.0$ & $100.0$ & $100.0$ & $100.0$ & $100.0$ & $100.0$  \\
\midrule

\texttt{o3}  & $58.2$ & $61.2$ & N/A & $31.0$ & $29.0$ & N/A & $50.8$ & $57.8$ & N/A \\

\texttt{o4-mini} & $53.8$ & $58.2$ & N/A & $27.5$ & $26.8$ & N/A & $47.8$ & $43.8$ & N/A  \\

\texttt{gemini-2.5-pro}  & $54.2$ & $51.2$ & N/A & $28.2$ & $26.8$ & N/A & $55.0$ & $35.8$ & N/A \\

\texttt{gemini-2.5-flash} & $42.0$ & $37.0$ & N/A & $23.5$ & $23.2$ & N/A & $49.0$ & $50.2$ & N/A \\

\texttt{claude-3-7-sonnet} & $45.0$ & $39.0$ & N/A & $25.0$ & $24.2$ & N/A & $56.0$ & $42.5$ & N/A  \\

\texttt{doubao-1-5-thinking-pro} & $34.5$ & $32.8$ & N/A & $23.5$ & $19.8$ & N/A & $57.2$ & $45.2$ & N/A  \\

\texttt{ui-tars-1-5}   & $24.2$ & $25.5$ & N/A & $27.8$ & $23.3$ & N/A & $34.0$ & $41.5$ & N/A \\

\texttt{qvq-max} & $41.0$ & $32.2$ & N/A & $27.5$ & $21.8$ & N/A & $48.2$ & $21.5$ & N/A  \\

\midrule

\texttt{gemini-2.5} w/o thinking & $24.5$ & $21.5$ & $24.0$ & $20.5$ & $14.8$ & $21.5$ & $40.2$ & $30.0$ & $45.0$  \\

\texttt{claude-3-7} w/o thinking & $19.2$ & $9.8$ & $32.8$ & $19.2$ & $18.0$ & $19.0$ & $37.0$ & $26.0$ & $43.8$ \\

\texttt{gpt-4.1} & $40.0$ & $23.0$ & $49.8$ & $20.5$ & $22.5$ & $27.5$ & $41.0$ & $36.8$ & $34.5$  \\

\texttt{qwen-vl-max} & $17.0$ & $26.5$ & $20.0$ & $19.0$ & $19.5$ & $17.2$ & $31.8$ & $23.5$ & $35.5$  \\

\texttt{doubao-1-5-vision-pro} & $19.5$ & $15.0$ & $25.2$ & $17.2$ & $15.8$ & $16.8$ & $33.2$ & $36.0$ & $39.2$  \\

\texttt{grok-2-vision} & $23.8$ & $12.8$ & $22.5$ & $14.0$ & $10.8$ & $18.2$ & $42.5$ & $31.5$ & $33.2$  \\

\midrule

\texttt{Qwen2.5-VL-72B-Ins.} & $18.5$ & $26.8$ & $22.2$ & $19.2$ & $23.8$ & $16.5$ & $32.2$ & $27.2$ & $37.5$  \\

\texttt{InternVL3-78B} & $26.8$ & $25.2$ & $20.5$ & $17.5$ & $14.0$ & $16.0$ & $34.2$ & $30.0$ & $37.8$  \\

\texttt{Llama-3.2-90B-Vision-Ins.} & $26.8$ & $20.0$ & $14.8$ & $6.5$ & $11.8$ & $14.0$ & $36.8$ & $26.2$ & $27.8$  \\

\midrule

Random & $8.8$ & $8.8$ & $8.8$ & $4.3$ & $4.3$ & $4.3$ & $29.3$ & $29.3$ & $29.3$  \\

\bottomrule
\end{tabular}
\setlength{\tabcolsep}{6pt}

}
\label{tab:ablation_reasoning}
\end{table}

\begin{table}[h]
\centering
\scriptsize
\caption{Decision-making results on multimodal input and CoT prompting.}
\makebox[\textwidth][c]{

\setlength{\tabcolsep}{5pt}
\centering
\begin{tabular}{cccccccccc}
\toprule
\multirow{2}{*}{Model} & \multicolumn{3}{c}{\textit{Hanabi}} & \multicolumn{3}{c}{\textit{Breakthrough}} & \multicolumn{3}{c}{\textit{Monster Hunt}} \\
& Text-Only & Multimodal & CoT & Text-Only & Multimodal & CoT& Text-Only & Multimodal & CoT \\
\midrule
Optimal & $100.0$ & $100.0$ & $100.0$ & $100.0$ & $100.0$ & $100.0$ & $100.0$ & $100.0$ & $100.0$  \\
\midrule

\texttt{o3} & $62.9_{\pm7.1}$ & $55.8_{\pm20.9}$ & N/A & $95.0_{\pm15.0}$ & $80.0_{\pm40.0}$ & N/A & $45.3_{\pm7.7}$ & $24.0_{\pm3.2}$ & N/A  \\

\texttt{o4-mini} & $37.1_{\pm26.1}$ & $42.9_{\pm30.5}$ & N/A & $30.0_{\pm47.0}$ & $30.0_{\pm47.0}$ & N/A & $23.2_{\pm4.2}$ & $24.9_{\pm8.2}$ & N/A  \\

\texttt{gemini-2.5-pro} & $22.5_{\pm36.7}$ & $32.9_{\pm35.5}$ & N/A & $55.0_{\pm49.5}$ & $55.0_{\pm49.5}$ & N/A & $50.4_{\pm9.0}$ & $21.5_{\pm6.0}$ & N/A  \\

\texttt{gemini-2.5-flash} & $40.8_{\pm21.9}$ & $27.1_{\pm36.0}$ & N/A & $30.0_{\pm42.3}$ & $20.0_{\pm25.7}$ & N/A & $15.63_{\pm4.6}$ & $26.2_{\pm5.8}$ & N/A\\

\texttt{claude-3-7-sonnet} & $33.8_{\pm35.8}$ & $6.7_{\pm21.1}$ & N/A & $45.0_{\pm50.0}$ & $20.0_{\pm39.8}$ & N/A & $27.3_{\pm4.6}$ & $19.9_{\pm3.5}$ & N/A  \\

\texttt{doubao-1-5-thinking-pro} & $37.5_{\pm32.9}$ & $56.7_{\pm22.8}$ & N/A & $15.0_{\pm37.0}$ & $10.0_{\pm21.0}$ & N/A & $27.6_{\pm6.1}$ & $17.2_{\pm11.3}$ & N/A  \\

\texttt{ui-tars-1-5} & $0.0_{\pm0.0}$ & $0.0_{\pm0.0}$ & N/A & $20.0_{\pm40.0}$ & $30.0_{\pm47.0}$ & N/A & $12.6_{\pm5.1}$ & $24.8_{\pm3.3}$ & N/A \\

\texttt{qvq-max} & $0.0_{\pm0.0}$ & $0.0_{\pm0.0}$ & N/A & $5.0_{\pm15.8}$ & $5.0_{\pm15.8}$ & N/A & $15.1_{\pm5.7}$ & $0.7_{\pm4.5}$ & N/A \\

\midrule

\texttt{gemini-2.5} w/o thinking & $0.0_{\pm0.0}$ & $0.0_{\pm0.0}$ & $3.3_{\pm10.5}$ & $0.0_{\pm0.0}$ & $0.0_{\pm0.0}$ & $20.0_{\pm39.8}$ & $5.1_{\pm4.9}$ & $0.7_{\pm8.9}$ & $6.3_{\pm9.8}$  \\

\texttt{claude-3-7} w/o thinking & $0.0_{\pm0.0}$ & $0.0_{\pm0.0}$ & $0.0_{\pm0.0}$ & $5.0_{\pm15.8}$ & $5.0_{\pm15.8}$ & $10.0_{\pm31.5}$ & $15.0_{\pm5.8}$ & $0.2_{\pm8.2}$ & $12.4_{\pm8.6}$ \\

\texttt{gpt-4.1} & $0.0_{\pm0.0}$ & $0.0_{\pm0.0}$ & $0.0_{\pm0.0}$ & $0.0_{\pm0.0}$ & $0.0_{\pm0.0}$ & $10.0_{\pm31.5}$ & $13.0_{\pm6.8}$ & $11.2_{\pm5.6}$ & $18.5_{\pm10.9}$  \\

\texttt{qwen-vl-max} & $0.0_{\pm0.0}$ & $1.2_{\pm2.0}$ & $0.0_{\pm0.0}$ & $5.0_{\pm15.8}$ & $0.0_{\pm0.0}$ & $0.0_{\pm0.0}$ & $14.1_{\pm7.3}$ & $13.2_{\pm20.2}$ & $-0.6_{\pm8.3}$  \\

\texttt{doubao-1-5-vision-pro} & $0.0_{\pm0.0}$ & $0.0_{\pm0.0}$ & $5.0_{\pm5.0}$ & $10.0_{\pm21.0}$ & $0.0_{\pm0.0}$ & $5.0_{\pm31.5}$ & $7.8_{\pm8.8}$ & $8.1_{\pm6.3}$ & $16.2_{\pm15.0}$  \\

\texttt{grok-2-vision} & $0.0_{\pm0.0}$ & $0.0_{\pm0.0}$ & $0.0_{\pm0.0}$ & $0.0_{\pm0.0}$ & $0.0_{\pm0.0}$ & $0.0_{\pm0.0}$ & $5.1_{\pm5.3}$ & $-0.4_{\pm5.8}$ & $3.0_{\pm3.9}$ \\

\midrule

\texttt{Qwen2.5-VL-72B-Ins.} & $6.2_{\pm6.6}$ & $0.8_{\pm1.8}$ & $2.9_{\pm6.2}$ & $0.0_{\pm0.0}$ & $0.0_{\pm0.0}$ & $0.0_{\pm0.0}$ & $16.2_{\pm9.6}$ & $19.6_{\pm25.7}$ & $23.3_{\pm22.9}$  \\

\texttt{InternVL3-78B} & $0.0_{\pm0.0}$ & $0.0_{\pm0.0}$ & $1.7_{\pm5.2}$ & $0.0_{\pm0.0}$ & $0.0_{\pm0.0}$ & $0.0_{\pm0.0}$ & {$4.8_{\pm3.9}$} & $-1.8_{\pm9.2}$ & $8.2_{\pm7.6}$  \\

\texttt{Llama-3.2-90B-Vision-Ins.} & $0.0_{\pm0.0}$ & $0.0_{\pm0.0}$ & $0.0_{\pm0.0}$ & $0.0_{\pm0.0}$ & $0.0_{\pm0.0}$ & $0.0_{\pm0.0}$ & $5.5_{\pm3.7}$ & $3.6_{\pm4.9}$ & $3.0_{\pm8.8}$  \\

\midrule

Random & $0.0$ & $0.0$ & $0.0$ & $0.0$ & $0.0$ & $0.0$ & $0.0$ & $0.0$ & $0.0$  \\

\bottomrule
\end{tabular}
\setlength{\tabcolsep}{6pt}

}
\label{tab:ablation_decision}
\end{table}

\subsection{Multimodal observation results}
\label{app:results:multimodal}

We select a cooperative card game \textit{Hanabi}, a competitive board game \textit{Breakthrough}, and a mixed-motive video game \textit{Monster Hunt}, and evaluate the performance of fifteen VLMs with multimodal observations and text-only observations.
The evaluation results of strategic reasoning are shown in Table~\ref{tab:ablation_reasoning} and the evaluation results of decision-making are shown in  Table~\ref{tab:ablation_decision}.
In general, most VLMs achieve better results with text-only observations in both strategic reasoning and decision-making.
However, multimodal observations provide richer information and should lead to at least the same performance. 
These results demonstrate that VLMs can fail to extract visual information in multimodal observations and improve strategic reasoning and decision-making performance.

\subsection{Test-time scaling results}
\label{app:results:scaling}

We consider the same three games as in multimodal observation experiments and evaluate the performance of six chat VLMs and three open-source VLMs with IO prompting and Chain-of-Thought (CoT) prompting~\cite{wei2022chain}.
We do not consider CoT prompting for reasoning models because they already generate step-by-step reasoning by themselves.
The evaluation results of strategic reasoning are shown in Table~\ref{tab:ablation_reasoning}, and the evaluation results of decision-making are shown in  Table~\ref{tab:ablation_decision}.
In general, CoT prompting leads to much better performance, which shows that test-time scaling methods like reasoning and CoT prompting can substantially improve VLMs' performance.

\subsection{Social behaviors and personas results}
\label{app:results:social}

\begin{figure}[t]
\centering

\begin{subfigure}[b]{0.32\linewidth}
    \centering
    \includegraphics[width=\linewidth]{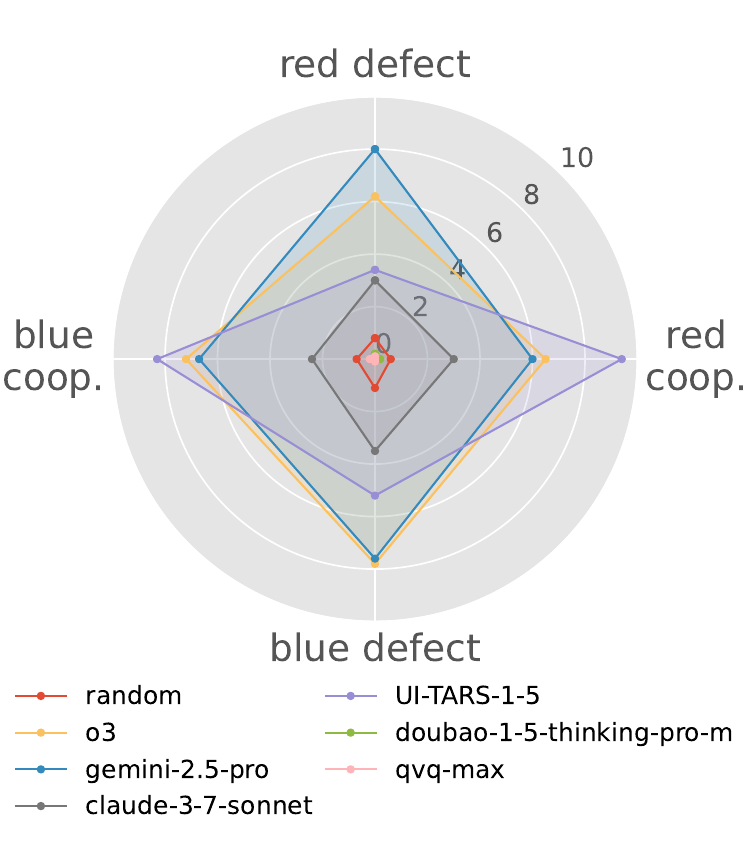}
    \caption{Reasoning VLMs in \textit{Dilemma}.}
    \label{fig:app:reasoning:dilemma}
\end{subfigure}
\hfill
\begin{subfigure}[b]{0.32\linewidth}
    \centering
    \includegraphics[width=\linewidth]{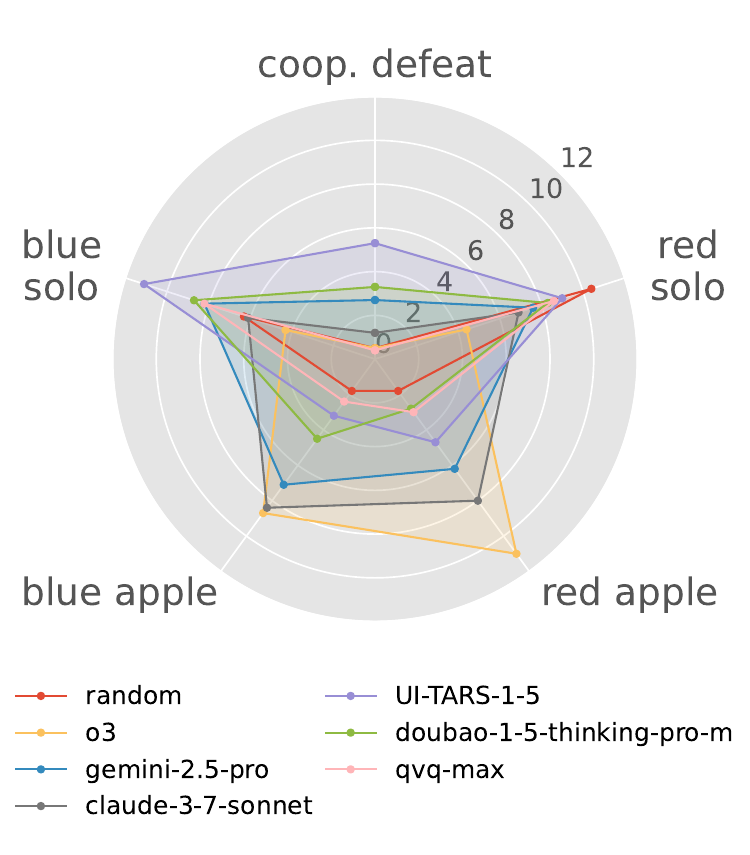}
    \caption{Reasoning VLMs in \textit{Hunt}.}
    \label{fig:app:reasoning:hunt}
\end{subfigure}
\hfill
\begin{subfigure}[b]{0.32\linewidth}
    \centering
    \includegraphics[width=\linewidth]{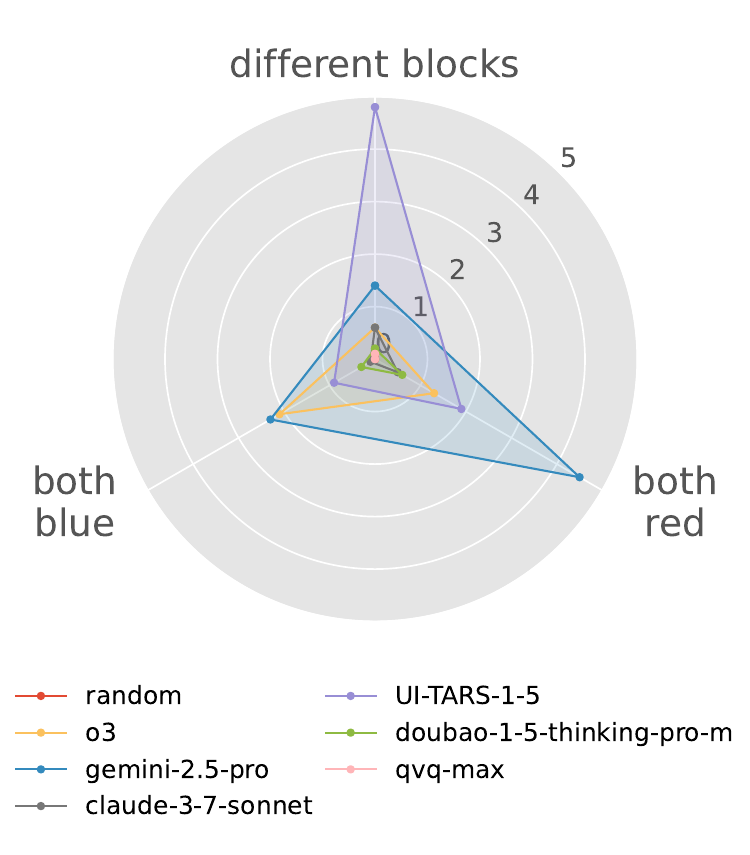}
    \caption{Reasoning VLMs in \textit{Battle}.}
    \label{fig:app:reasoning:battle}
\end{subfigure}

\begin{subfigure}[b]{0.32\linewidth}
    \centering
    \includegraphics[width=\linewidth]{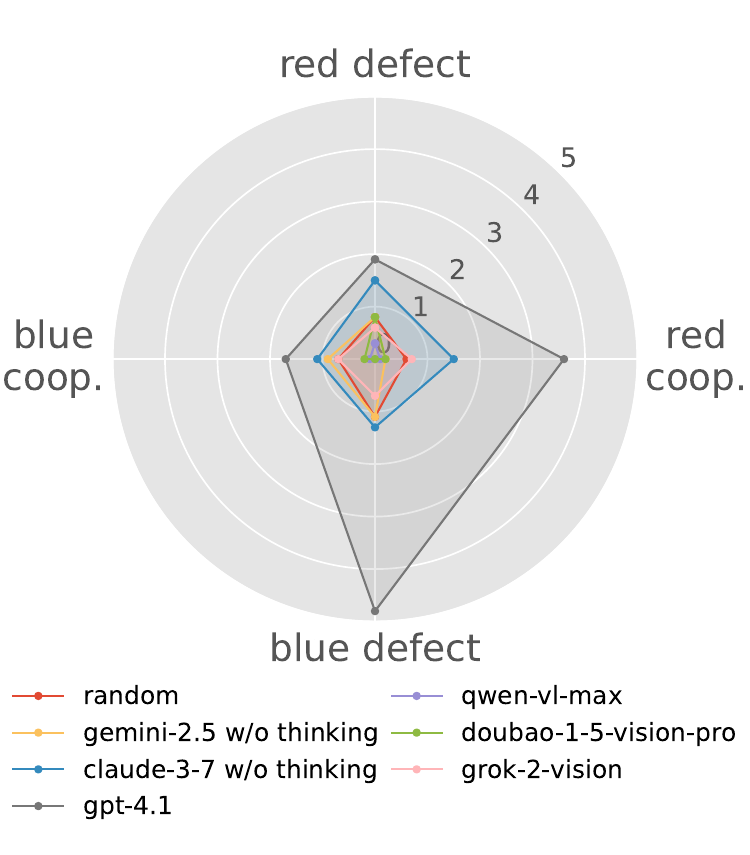}
    \caption{Chat VLMs in \textit{Dilemma}.}
    \label{fig:app:chat:dilemma}
\end{subfigure}
\hfill
\begin{subfigure}[b]{0.32\linewidth}
    \centering
    \includegraphics[width=\linewidth]{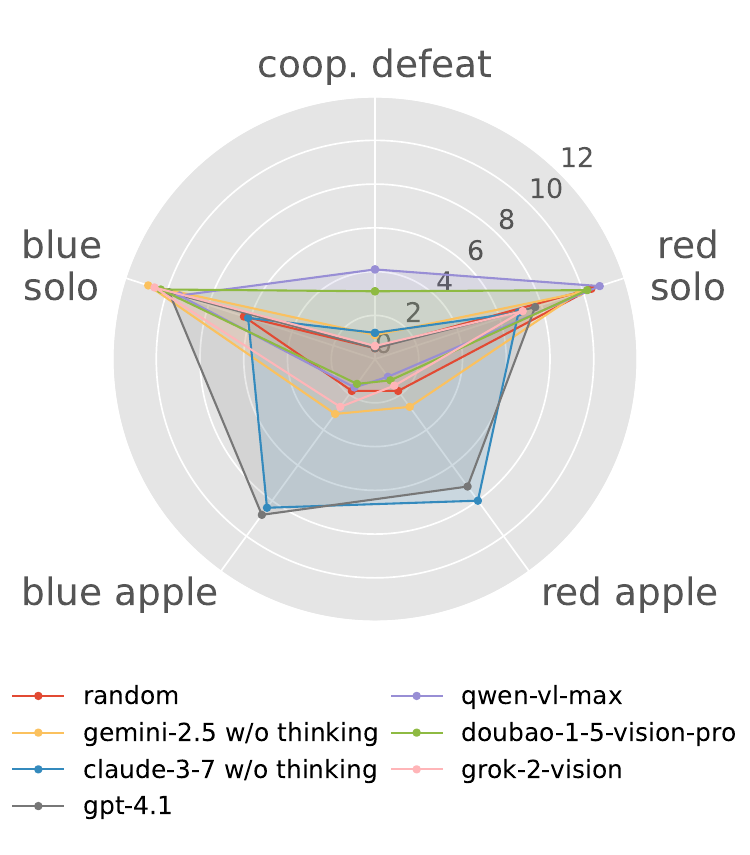}
    \caption{Chat VLMs in \textit{Hunt}.}
    \label{fig:app:chat:hunt}
\end{subfigure}
\hfill
\begin{subfigure}[b]{0.32\linewidth}
    \centering
    \includegraphics[width=\linewidth]{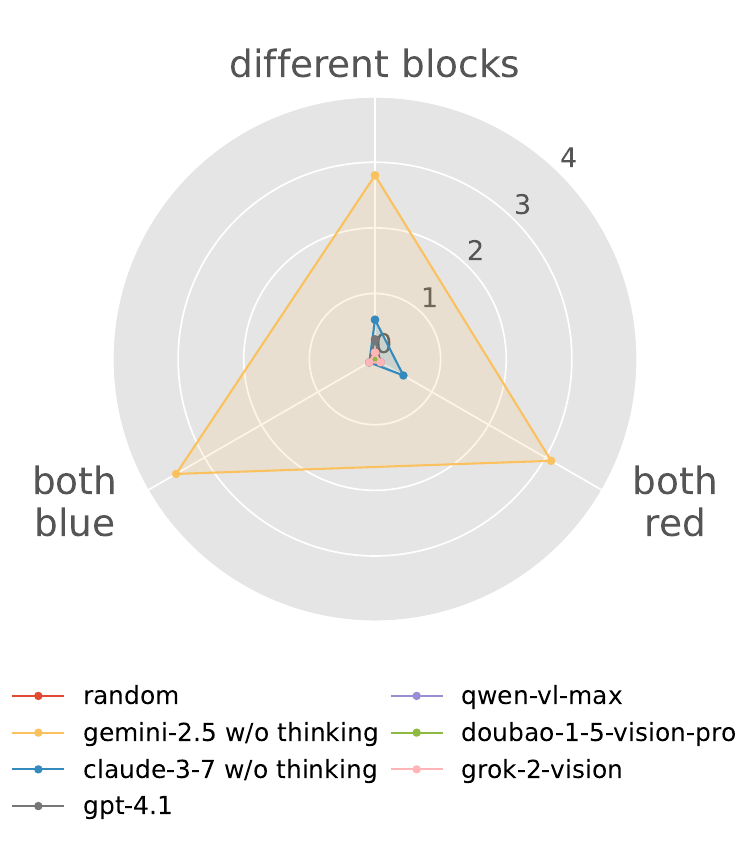}
    \caption{Chat VLMs in \textit{Battle}.}
    \label{fig:app:chat:battle}
\end{subfigure}

\begin{subfigure}[b]{0.32\linewidth}
    \centering
    \includegraphics[width=\linewidth]{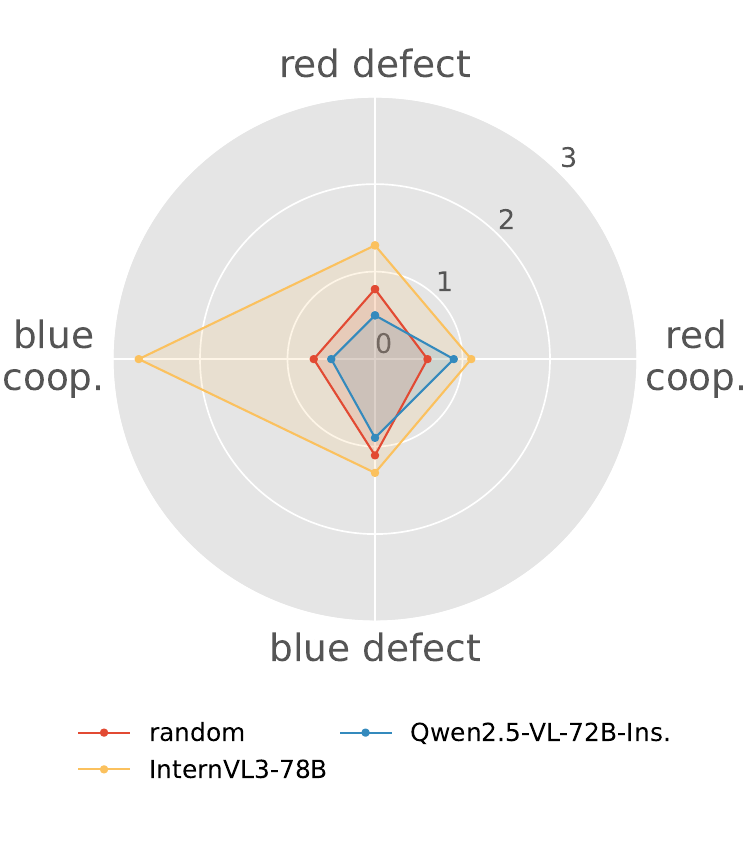}
    \caption{Open-source VLMs in \textit{Dilemma}.}
    \label{fig:app:open:dilemma}
\end{subfigure}
\hfill
\begin{subfigure}[b]{0.32\linewidth}
    \centering
    \includegraphics[width=\linewidth]{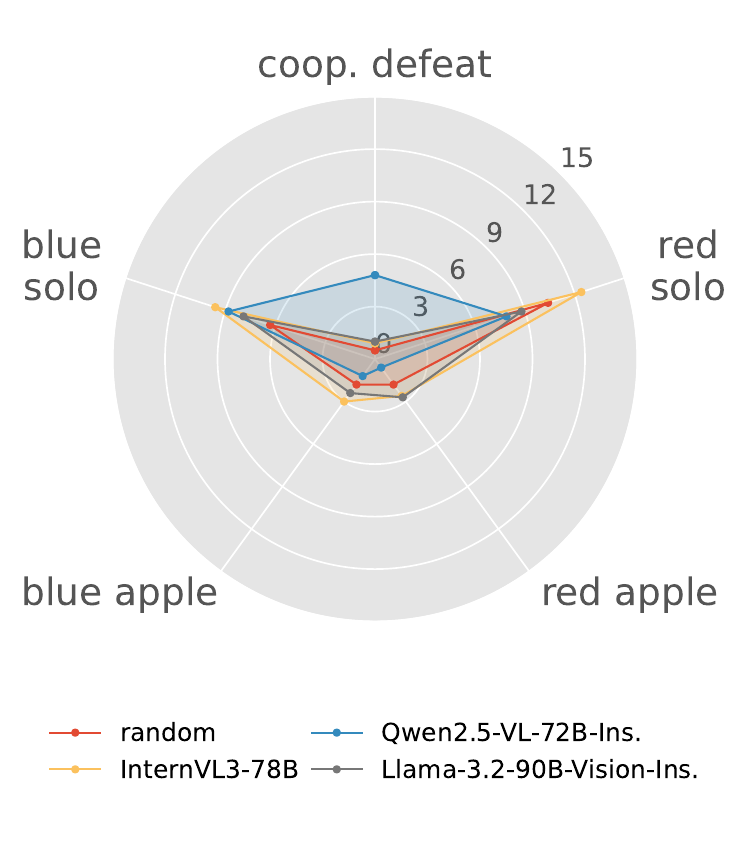}
    \caption{Open-source VLMs in \textit{Hunt}.}
    \label{fig:app:open:hunt}
\end{subfigure}
\hfill
\begin{subfigure}[b]{0.32\linewidth}
    \centering
    \includegraphics[width=\linewidth]{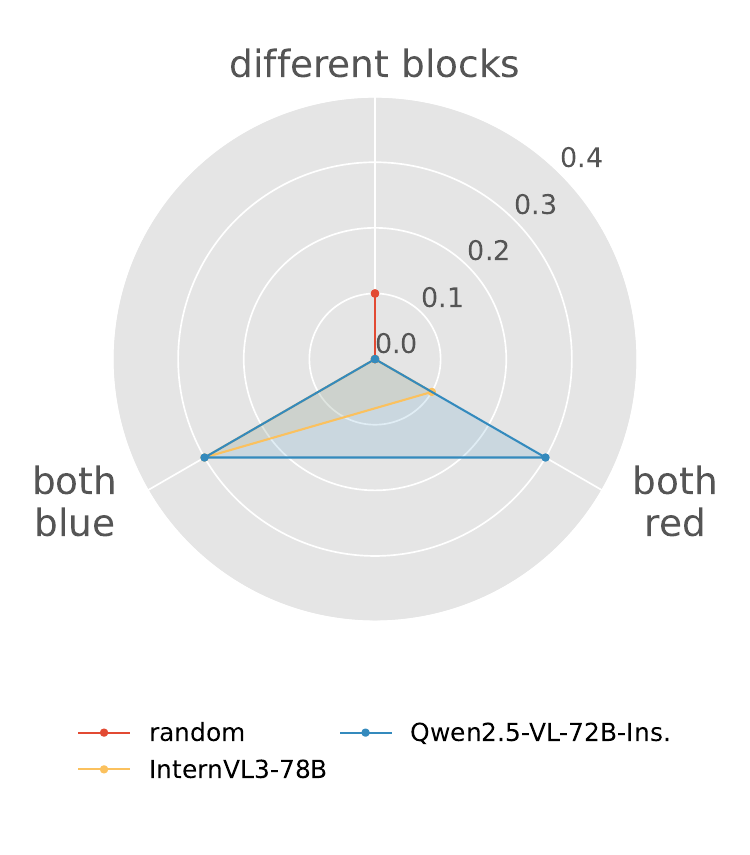}
    \caption{Open-source VLMs in \textit{Battle}.}
    \label{fig:app:open:battle}
\end{subfigure}
\caption{Social behaviors of all models in mixed-motive social dilemma games. Dimensions are agents' behaviors described in Sec.4.3.}
\label{fig:app:social}
\end{figure}

We evaluated the social behavior of all models in three mixed-motive games. Additionally, we selected three well-performing reasoning models to test their behavior under different persona settings.

\subsubsection{Social behaviors}

For \textit{Coin Dilemma}, the social behaviors of different models are shown in Fig.~\ref{fig:app:reasoning:dilemma}, \ref{fig:app:chat:dilemma}, and \ref{fig:app:open:dilemma}.
Compared to reasoning models, chat models exhibit inferior performance in coin collection, leading to fewer instances of both cooperation and defection. The best-performing chat model, \texttt{GPT-4.1}, demonstrates more defections than cooperative actions, suggesting a behavioral tendency that prioritizes self-interest over collective welfare. In contrast, the open-source model \texttt{InternVL3-78B} displays the opposite behavioral pattern.

For \textit{Monster Hunt}, the social behaviors of different models are shown in Fig.~\ref{fig:app:reasoning:hunt}, \ref{fig:app:chat:hunt}, and \ref{fig:app:open:hunt}.
For reasoning models, \texttt{o3}, \texttt{gemini-2.5-pro}, and \texttt{claude-3-7-sonnet} prefer eating apples alone; \texttt{doubao-1-5-thinking-pro-m} prefers cooperatively defeating the monster; \texttt{qvq-max} performs a little better than random.
For chat models, \texttt{gpt-4.1} prefer eating apples alone; \texttt{doubao-1-5-vision-pro} and \texttt{qwen-vl-max} prefer cooperatively defeating the monster; the other models perform a little better than random.
For open-source models, \texttt{Qwen2.5-VL-72B-Instruct} prefers cooperatively defeating the monster; the other models perform a little better than random.

For \textit{Battle of the Colors}, the social behaviors of different models are shown in Fig.~\ref{fig:app:reasoning:battle}, \ref{fig:app:chat:battle}, and \ref{fig:app:open:battle}. Both \texttt{gemini-2.5-pro}, \texttt{gemini-2.5} w/o thinking, \texttt{o3}, and \texttt{ui-tars-1-5} achieve a considerable number of game events, demonstrating strong decision-making capabilities. Beyond this, \texttt{gemini-2.5-pro} and \texttt{gemini-2.5} w/o thinking exhibit a pronounced willingness to cooperate with others, even when such cooperation yields higher rewards for their counterparts.

\subsubsection{Persona}

To investigate the influence of different prompts on the outputs of VLMs, we conducted additional experiments on ``personas'' across three mixed-motive games. Specifically, we selected three reasoning models---\texttt{o3}, \texttt{gemini-2.5-pro}, and \texttt{doubao-1-5-thinking-pro}---and prompted them to adopt either a \textit{selfish persona} (aiming to maximize their own return) or a \textit{cooperative persona} (aiming to maximize the group's total return). The complete prompt instructions are listed below.

\begin{enumerate}[leftmargin=2em]
    \item \textit{Coin Dilemma}
    \begin{itemize}[leftmargin=*,labelsep=0.5em]
        \item \textbf{Common Welfare:} ``Consider the common welfare and collect only coins of your own color to avoid harming the other player's score.''
        \item \textbf{Self Interest:} ``Consider only your own welfare and collect coins of all colors to maximize your score.''
    \end{itemize}

    \item \textit{Monster Hunt}
    \begin{itemize}[leftmargin=*,labelsep=0.5em]
        \item \textbf{Common Welfare:} ``To maximize the common welfare, go to the bottom-left corner and wait for the other player to join you.''  
        ``Stay there and let the monster come to you so that you can defeat the monster together.''
        \item \textbf{Self Interest:} ``To minimize your risk on behalf of your own welfare, do not cooperate with the other player and only collect apples to maximize your own score.''
    \end{itemize}

    \item \textit{Battle of the Colors}
    \begin{itemize}[leftmargin=*,labelsep=0.5em]
        \item \textbf{Common Welfare:} ``To maximize the common welfare, cooperate with the other player and go to the nearest color block to both players (regardless of the color).''
        \item \textbf{Self Interest:} ``To maximize your own welfare, do not cooperate with the other player and only go to the color block of your own color.''
    \end{itemize}
\end{enumerate}

The complete experimental results are presented in Table~\ref{tab:persona}. It can be clearly observed that, across all three mixed-motive games, the choice of prompt significantly shapes the action decisions of VLMs, thereby exerting a substantial impact on the final scores.

\begin{table}[h]
\centering
\scriptsize
\caption{Raw decision-making results under different personas in three mixed-motive games across three reasoning models.}
\label{tab:persona}
\makebox[\textwidth][c]{\setlength{\tabcolsep}{8pt}
\centering
\begin{tabular}{cccccccc}
\toprule
\multirow{2}{*}{Models} & \multicolumn{2}{c}{\textit{Hanabi}} & \multicolumn{2}{c}{\textit{Breakthrough}} & \multicolumn{2}{c}{\textit{Monster Hunt}} \\
& Common Welfare & Self Interest & Common Welfare & Self Interest & Common Welfare & Self Interest \\

\midrule

\texttt{o3} & $11.2_{\pm1.8}$ & $-0.1_{\pm5.9}$ & $54.3_{\pm5.7}$ & $27.5_{\pm5.8}$ & $13.1_{\pm4.4}$ & $0.3_{\pm0.6}$ \\

\texttt{gemini-2.5-pro} & $11.2_{\pm2.1}$ & $-0.7_{\pm5.7}$ & $51.0_{\pm6.6}$ & $26.6_{\pm4.3}$ & $14.7_{\pm4.2}$ & $0.2_{\pm0.4}$ \\

\texttt{doubao-1.5-thinking-pro-m} & $9.9_{\pm4.5}$ & $-3.5_{\pm9.1}$ & $13.1_{\pm41.1}$ & $6.7_{\pm5.6}$ & $5.0_{\pm4.8}$ & $0.1_{\pm1.2}$ \\

\bottomrule
\end{tabular}
\setlength{\tabcolsep}{6pt}
}
\end{table}

\subsection{Human experiments}
\label{app:results:human}

In order to better evaluate the capabilities of VLMs, we additionally conducted a human study. 
A total of 26 college students were participated in our experiments, where they were grouped to perform all decision-making games. 
This provides an effective human baseline for \textsc{VS-Bench}, offering an intuitive reference to measure the performance gap between current VLMs and human-level decision making. 
The complete results of the human study are reported in Table~\ref{tab:human}. 
For a more comprehensive comparison, we also include the best-performing reasoning model \texttt{o3}, the chat model \texttt{gemini-2.5} w/o thinking, and the open-source model \texttt{Qwen2.5-VL-72B-Ins}. 
The results clearly show that, at the current stage, humans still substantially outperform even the strongest reasoning models. 
Specifically, the normalized score of human participants is nearly twice as high as that of \texttt{o3}.

\begin{table}[h]
\centering
\scriptsize
\caption{Decision-making results for the human baseline. 
For comparison, we also report the performance of the best reasoning, chat, and open-source models. 
All values are normalized scores. For each environment, the \colorbox{first}{best} result is highlighted in green.}
\label{tab:human}
\makebox[\textwidth][c]{\setlength{\tabcolsep}{6.5pt}
\begin{tabular}{cccccccccccc}
\toprule
\multirow{2}{*}{Models}
& \multirow{2}{*}{Overall} & \multicolumn{3}{c}{Cooperative} & \multicolumn{4}{c}{Comptitive} & \multicolumn{3}{c}{Mixed-Motive} \\
& & \textit{Hanabi} & \textit{Overcooked} & \textit{KAZ} & \textit{Board} 
& \textit{Poker}    & \textit{Pong} & \textit{MPE} & \textit{Dilemma} & \textit{Hunt} & \textit{Battle} \\
\midrule
Oracle & $100.0$ & $100.0$ & $100.0$ & $100.0$ & $100.0$ & $100.0$ & $100.0$ & $100.0$ & $100.0$ & $100.0$ & $100.0$ \\
\midrule

Human & \cellcolor{first} $62.7$ & \cellcolor{first} $72.9$ & \cellcolor{first} $87.9$ & \cellcolor{first} $84.2$ & \cellcolor{first} $100.0$ &   $7.6$ & \cellcolor{first} $66.5$ &   $22.4$ & \cellcolor{first} $38.1$ & \cellcolor{first} $66.0$ & \cellcolor{first} $81.7$ \\

\midrule

\texttt{o3} &   $31.4$ &   $55.8$ &   $15.6$ &   $29.5$ &   $65.0$ & \cellcolor{first} $61.8$ &   $8.6$ & \cellcolor{first} $37.2$ &  $-0.4$ &   $24.0$ &   $16.7$ \\

\texttt{gpt-4.1} & $4.8$ &   $0.0$ &  $-0.5$ &  $-5.3$ &   $0.0$ &  $-7.1$ & $0.2$ & $31.5$ &   $17.8$ & $11.2$ & $0.5$ \\

\texttt{Qwen2.5-VL-72B-Ins.} & $3.0$ & $0.8$ &  $-0.5$ &  $-5.3$ &   $0.0$ &  $-3.1$ &  $-0.8$ & $19.6$ &   $0.0$ & $19.6$ &  $-0.5$ \\

\midrule
Random & $0.0$ & $0.0$ & $0.0$ & $0.0$ & $0.0$ & $0.0$ & $0.0$ & $0.0$ & $0.0$ & $0.0$ & $0.0$ \\
\bottomrule
\end{tabular}
}
\end{table}

\subsection{Beyond the two-player setting}
\label{app:results:multi}

Although the main body of our work primarily emphasizes the two-player setting---because it is the most canonical and widely studied case in multi-agent research, serving as a foundational step toward more general $n$-agent evaluation---our benchmark is in fact a general framework that can be readily applied to $n$-player games with $n > 2$. To substantiate this claim, we further conduct experiments on two additional environments: (1) three-player \textit{Hanabi} and (2) the three-player \textit{Coin Dilemma}. We evaluate several representative models and the corresponding results are reported in Table~\ref{tab:multi}.

\begin{table}[t]
\centering
\scriptsize
\caption{Decision-making results of 7 representative VLMs on multi-agent games, including three-player \textit{Hanabi} and the three-player \textit{Coin Dilemma}.}
\label{tab:multi}
\makebox[\textwidth][c]{\setlength{\tabcolsep}{12pt}
\centering
\begin{tabular}{cccc}
\toprule
\multirow{2}{*}{Models} & \multicolumn{2}{c}{Three-Player \textit{Hanabi}} & \multirow{2}{*}{Three-Player \textit{Dilemma}} \\
 & standard & firework & \\

\midrule

\texttt{o3} & $14.3_{\pm2.19}$ & $14.3_{\pm2.19}$ & $-8.7_{\pm4.45}$ \\

\texttt{doubao-1.5-thinking-pro-m} & $8.7_{\pm6.14}$ & $10.4_{\pm3.95}$ & $1.07_{\pm10.37}$ \\

\midrule

\texttt{gpt-4.1} & $0_{\pm0}$ & $3.8_{\pm1.40}$ & $5.34_{\pm8.87}$ \\

\texttt{gemini-2.5-flash} & $0_{\pm0}$ & $2.3_{\pm1.41}$ & $-3.56_{\pm10.89}$ \\

\texttt{doubao-1.5-vision-pro} & $0_{\pm0}$ & $4.3_{\pm1.41}$ & $-2.14_{\pm4.98}$ \\

\midrule

\texttt{Qwen2.5-VL-72B-Instruct} & $0.3_{\pm0.48}$ & $0.4_{\pm0.52}$ & $-4.63_{\pm8.90}$ \\

\texttt{Llama-3.2-90B-Vision-Instruct} & $0_{\pm0}$ & $1_{\pm1.05}$ & $-1.07_{\pm5.10}$ \\

\bottomrule
\end{tabular}
\setlength{\tabcolsep}{6pt}
}
\end{table}

\section{Failure case examples}
\label{app:fail}
\begin{figure}[t]
  \begin{minipage}[t]{0.4\textwidth}
    \textbf{Visual observation:}\\[4pt]
    
    \centering
    \includegraphics[width=\linewidth]{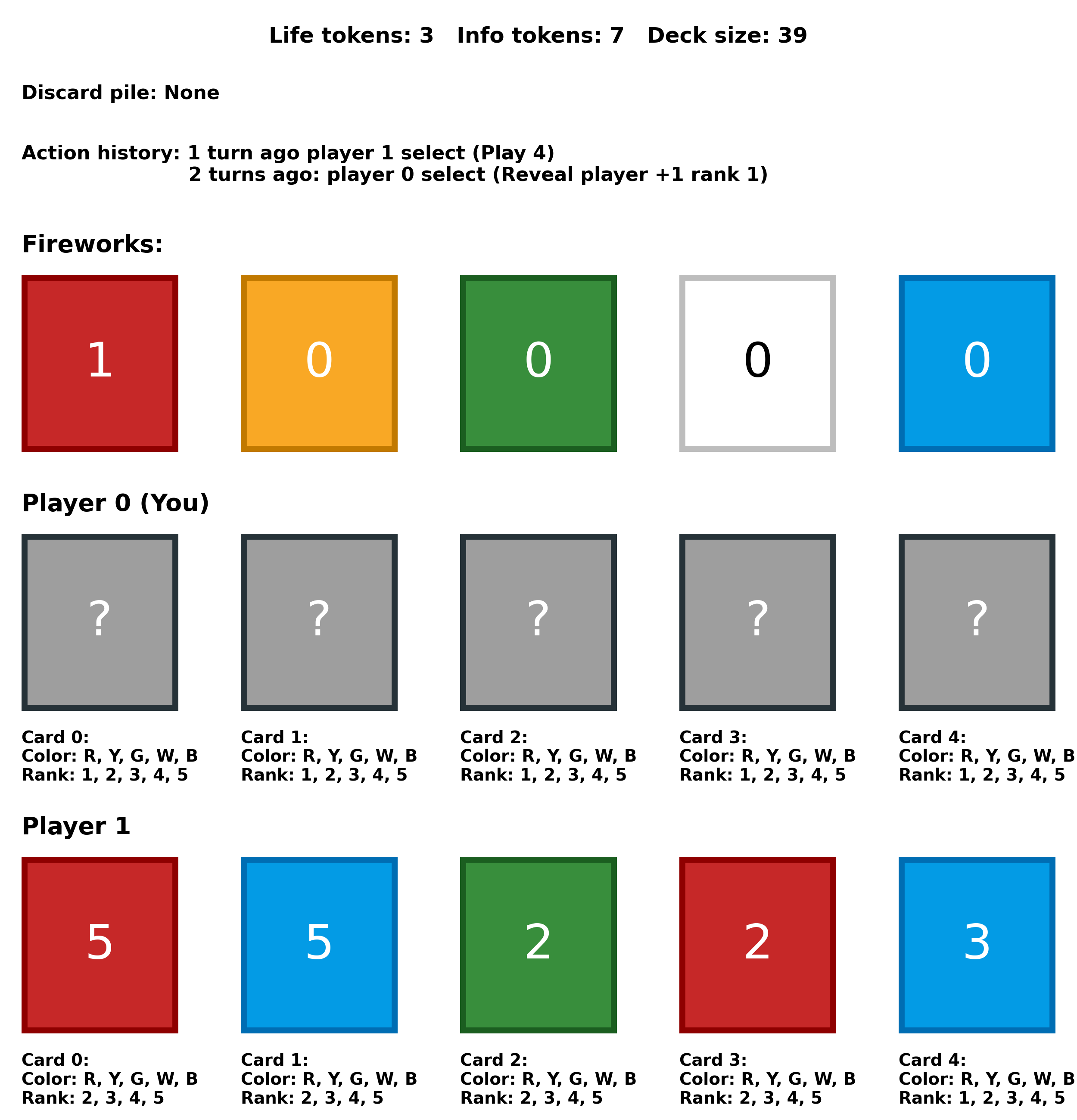}
  \end{minipage}\hfill
  \begin{minipage}[t]{0.58\textwidth}
    \textbf{VLM response:}
\begin{tcolorbox}[boxsep=1pt]
\begin{Verbatim}[breaklines=true,breakanywhere=true,breaksymbol={},breaksymbolleft={},fontsize=\small]
...
Now, Player 1's possible actions: play, discard, or reveal. Let's consider playing first. The fireworks for Red have 1, so the next needed is Red 2. Player 1 has R2 (card 3), which is the next in Red. Playing that would be valid. Green needs 1 first, but Player 1 has G2, which can't be played yet since Green's 1 isn't down. The 5s (R5, B5) can't be played until their sequences are up to 4. B3 is Blue 3, but Blue needs 1 first. So the valid play here is R2 (card 3) since Red is at 1, and R2 is the next. 
...
      \end{Verbatim}
    \end{tcolorbox}
  \end{minipage}
  \caption{Failure case example of strategic reasoning in \textit{Hanabi}.}
  \label{fig:app:fail:hanabi}
\end{figure}

\begin{figure}[t]
\begin{minipage}[t]{\textwidth}
\textbf{Visual observation:}
\end{minipage}
\begin{minipage}[t]{0.24\textwidth}
\centering
\includegraphics[trim=0 0 1100 0,clip,width=\linewidth]{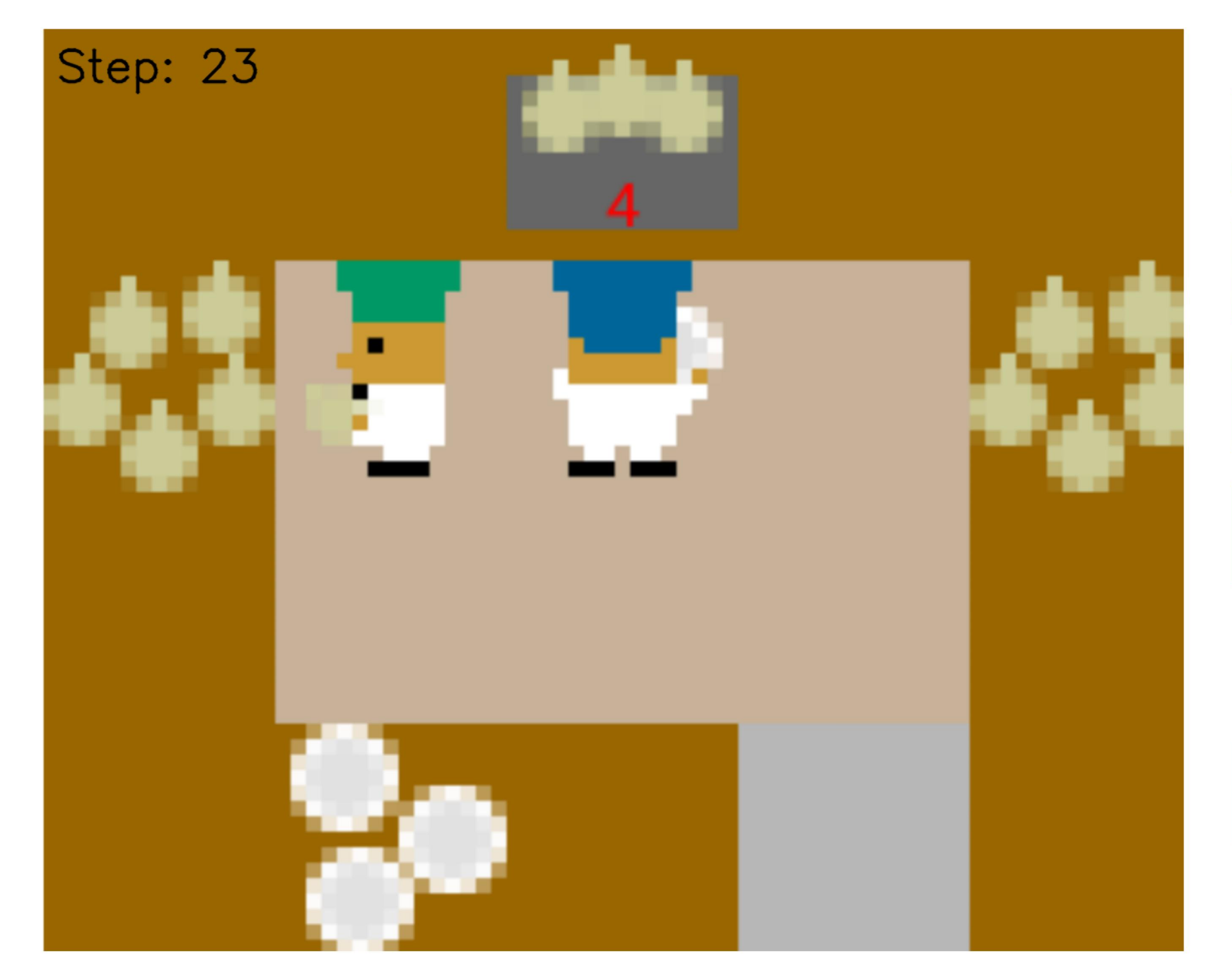}
frame 1
\end{minipage}
\hfill
\begin{minipage}[t]{0.24\textwidth}
\centering
\includegraphics[trim=0 0 1100 0,clip,width=\linewidth]{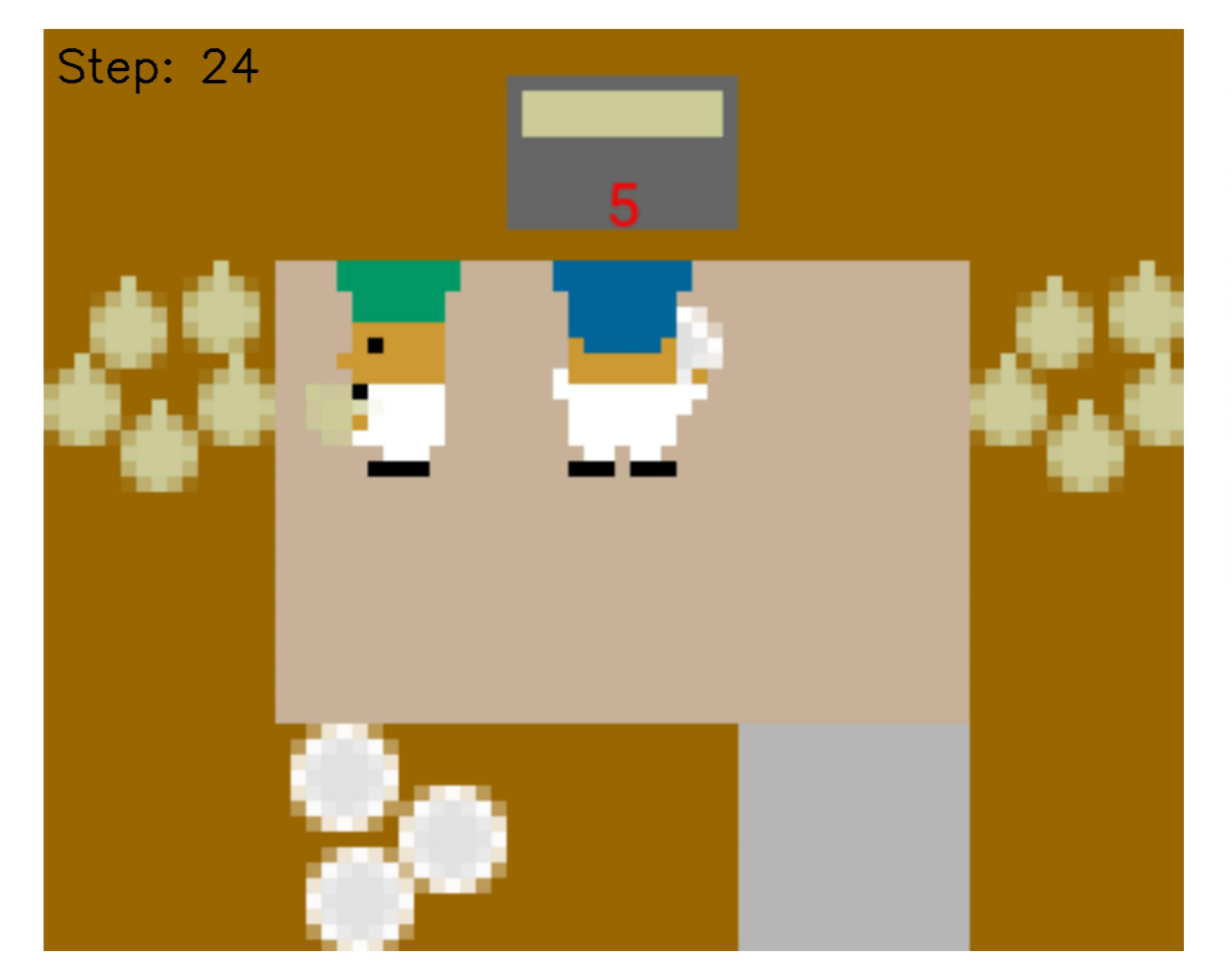}
frame 2
\end{minipage}
\hfill
\begin{minipage}[t]{0.24\textwidth}
\centering
\includegraphics[trim=0 0 1100 0,clip,width=\linewidth]{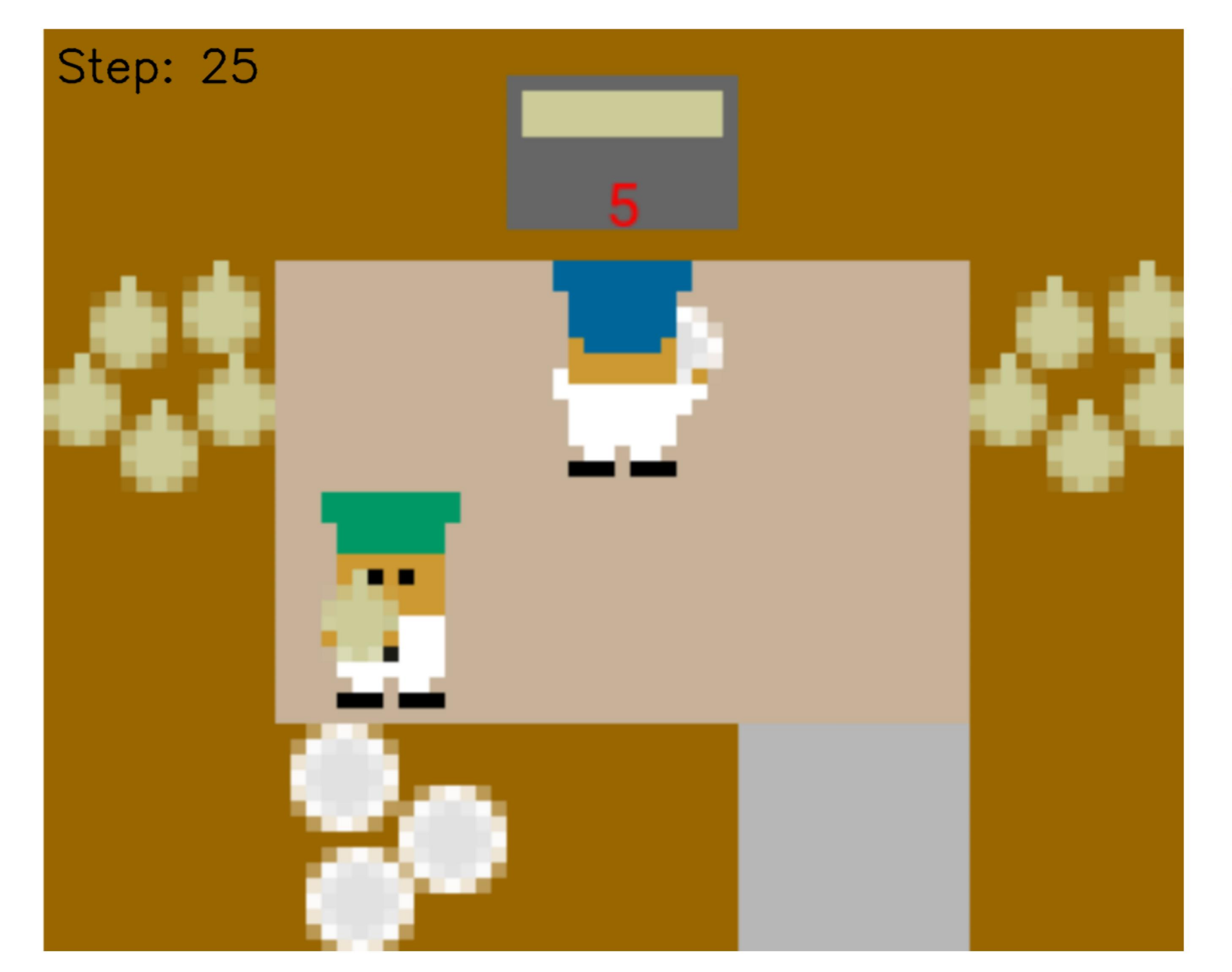}
frame 3
\end{minipage}
\hfill
\begin{minipage}[t]{0.24\textwidth}
\centering
\includegraphics[trim=0 0 1100 0,clip,width=\linewidth]{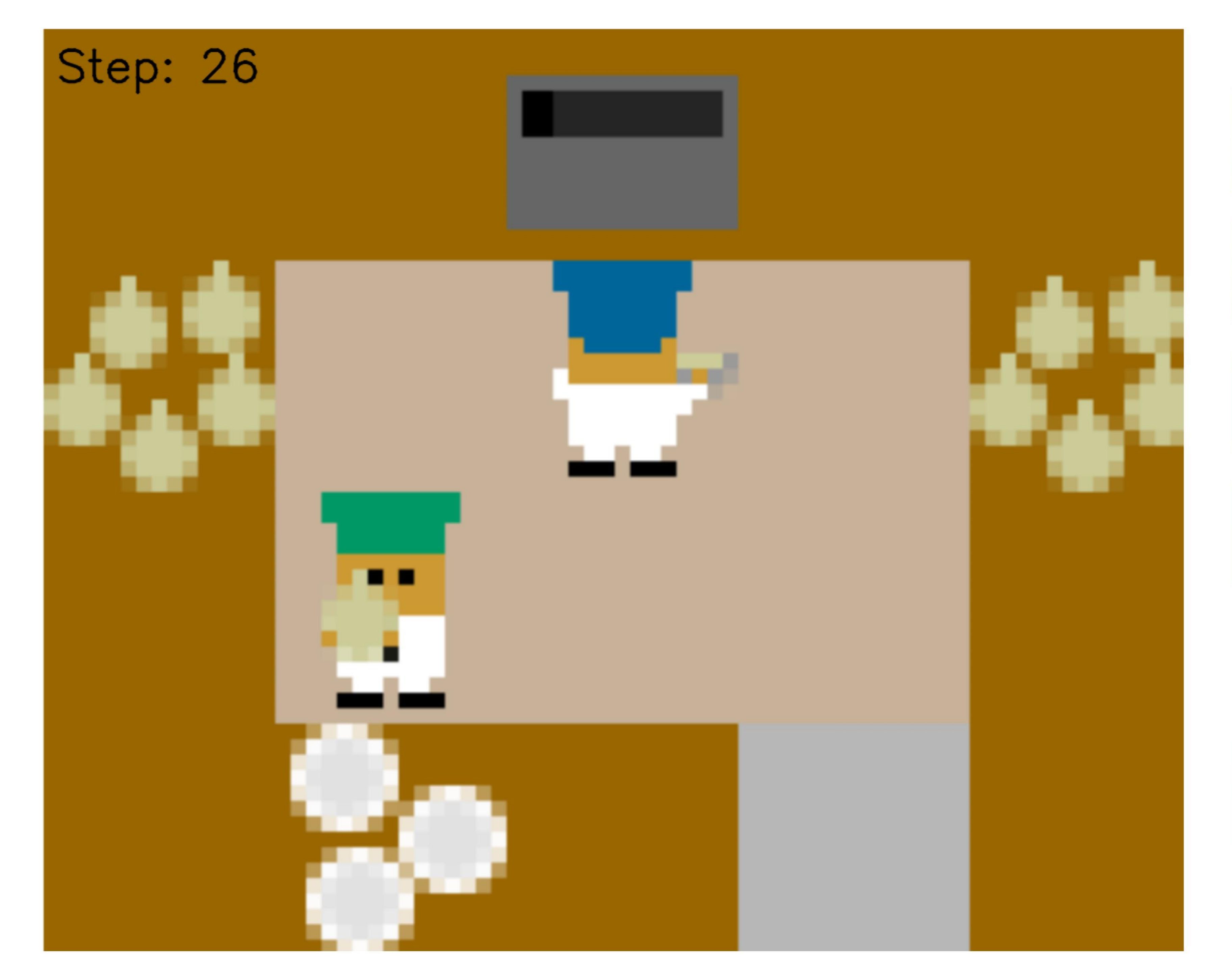}
frame 4
\vspace{2mm}
\end{minipage}
\begin{minipage}[t]{\textwidth}
\textbf{VLM response:}
\begin{tcolorbox}
\begin{Verbatim}[breaklines=true, breakanywhere=true,breaksymbol={},breaksymbolleft={},fontsize=\small]
...
Actually, let me look more carefully at what chef_0 is doing. In frames 23, 24, and 25, chef_0 is standing in the same position and hasn't moved. In frame 26, the pot shows a black strip, indicating the soup is fully cooked. 
According to the rules, after the soup is cooked for 5 timesteps, the next step is to plate it using a dish. Since chef_0 is directly in front of the pot, the most likely action would be <INTERACT> to take the soup from the pot. 
However, this would require chef_0 to have a dish already. Since I don't clearly see chef_0 holding a dish, they would first need to get one.
Given this analysis, I believe chef_0 will most likely choose <INTERACT> to interact with the pot.
...
\end{Verbatim}
\end{tcolorbox}
\end{minipage}
\caption{Failure case example of reasoning in \textit{Overcooked}.}
\label{fig:app:fail:overcooked-reasoning}
\end{figure}

\subsection{Strategic reasoning}
We present three illustrative failure cases in strategic reasoning from different game environments.

In \textit{Hanabi}, VLM agents only observe the other agent’s hands but not their own hands, creating a distinct information asymmetry.
An example with visual observation and the VLM’s response is shown in Fig.~\ref{fig:app:fail:hanabi}.
When predicting the other agent’s next move, VLMs often immediately identify which visible card could yield points and assume the other agent will play it. This prediction, however, overlooks a critical fact: from the other agent’s perspective, their information about that card is incomplete, so they would not risk losing a life token by playing it prematurely. Instead, they would wait for more definitive clues before making that play.

In \textit{Overcooked}, accurately predicting the next action of the other agent requires robust visual perception capabilities and a thorough comprehension of image information.
An example with visual observation and the VLM’s response is shown in Fig.~\ref{fig:app:fail:overcooked-reasoning}.
In this case, the VLM should predict the blue chef's action from the green chef's perspective. Despite correctly identifying from historical frames that the soup was cooked, the VLM's visual perception is inadequate: it fails to recognize that the blue chef was already holding the soup, plated in a dish. 
In addition, the VLM overlooks a critical game rule, mistakenly believing that soup could be collected from the pot before a dish was acquired. As a result of these combined deficiencies in visual understanding and rule application, the VLM predicts the erroneous action <INTERACT>.

\begin{figure}[t]
\begin{minipage}[t]{0.5\textwidth}
\textbf{Visual observation:}\\[6pt]
\includegraphics[trim=30 0 0 0,clip,width=\linewidth]{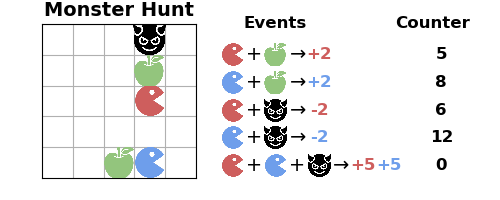}
\end{minipage}
\hfill
\begin{minipage}[t]{0.48\textwidth}
\textbf{VLM response:}
\begin{tcolorbox}[boxsep=1pt]
\begin{Verbatim}[breaklines=true,breakanywhere=true,breaksymbol={},breaksymbolleft={},fontsize=\small]
...
If the red player moves <UP>, it collects an apple, and gets +2 points. The red player's best move to maximize their own score is to go up.
...
      \end{Verbatim}
    \end{tcolorbox}
  \end{minipage}
  \caption{Failure case example of strategic reasoning in \textit{Monster Hunt}.}
  \label{fig:app:fail:hunt}
\end{figure}

In \textit{Monster Hunt}, in order to predict the next action of the other player, it is important to track the positions and movements of all game elements, rather than focusing on only part of the game mechanics and make short-sighted predictions. 
An example with visual observation and the VLM’s response is shown in Fig.~\ref{fig:app:fail:hunt}.
The VLM notices the apple is just one step above the red player, but does not pay attention to the monster on top of the apple. The VLM predicts the red player to go up to collect the apple, while the red player actually goes down to flee from the monster.

\begin{figure}[t]
  \begin{minipage}[t]{0.5\textwidth}
    \vspace{0pt}
    \textbf{Visual observation:}\\[-5pt]
    
    \includegraphics[width=\linewidth]{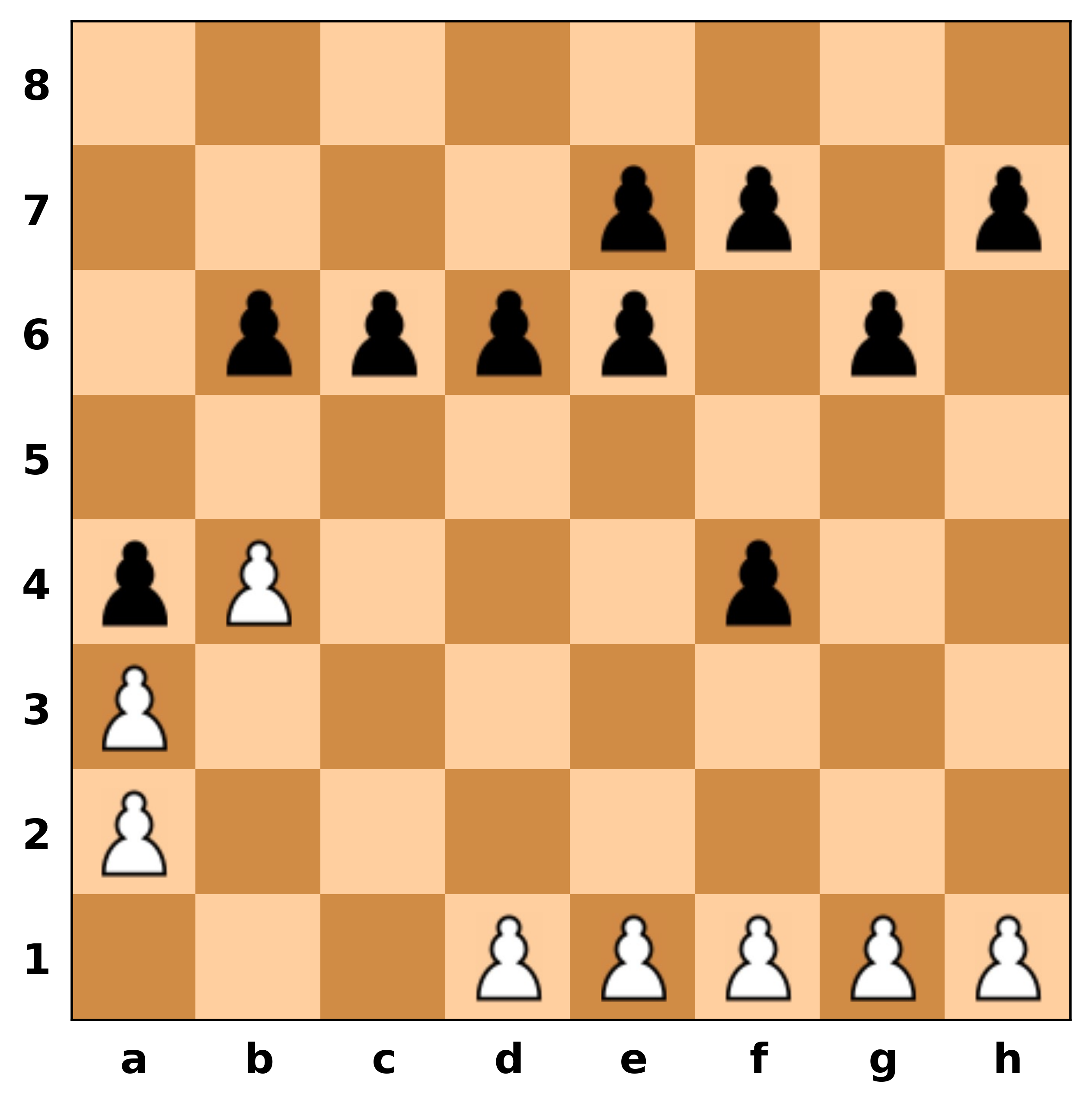}
  \end{minipage}\hfill
  \begin{minipage}[t]{0.5\textwidth}
    \vspace{0pt}
    \textbf{VLM response:}
\begin{tcolorbox}[boxsep=1pt]
\begin{Verbatim}[breaklines=true,breakanywhere=true,breaksymbol={},breaksymbolleft={},fontsize=\small]
...
Now, considering optimal play, White would want to advance their pieces towards Black's side, possibly creating threats or blocking Black's moves. The b4 piece is already advanced. Moving it to c5 (b4c5) would bring it closer to the center and maybe set up a capture or block. 
Looking at the starting row pieces (d1 to h1), moving them straight forward (like d1d2, e1e2, f1f2, g1g2, h1h2) or diagonally. Straight moves are safe but maybe less aggressive. Diagonal moves could potentially set up captures later.
...
      \end{Verbatim}
    \end{tcolorbox}
  \end{minipage}
  \caption{Failure case example of decision-making in \textit{Breakthrough}.}
  \label{fig:app:fail:breakthrough}
\end{figure}

\begin{figure}[t]
\begin{minipage}[t]{0.24\textwidth}
\centering
\textbf{Frame 1}
\includegraphics[trim=0 0 1150 0,clip,width=\linewidth]{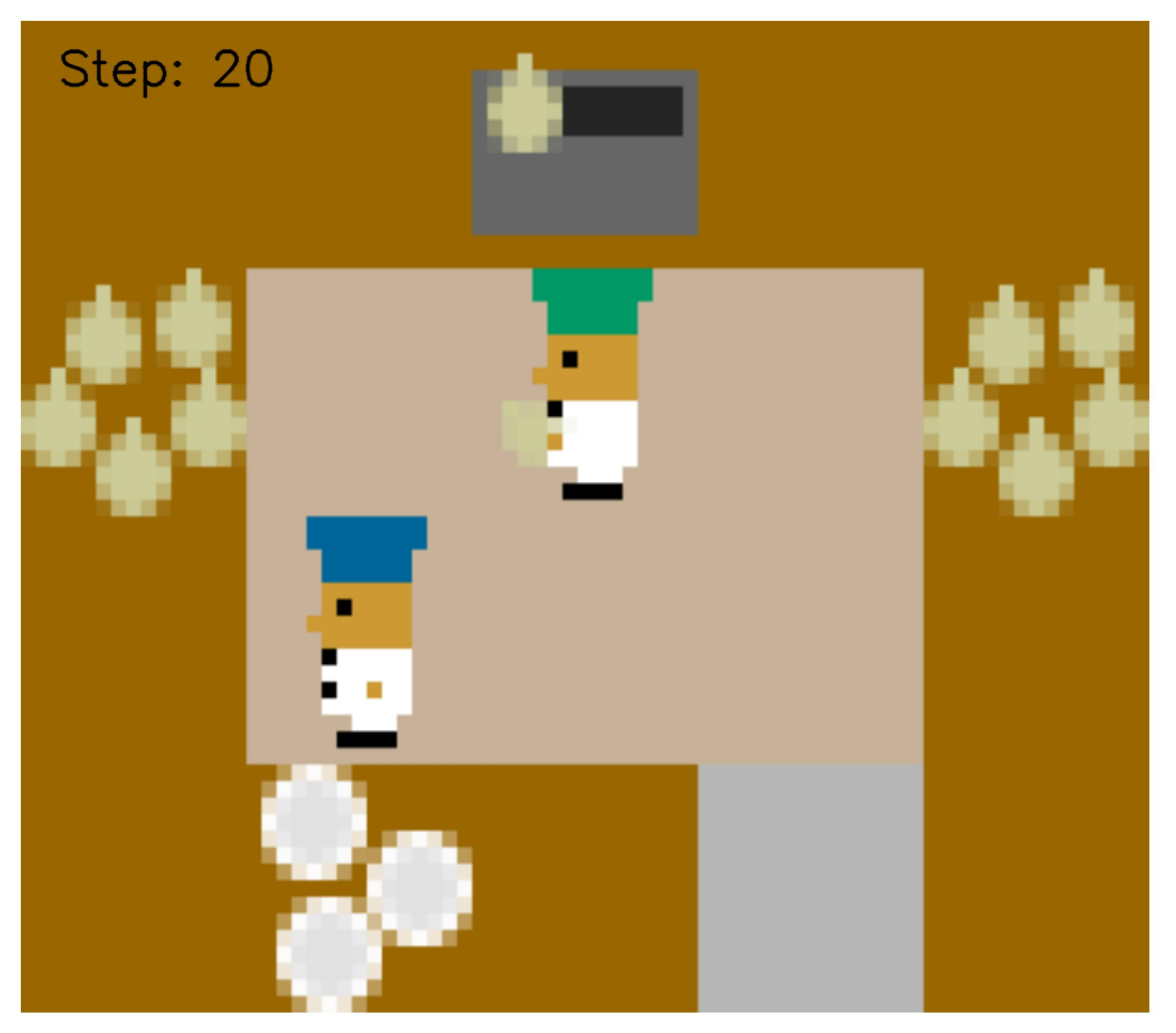}
\end{minipage}
\hfill
\begin{minipage}[t]{0.24\textwidth}
\centering
\textbf{Frame 2}
\includegraphics[trim=0 0 1150 0,clip,width=\linewidth]{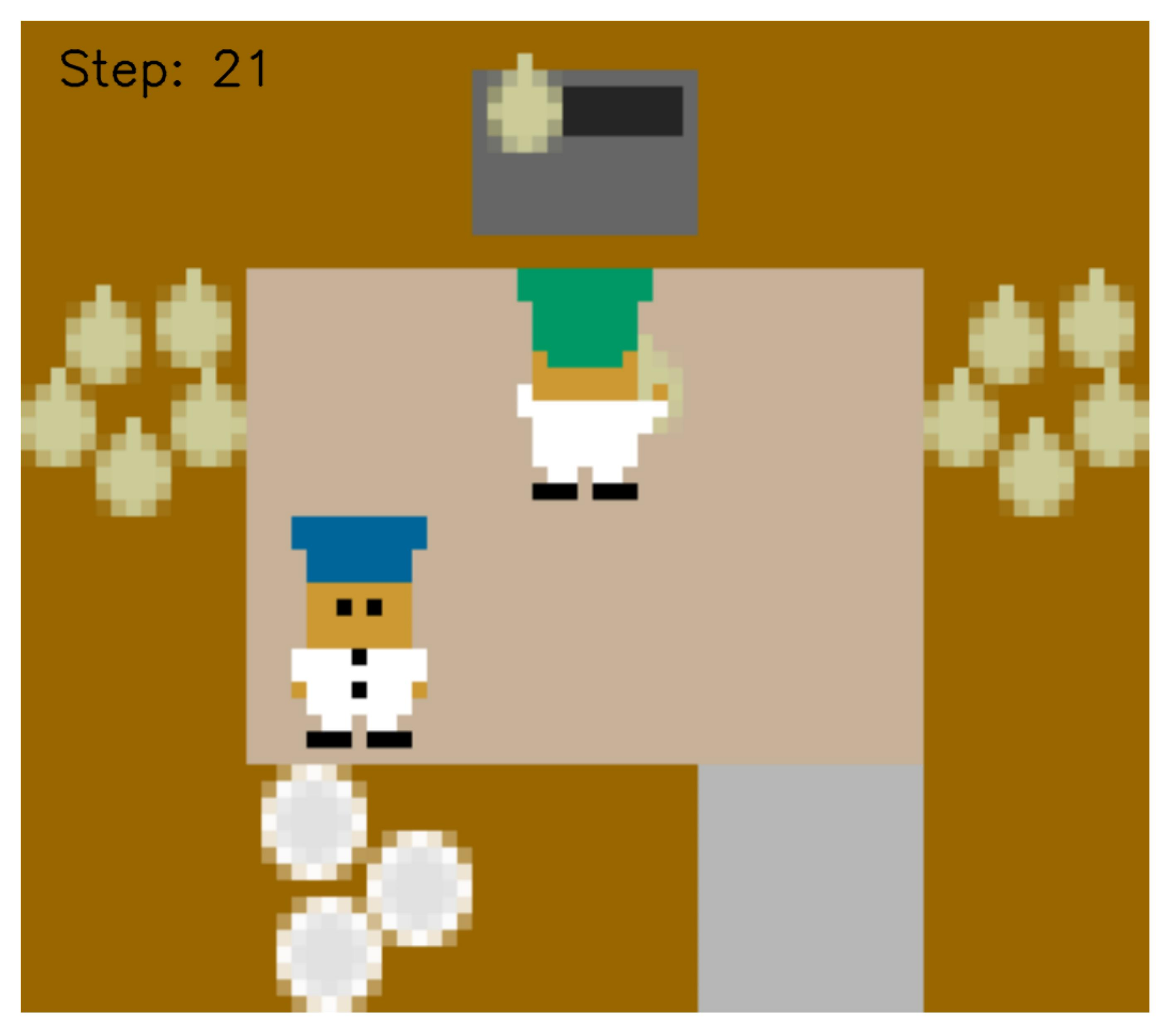}
\end{minipage}
\hfill
\begin{minipage}[t]{0.24\textwidth}
\centering
\textbf{Frame 3}
\includegraphics[trim=0 0 1150 0,clip,width=\linewidth]{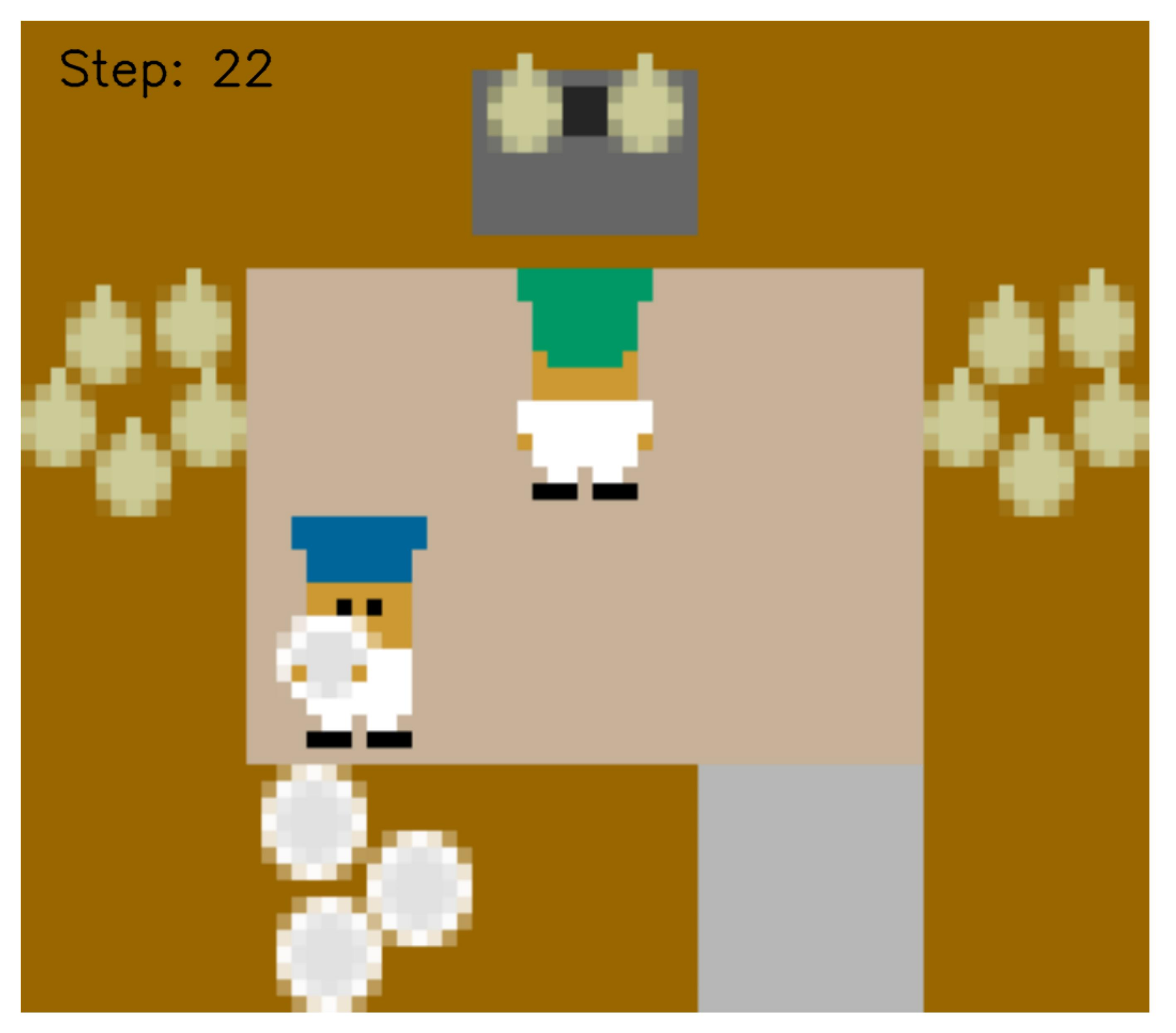}
\end{minipage}
\hfill
\begin{minipage}[t]{0.24\textwidth}
\centering
\textbf{Frame 1}
\includegraphics[trim=0 0 1150 0,clip,width=\linewidth]{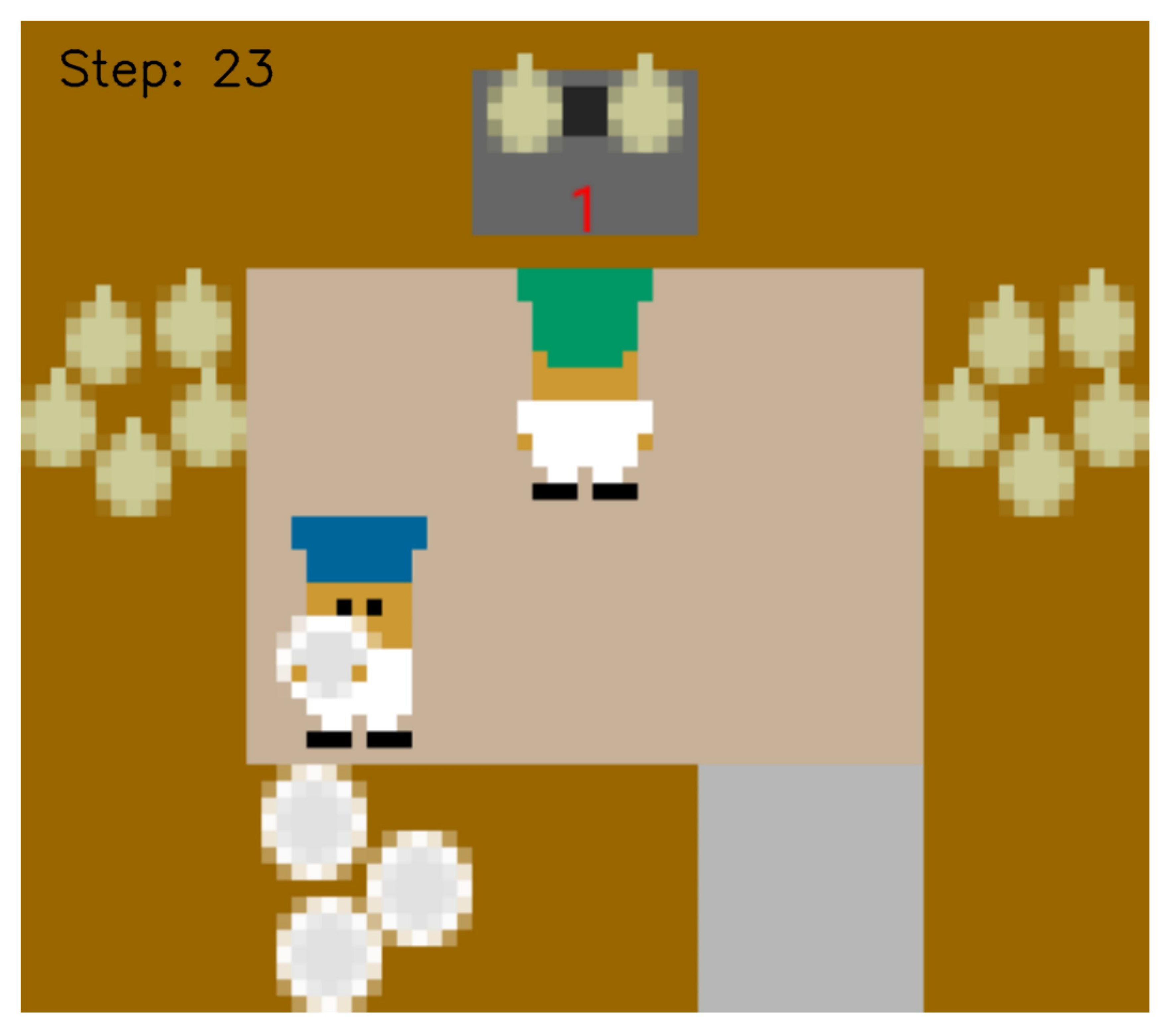}
\end{minipage}
\caption{Failure case example of decision-making in \textit{Overcooked}.}
\label{fig:app:fail:overcooked-decision}
\end{figure}

\begin{figure}[t]
\centering
\begin{minipage}[t]{0.2\textwidth}
\textbf{Frame 1}
\centering
\includegraphics[width=\linewidth]{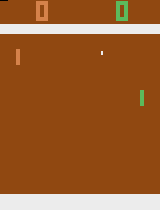}
\end{minipage}\hfill
\begin{minipage}[t]{0.2\textwidth}
\textbf{Frame 2}
\centering
\includegraphics[width=\linewidth]{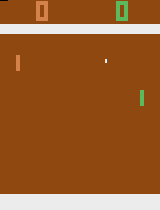}
\end{minipage}\hfill
\begin{minipage}[t]{0.2\textwidth}
\textbf{Frame 3}
\centering
\includegraphics[width=\linewidth]{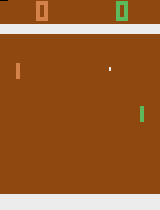}
\end{minipage}\hfill
\begin{minipage}[t]{0.2\textwidth}
\textbf{Frame 4}
\centering
\includegraphics[width=\linewidth]{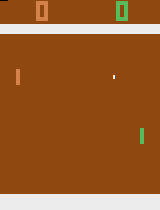}
\end{minipage}\hfill
\caption{Failure case example of decision-making in \textit{Atari Pong}.}
\label{fig:app:fail:pong}
\end{figure}

\subsection{Decision‑making}
We also present three illustrative failure cases in decision‑making from different game environments.

In \emph{Breakthrough}, a strong strategy must excel at both offense and defense. However, VLMs occasionally overlook situations in which the opponent’s pieces are on the verge of victory and require immediate defensive intervention. Instead, they persistently advance their most distant piece by one step, thereby neglecting the critical balance between attack and defense.
An example with visual observation and the VLM’s response is shown in Fig.~\ref{fig:app:fail:breakthrough}.

In \textit{Overcooked}, to correctly prepare a soup with three onions, the VLM agent needs to accurately manage the number of onions. A common failure mode of VLMs is that they often neglect these precise quantity requirements. As illustrated in Fig.~\ref{fig:app:fail:overcooked-decision}, the green chef initiates cooking by performing an <INTERACT> operation after adding only the second onion for this recipe. Concurrently, the blue chef prematurely begins to retrieve a plate. As a consequence, the dish, even when served, fails to earn the delivery reward because the ingredient count requirement is not met.

In \textit{Atari Pong}, in order to successfully intercept the ball, the VLM agent needs to accurately predict the landing point of the ball. A common failure mode of the VLMs is that they may over-adjust their paddle. As illustrated in Fig.~\ref{fig:app:fail:pong}, the VLM on the right detects that the ball is moving downward and then move the paddle down, but misses the landing point due to over-aggressive adjustment.

\section{Additional environments}
\label{app:extending}
To provide a more comprehensive evaluation, we extend \textsc{VS-Bench} with four additional environments. Specifically, we introduce two simpler games, a cooperative game \textit{Tiny-Hanabi} and a competitive game \textit{Tic-Tac-Toe}, which serve as lightweight testbeds. In addition, since our benchmark is a general framework that can be readily applied to $n$-player games with $n > 2$, we further include two three-player extensions, a cooperative game given by the three-player version of \textit{Hanabi} and a mixed-motive game given by the three-player version of \textit{Coin Dilemma}, which naturally extend the original two-player settings.

\begin{figure}[t]
\centering
\hfill
    \begin{minipage}[b]{0.45\textwidth}
    \centering
    \includegraphics[height=10cm]{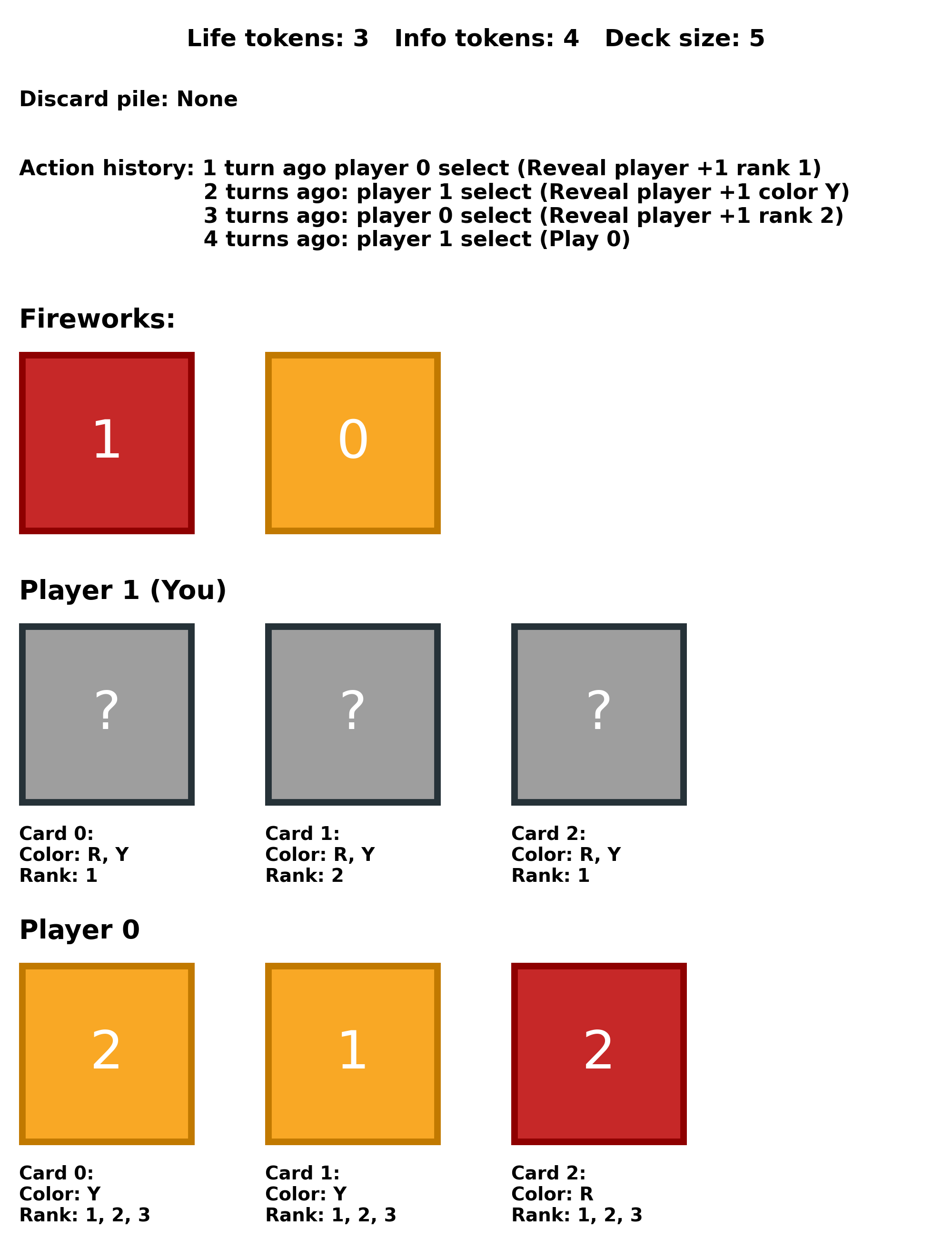}
    \caption{\textit{Tiny Hanabi}.}
    \label{fig:app:tiny-hanabi}
    \end{minipage}
\hfill
    \begin{minipage}[b]{0.45\textwidth}
    \centering
    \includegraphics[height=10cm]{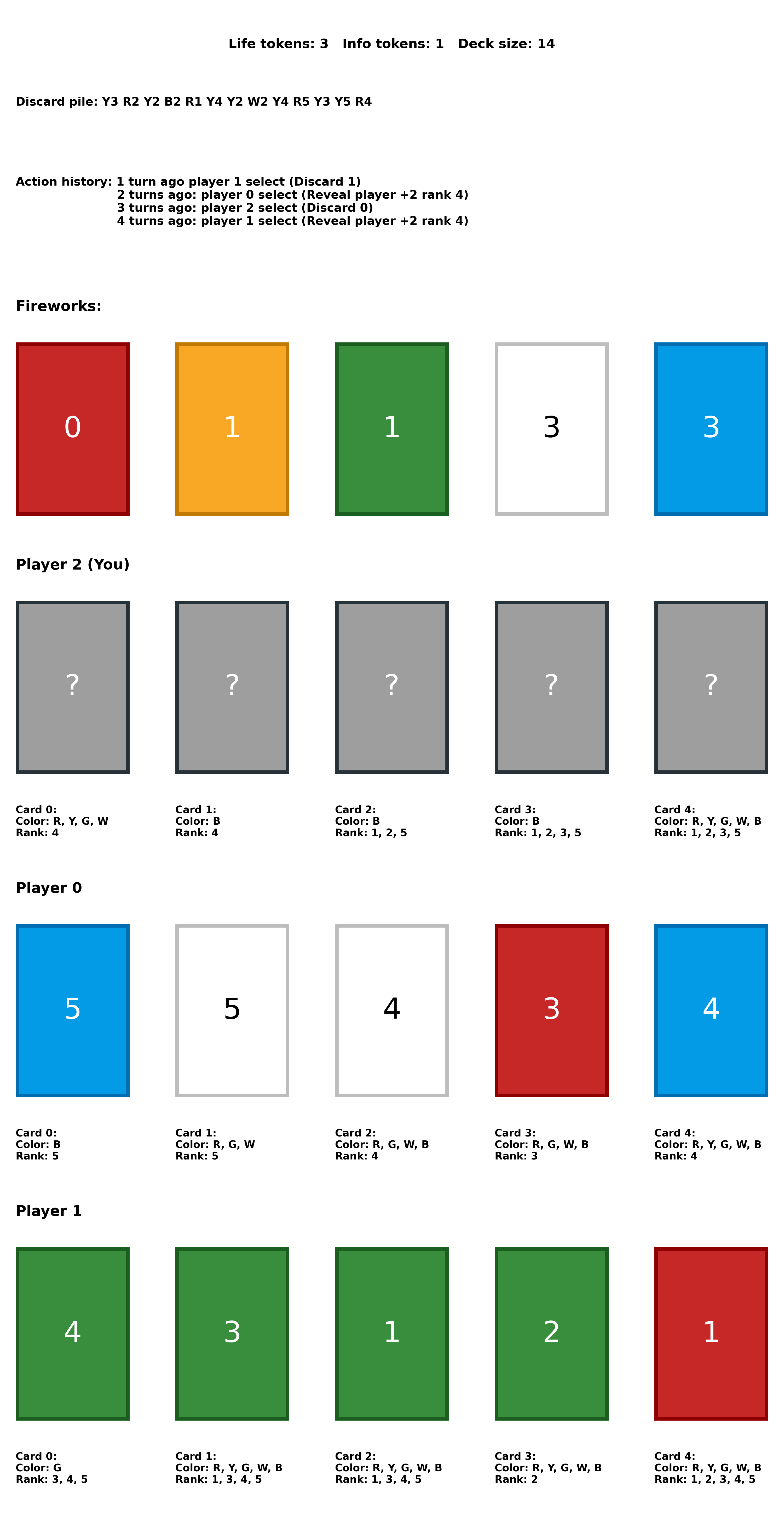}
    \caption{Three-player \textit{Hanabi}.}
    \label{fig:app:3-hanabi}
    \end{minipage}    
\hfill
\end{figure}

\begin{figure}[t]
\centering
\hfill
    \begin{minipage}[b]{0.28\textwidth}
    \centering
    \includegraphics[height=4cm]{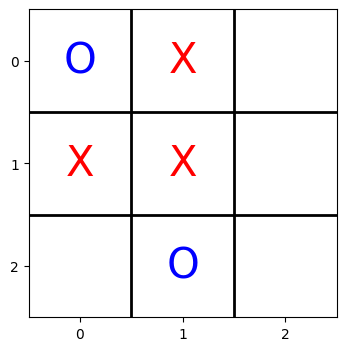}
    \caption{\textit{Tic-Tac-Toe}.}
    \label{fig:app:tic-tac-toe}
    \end{minipage}
\hfill
    \begin{minipage}[b]{0.7\textwidth}
    \centering
    \includegraphics[height=4cm]{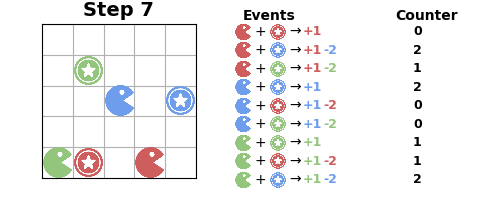}
    \caption{Three-player \textit{Coin Dilemma}.}
    \label{fig:app:3-dilemma}
    \end{minipage}
\hfill
\end{figure}

\begin{enumerate}[leftmargin=2em]

    \item \textbf{\textit{Tiny Hanabi}} is a simplified version of \textit{Hanabi} with only two colors, three ranks, and a hand size of three cards. Detailed rules are provided in Appendix~\ref{app:hanabi_obs}. An example visual observation is shown in Fig.~\ref{fig:app:tiny-hanabi}.

    \item \textbf{\textit{Tic-Tac-Toe}} is a competitive two-player board game played on a $3 \times 3$ grid. The first player to align three marks horizontally, vertically, or diagonally wins. This game has a relatively small state space, and its optimal strategies are known to guarantee a draw under perfect play. We use it as a preliminary testbed for evaluating VLMs in competitive multi-agent environments. An example observation is shown in Fig.~\ref{fig:app:tic-tac-toe}.

    \item \textbf{Three-player \textit{Hanabi}} extends the original \textit{Hanabi} to three players with other rules unchanged. Detailed rules are provided in Appendix~\ref{app:hanabi_obs}. An example observation is shown in Fig.~\ref{fig:app:3-hanabi}.
    
    \item \textbf{Three-player \textit{Coin Dilemma}} extends the original game to a three-player setting with other rules unchanged. Detailed rules are provided in Appendix~\ref{app:dilemma_obs}. An example observation is shown in Fig.~\ref{fig:app:3-dilemma}.

\end{enumerate}

\section{Environment details}
\label{app:envs}
In this section, we first introduce the four core abilities that \textsc{VS-Bench} aims to evaluate in multi-agent settings. We then provide detailed descriptions of the ten games included in \textsc{VS-Bench}, covering their rules, gameplay dynamics, and relevant implementation details.

\subsection{Ability Definition}
In \textsc{VS-Bench}, we identify four core abilities that play crucial roles in multi-agent interaction:

\begin{enumerate}[leftmargin=2em]
    \item \textbf{Spatial Reasoning.} This ability emphasizes a model’s capacity for spatial perception, namely, whether it can interpret the current environment based on visual inputs and the rules of the game. It serves as the foundation for VLMs in interactions. In particular, continuous-space games such as \textit{Pong} and \textit{KAZ} are highly dependent on visual understanding.

    \item \textbf{Theory of Mind (ToM).} This ability evaluates whether a VLM can accurately infer an opponent’s intentions and subsequently make better action choices. For example, in \textit{Hanabi}, players must reason about their teammates’ true intentions based on very limited hints.

    \item \textbf{Long-Term Planning.} This ability reflects a model’s competence in long-horizon reasoning, particularly its capacity to anticipate future actions at each decision point in order to maximize cumulative rewards. Both \textit{Hanabi} and \textit{Breakthrough} exemplify this demand for strong planning skills, with the former requiring players to operate under a limited set of cards, while the latter embodies the general characteristics of board games.

    \item \textbf{Team Collaboration.} This ability evaluates how effectively models coordinate with one another. It is not relevant in purely competitive \textit{games}, but becomes crucial in cooperative settings, where improved collaboration directly translates into higher returns. For instance, \textit{Hanabi} presents a considerable challenge by demanding close cooperation between players.
\end{enumerate}

\subsection{Hanabi}
\label{app:hanabi_obs}

\textit{Hanabi}~\citep{bard2020hanabi} is a partially observable card game where players can observe others' cards but not their own. 
Each card has a color and a rank that can only be revealed through hint actions at the cost of an information token. 
To succeed, agents must coordinate to play cards in rank order for five colors.
We consider the two-player full game, which is widely used for research on theory of mind, zero-shot coordination, and ad-hoc teamplay~\citep{hu2020other,hu2023language}.

\textbf{Game rules.}

\begin{enumerate}[leftmargin=2em]
    \item \textit{Hanabi} is a cooperative card game for 2 players.
    \item The deck consists of 5 colors: R (Red), Y (Yellow), G (Green), W (White), B(Blue), with ranks ranging from 1 to 5. Each color contains 10 cards: three of rank 1, two each of rank 2 through 4, and one of rank 5, for a total of 50 cards.
    \item Each player holds 5 cards in hand.
    \item There are 8 Info tokens (used to give hints) and 3 Life tokens (penalties for misplays).
    \item As in blind man's bluff, players can see each other's cards but they cannot see their own. Play proceeds around the table; each turn, a player must take one of the following actions: <PLAY>, <DISCARD i>, <REVEAL color c>, <REVEAL rank r>.
    \item The game ends immediately when either all Life tokens are used up, resulting in a game loss with a score of 0, or when all 5s have been successfully played, resulting in a game win with a score of 25. Otherwise, the game continues until the deck runs out and one final round is completed. At the end of the game, the final score is calculated as the sum of the highest card played in each suit, up to a maximum of 25 points.
\end{enumerate}

\textbf{Visual observation.}
An example is shown in Fig.~\ref{fig:app:hanabi}. The visual observation has four parts:
\begin{enumerate}[leftmargin=2em]
  \item Basic Information: counts of remaining life tokens, information tokens, and deck size.
  \item Historical Information: all discarded cards, together with each player’s two most recent actions.
  \item Fireworks: current progress of each color’s fireworks stack, indicating the highest played rank.
  \item Players’ Hands: each player’s hand, with annotations beneath each card showing all possible colors and ranks deduced from received hints.
\end{enumerate}

\subsection{Overcooked}
\label{app:overcooked_obs}

\textit{Overcooked}~\citep{ghost2016overcooked} is a popular video game where two chefs cooperate to cook and serve dishes in a kitchen.
Each dish requires multiple operations like navigating, chopping, cooking, plating, and delivering, which are difficult to coordinate even for human players.
Our implementation is based on Overcooked-AI~\citep{carroll2019utility}, a well-known environment for zero-shot coordination and human-AI interactions~\citep{strouse2021collaborating,zhang2024proagent}.

\textbf{Game rules.}

\begin{enumerate}[leftmargin=2em]
    \item Overcooked is a cooperative game where two chefs collaborate to cook and serve soups in 50 timesteps.
    \item The chefs can move in the available area and cannot move to the counter.
    \item The chefs can interact with the object on the tile that they are facing.
    \item A soup is cooked in the following steps:
    \begin{enumerate}[leftmargin=2em,label=\alph*.]
        \item Pick up (interact) 1 onion and place (interact) it in the pot.
        \item After placing 3 onions in the pot, open (interact) the pot and cook for 5 timesteps. The pot will show how long the soup has been cooked.
        \item When the pot shows the number 5, the soup is finished. Pick up (interact) a dish to plate (interact) the soup.
        \item Deliver the soup and put (interact) it on the serving location.
    \end{enumerate}
\end{enumerate}

\textbf{Visual observation.}
An example is shown in Fig.~\ref{fig:app:overcooked}. On the left is the current game state, showing the overall kitchen layout, the positions and orientations of both chefs, and the items they hold. On the right is a legend explaining the visual representations of game elements, such as objects and chef orientations, used in the game state.

\subsection{Knights Archers Zombies}
\label{app:kaz_obs}

\textbf{Game rules.}

\begin{enumerate}[leftmargin=2em]
    \item \textit{Knights Archers Zombies} (KAZ) is a cooperative survival game played on a 2D battlefield. The objective is to survive as long as possible while maximizing zombie kills and protecting both yourself and your teammate.
    \item Zombies spawn from the top and walk downward toward the bottom along unpredictable paths.
    \item Each player controls either a Knight (melee fighter) or an Archer (ranged fighter), both starting at the bottom of the field.
    \item In the illustration, green units represent zombies, red units represent Archers, and white units represent Knights.
    \item The game ends under either of the following conditions: (a) one agent dies, or (b) a zombie reaches the bottom border.
    \item Rewards are assigned as follows: each zombie killed yields +1 point.
    \item Knights attack with a mace, striking in an arc in front of them, whereas Archers attack by shooting arrows in straight lines.
    \item All agents can move forward or backward and rotate left or right to change their direction.
\end{enumerate}

\textbf{Visual observation.}
An example is shown in Fig.~\ref{fig:app:kaz-origin}. The environment is a two-dimensional plane where zombies randomly spawn from the top and continuously move downward, with their horizontal movement being stochastic. Both players are initialized at the bottom side of the environment.

\subsection{Breakthrough}
\label{app:breakthrough_obs}

\textit{Breakthrough}~\citep{troyka2000breakthrough} is a chess-like board game with simplified rules and identical pawns.
Two players compete to advance their pieces across an $8\times8$ grid to reach the opponent's back row.
The game is deceptively simple, yet it exhibits deep combinatorial complexity and sharp tempo imbalance between attack and defense, making it a suitable environment for studying multi-step lookahead and adversarial decision-making~\citep{lorentz2013programming,saffidine2012solving}.

\textbf{Game rules.}

\begin{enumerate}[leftmargin=2em]
    \item Breakthrough is a two-player strategy game played on an 8x8 grid.
    \item Each player controls pieces of a color: `White' or ``Black''. `White' starts at the bottom (rows 1 and 2), while ``Black'' starts at the top (rows 7 and 8).
    \item If `White' moves a piece to row 8, `White' wins the game. Conversely, if ``Black'' moves a piece to row 1, ``Black'' wins the game.
    \item Players alternate turns, moving one piece per turn, with ``Black'' going first.
    \item A piece may only move one space straight or diagonally forward, and only if the destination square is empty.
    \item A piece may only capture an opponent's piece by moving one space diagonally forward into its square. In this case, the opponent's piece is removed, and your piece takes its place.
    \item ``Black'' moves forward by decreasing row indices (downward), while `White' moves forward by increasing them (upward).
    \item Moves are specified by their start and end positions. For example, 'a2a3' indicates moving a piece from a2 (column a, row 2) to a3 (column a, row 3).
    \item The board is labeled with columns a-h and rows 1-8. Thus, h8 is the top-right corner, and a1 is the bottom-left corner.
\end{enumerate}

\textbf{Visual observation.}
An example is shown in Fig.~\ref{fig:app:breakthrough}.
The figure illustrates the current positions of both black and white pieces on the board. Row and column indices are annotated on the left and bottom sides of the image, respectively.

\subsection{Kuhn Poker}
\label{app:poker_obs}

\textit{Kuhn Poker}~\citep{kuhn1950simplified} is a simplified variant of Texas Hold'em~\citep{brown2019superhuman,moravvcik2017deepstack} designed to study imperfect-information games for game-theoretic analysis.
It has a three-card deck and a single betting round where two players take turns to either check or bet with limited stakes. 
The game has been used as a classic environment for counterfactual reasoning and decision-making with imperfect information~\citep{lanctot2019openspiel,muller2019generalized}.

\textbf{Game rules.}

\begin{enumerate}[leftmargin=2em]
    \item Kuhn poker is a two-player card game. The deck includes only three cards: King (K) > Queen (Q) > Jack (J).
    \item At the start of each game, both player 0 and player 1 place 1 chip into the pot as a blind ante.
    \item Each player is dealt a card as private information, and the third card is set aside unseen.
    \item The two players take turns acting, starting with player 0. A player can choose to:
    \begin{enumerate}[leftmargin=2em,label=\alph*.]
        \item <PASS>: place no additional chips into the pot.
        \item <BET>: place 1 additional chip into the pot.
    \end{enumerate}
    \item If a player chooses to <PASS> after the other player's <BET>, the betting player wins the pot.
    \item If both players choose to <PASS> or both players choose to <BET>, the player with the higher card wins the pot.
\end{enumerate}

\textbf{Visual observation.}
An example is shown in Fig.~\ref{fig:app:kuhn_cards}. 
Each player receives a visual representation of their private card and the current chips in the pot.

\subsection{MPE}
\label{app:mpe_obs}

\textbf{Game rules.}

\begin{enumerate}[leftmargin=2em]
    \item The environment consists of two moving players (a blocker and a charger) and a target position within a two-dimensional space.
    \item The charger's goal is to reach the target location.
    \item The blocker's goal is to prevent the charger from reaching the target by blocking and pushing.
    \item At each step:
    \begin{enumerate}[label=(\alph*), leftmargin=2em]
        \item The charger is rewarded based on its distance to the target --- the smaller the distance, the higher the reward.
        \item The blocker is rewarded when it is close to the target and when the charger remains far from the target (i.e., the difference in distances).
    \end{enumerate}
    \item The player with the higher cumulative reward at the end of the game is declared the winner.
\end{enumerate}

\textbf{Visual observation.}
An example is shown in Fig.~\ref{fig:app:mpe}. The target is represented by a red X. The charger and blocker are represented by green and blue circles, respectively. The cordinate grid on the image indicates the positions and distances between the players and the target.

\subsection{Atari Pong}
\label{app:pong_obs}

\textit{Atari Pong}~\citep{atari1972pong} is a classic arcade video game where two players control paddles to hit a ball across the screen.
With raw pixel observations and competitive dynamics, the game has become a canonical environment in the Arcade Learning Environment (ALE)~\citep{bellemare2013arcade}, which requires spatio-temporal reasoning and strategic gameplay~\citep{mnih2013playing,mnih2015human}.

\textbf{Game rules.}

\begin{enumerate}[leftmargin=2em]
    \item Atari Pong is a zero-sum game played on a 2D screen with two players (left and right) and a ball.
    \item Each player controls a paddle and receives rewards on different events:
    \begin{enumerate}[leftmargin=2em,label=\alph*.]
        \item If the ball passes your paddle: the opponent +1 point.
        \item If the ball passes the opponent's paddle: you +1 point.
    \end{enumerate}
    \item The ball bounces off the top/bottom walls and the paddles.
    \item Paddles can only move vertically within the top and bottom walls.
    \item First player to score 3 points wins.
\end{enumerate}

\textbf{Visual observation.}
An example is shown in Fig.~\ref{fig:app:pong-origin}. The two players each control a paddle on the side of the screen to hit a ball back and forth with each other. The paddles are vertical rectangles, and the ball is a white square. The players score if the ball passes their opponent's paddle. The built-in bot controls the left paddle, while the VLM agent controls the right paddle. The scores of both players are displayed at the top of the screen.

\subsection{Coin's Dilemma}
\label{app:dilemma_obs}

\textit{Coin Dilemma}~\citep{lerer2017maintaining} is a grid-world environment inspired by the classic Prisoner's Dilemma~\citep{rapoport1965prisoner} in game theory.
A red player and a blue player move in a $5\times5$ grid world to collect red and blue coins.
A player earns 1 point for collecting any coin.
However, if the red player collects a blue coin, the blue player is penalized 2 points, and vice versa.
This setup creates a tension between mutual benefit and self-interest: while both players collecting coins of their own color leads to a win-win result, unilateral defection to collect all coins maximizes one's own gains at the other's expense.
Therefore, the game has been a common environment for studying rational reasoning, opponent shaping, and social dilemma resolution~\citep{foerster2017learning,lu2022model,rutherford2024jaxmarl}.

\textbf{Game rules.}

\begin{enumerate}[leftmargin=2em]
    \item Coin Dilemma is a general-sum game played on a 5x5 grid board with two players (red and blue) and two types of coins (red and blue).
    \item Players receive rewards on different events:
    \begin{enumerate}[leftmargin=2em,label=\alph*.]
        \item A player collects one coin of its own color: the player +1 point.
        \item A player collects one coin of the other player's color: the player +1 point, the other player -2 points.
    \end{enumerate}
    \item New coins spawn randomly on the board after each collection.
\end{enumerate}

\textbf{Visual observation.}
An example is shown in Fig.~\ref{fig:app:dilemma}. On the left of the image is a grid map showing the current positions of all game elements, including two players (red and blue) and two coins (red and blue). The players are each represented by a Pac-Man icon, and the coins are each represented by a coin icon. On the right of the image is a table demonstrating the rewards of each event and a corresponding counter tracking the number of occurrences for that event.

\subsection{Monster Hunt}
\label{app:hunt_obs}

\textit{Monster Hunt}~\citep{peysakhovich2017prosocial} is a grid-world environment inspired by the classic Stag Hunt~\citep{rousseau1985discourse} in game theory.
Two players move in a $5\times5$ grid world to individually eat an apple for 2 points or jointly defeat a monster for 5 points.
A player who confronts the monster alone, however, is penalized 2 points.
This leads to multiple Nash equilibria where agents can both safely eat apples alone or take risks to cooperate for higher rewards.
The game is used to investigate trust formation and risk-sensitive decision-making~\citep{leibo2021scalable,tang2021discovering}.

\textbf{Game rules.}

\begin{enumerate}[leftmargin=2em]
    \item Monster Hunt is a general-sum game played on a 5x5 grid board with two players (red and blue), one monster, and two apples.
    \item The monster moves towards the closest player in each step.
    \item Players move in the grid-world and receive rewards on different events:
    \begin{enumerate}[leftmargin=2em,label=\alph*.]
        \item One player eats an apple: the player +2 points and the apple respawns at a random position.
        \item One player encounters the monster alone: the player -2 points and respawns at a random position.
        \item Two players defeat the monster together: both players +5 points and the monster respawns at a random position.
    \end{enumerate}
\end{enumerate}

\textbf{Visual observation.}
An example is shown in Fig.~\ref{fig:app:hunt}. On the left of the image is a grid map showing the current positions of all game elements, including two players (red and blue), two apples, and a monster. The players are each represented by a Pac-Man icon, the apples are each represented by a green apple icon, and the monster is represented by a black demon icon. On the right of the image is a table demonstrating the rewards of each event and a corresponding counter tracking the number of occurrences for that event.

\subsection{Battle of the Colors}
\label{app:battle_obs}

\textit{Battle of the Colors} is a grid-world environment inspired by the classic Battle of the Sexes~\citep{luce2012games} in game theory.
We propose and design this game in a manner similar to the previous two social dilemma games.
A red player and a blue player move in a $5\times5$ grid world with a red block and a blue block.
If both players move to the red block, the red player earns 2 points while the blue player earns 1 point, and vice versa.
If players move to two blocks of different colors, both players earn 0 points.
Therefore, while coordination is mutually beneficial, each player strictly prefers choosing the block of their own color, creating a conflict of interest that produces two payoff-asymmetric Nash equilibria and a mixed equilibrium.
This game thus challenges agents to solve conflicting preferences while avoiding coordination failure, making it suitable for studying equilibrium selection, bargaining dynamics, and social fairness.

\textbf{Game rules.}
\begin{enumerate}[leftmargin=2em]
    \item The Battle of the Colors is a general-sum game played on a 5x5 grid board with two players (red and blue) and two types of blocks (red and blue).
    \item Players receive rewards on different events:
    \begin{enumerate}[leftmargin=2em,label=\alph*.]
        \item When both players are on a red block: red player +2 points, blue player +1 point, and the red block will be refreshed to a new random position.
        \item When both players are on a blue block: red player +1 point, blue player +2 points, and the blue block will be refreshed to a new random position.
        \item When players are on different blocks: both players +0 points, and both blocks will be refreshed to new random positions.
    \end{enumerate}
\end{enumerate}

\textbf{Visual observation.}
An example is shown in Fig.~\ref{fig:app:battle}. On the left of the image is a grid map showing the current positions of all game elements, including two players (red and blue) and two colored blocks (red and blue). The players are each represented by a Pac-Man icon. On the right of the image is a table demonstrating the rewards of each event and a corresponding counter tracking the number of occurrences for that event.

\section{Use of LLMs}
For this manuscript, we use LLMs only to aid and polish the writing. Their use is limited to improving grammar, clarity, and overall readability.

\end{document}